\documentclass[10pt,twocolumn,letterpaper]{article}

\usepackage{wacv}
\usepackage{times}
\usepackage{epsfig}
\usepackage{graphicx}
\usepackage{amsmath}
\usepackage{amssymb}
\usepackage{bm}
\usepackage{xcolor,colortbl}
\usepackage{subfigure}
\usepackage{pifont}
\usepackage{array}
\usepackage{bbold}
\usepackage{nicefrac}
\makeatletter
\@namedef{ver@everyshi.sty}{}
\makeatother
\usepackage{tikz}
\usepackage{tikzscale}
\usepackage{pgfplots}
\pgfplotsset{compat=newest}
\usepackage[mode=buildnew]{standalone}
\usepackage{booktabs}
\usepackage{multirow}
\usepackage{makecell}
\usepackage{siunitx}

\usepackage[pagebackref=true,breaklinks=true,letterpaper=true,colorlinks,bookmarks=false]{hyperref}

\def\wacvPaperID{1} 

\wacvfinalcopy 

\ifwacvfinal
\fi

\ifwacvfinal
\usepackage[breaklinks=true,bookmarks=false]{hyperref}
\else
\usepackage[pagebackref=true,breaklinks=true,colorlinks,bookmarks=false]{hyperref}
\fi

\ifwacvfinal
\else
\fi

\newdimen\owntablesep
\newcommand*{\myplus}{\ensuremath{\boldsymbol{\pmb{+}}}}
\DeclareSymbolFont{AMSb}{U}{msb}{m}{n}
\DeclareSymbolFontAlphabet{\mathbb}{AMSb}
\begin{document}

\title{Unsupervised BatchNorm Adaptation (UBNA): A Domain Adaptation Method for Semantic Segmentation Without Using Source Domain Representations}

\author{Marvin Klingner\quad Jan-Aike Termöhlen\quad Jacob Ritterbach\quad Tim Fingscheidt\\
{\tt\small \{m.klingner, j.termoehlen, j.ritterbach, t.fingscheidt\}@tu-bs.de}\\[1.0em]
Technische Universität Braunschweig, Braunschweig, Germany}

\maketitle
\ifwacvfinal\thispagestyle{empty}\fi

\begin{abstract}
	In this paper we present a solution to the task of ``unsupervised domain adaptation (UDA) of a given pre-trained semantic segmentation model without relying on any source domain representations''. Previous UDA approaches for semantic segmentation either employed simultaneous training of the model in the source and target domains, or they relied on an additional network, replaying source domain knowledge to the model during adaptation. In contrast, we present our novel Unsupervised BatchNorm Adaptation (UBNA) method, which adapts a given pre-trained model to an unseen target domain without using---beyond the existing model parameters from pre-training---any source domain representations (neither data, nor networks) and which can also be applied in an online setting or using just a few unlabeled images from the target domain in a few-shot manner. Specifically, we partially adapt the normalization layer statistics to the target domain using an exponentially decaying momentum factor, thereby mixing the statistics from both domains. By evaluation on standard UDA benchmarks for semantic segmentation we show that this is superior to a model without adaptation and to baseline approaches using statistics from the target domain only. Compared to standard UDA approaches we report a trade-off between performance and usage of source domain representations.\footnote{Code is available at \href{https://github.com/ifnspaml/SGDepth}{\url{https://github.com/ifnspaml/UBNA}}}
\end{abstract}
 
\section{Introduction}
\label{sec:introduction}

\begin{figure}[t]
	\centering	
	\includegraphics[width=1.0\linewidth]{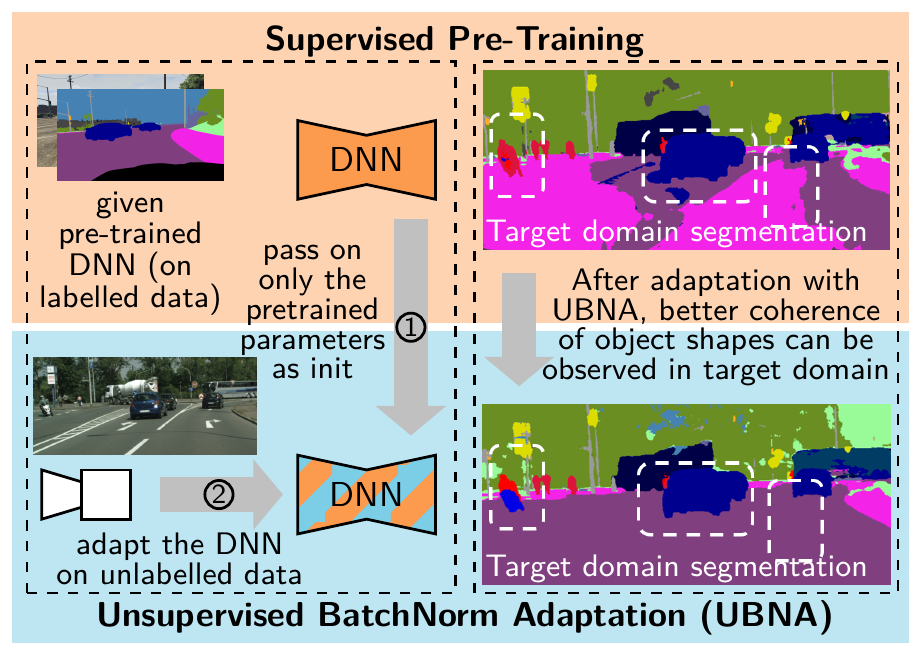}
    \caption{\textbf{UBNA} can be used to \textbf{adapt a given pre-trained model} to the target domain without using source domain representations. Target domain performance clearly improves after adaptation.}
	\label{fig:general_overview}
	\vspace{-0.3cm}
\end{figure} 

Current neural-network-based solutions to many perception tasks in computer vision rely on annotated datasets. However, the generation of these annotations, \ie, ground truth labels, is tedious and time-consuming, in particular for pixel-wise prediction tasks such as semantic segmentation \cite{Cordts2016, Neuhold2017}, depth estimation \cite{Geiger2013, Menze2015, Klingner2020, RaviKumar2021, RaviKumar2021a}, optical flow \cite{Geiger2013, Menze2015}, or instance segmentation \cite{lin2014microsoft}. Due to the domain shift between datasets, neural networks usually cannot be trained in one domain (source domain) and be deployed in a different one (target domain) without a significant loss in performance \cite{Ganin2015}. For applications such as autonomous driving, virtual reality, or medical imaging, these problems are often addressed by unsupervised domain adaptation (UDA) or domain generalization (DG) approaches.
\par
State-of-the-art approaches for UDA of semantic segmentation \cite{Kim2020, Mei2020, Yang2020a} usually employ annotated source domain data during adaptation. Alternatively, other approaches \cite{Kurmi2021, Liu2021} make use of an additional network to replay the source domain knowledge during the adaptation to the model. However, in practice, models are often pre-trained on non-public datasets. When adapting such a given model one might only have access to target domain data, since the source domain data cannot be passed on either for practical reasons or due to data privacy issues. In such cases, DG approaches are also inapplicable, as they have to be applied already during pre-training in the source domain. Therefore, in this work we do not aim at outperforming existing UDA or DG approaches but at solving a different very constrained task, where simultaneous access to source and target domain data is neither allowed nor required and a given model has been pre-trained in the source domain. We dub this task ``UDA for semantic segmentation without using source domain representations'', \ie, a given semantic segmentation model pre-trained in the source domain is adapted to the target domain without utilizing any kind of representation from the source domain beyond the existing model parameters from pre-training.
\par
To create a baseline method for this rather challenging task, we build on advances of UDA techniques on related image classification tasks \cite{Li2018d, Li2020a, Wulfmeier2018, Zhang2020a}, where a few approaches exist for UDA without source domain data. However, these are either not easily transferable to semantic segmentation \cite{Li2020a}, violate our condition of not using any source domain representations \cite{Wulfmeier2018}, or apply a data-dependent evaluation protocol, where the performance on a single image is dependent on other images in the test set \cite{Li2018d, Zhang2020a}. Therefore, we propose the Unsupervised BatchNorm Adaptation (UBNA) algorithm (that can be seen as an improved version of AdaBN \cite{Li2018d}), which adapts the statistics of the batch normalization (BN) layers to the target domain, while leaving all other model parameters untouched. In contrast to AdaBN \cite{Li2018d}, we apply a data-independent evaluation protocol after adaptation of the model. Also, we preserve some knowledge from the source domain BN statistics by not replacing these pre-trained parameters entirely with those from the target domain, instead combining them using an exponential BN momentum decay which leads to a significant boost in adaptation performance.
\par
Our contribution with this work is threefold. Firstly, we (i) outline the task of UDA for semantic segmentation \textit{without using source domain representations}. Secondly, we (ii) present Unsupervised BatchNorm Adaptation (UBNA) as a solution to this task and thereby, thirdly, (iii) enable an online-applicable UDA for semantic segmentation which we show to be operating even in a few-shot learning setup. We achieve a significant improvement over a pre-trained semantic segmentation model without adaptation and we also outperform respective baseline approaches making use of adapting the normalization layers, which we transferred to a comparable setting for semantic segmentation.

\section{Related Work}
\label{sec:related_work}

In this section we first discuss UDA and DG approaches for semantic segmentation. Afterwards, we discuss UDA approaches not relying on source domain data, and UDA approaches relying on the use of normalization layers.
\par
\textbf{UDA with Source Domain Representations}:
Approaches to UDA can be divided into three main categories. Firstly, \textit{style transfer} approaches \cite{Gong2019, Hoffman2018, Li2019b, Yang2020}, where the source domain images get altered to match the style of the target domain. Secondly, \textit{domain-adversarial training} \cite{Bolte2019a, Chen2019c, Dong2020, Du2019, Hoffman2016, Huang2020, Tsai2018, Vu2019, Xu2019b}, where a discriminator network is employed to enforce a domain-invariant feature extraction. Thirdly \textit{self-training}, where pseudo-labels are generated for the target domain samples, which are then used as additional training material \cite{Choi2019, Li2019b, Mei2020, Subhani2020, Zou2018}. Often methods from all three categories are combined to achieve state-of-the-art results \cite{Kim2020, Tranheden2021, Wang2020, Yang2020a, Yang2020, Zhang2019e}. On the other hand, the approaches of \cite{Bateson2020, Kurmi2021, Liu2021, Teja2021, Termoehlen2021, Wulfmeier2018} use auxiliary information (e.g., an auxiliary network) from the source domain during adaptation. While this removes the need for source \textit{data} during adaptation, it imposes the need of additional information about the source domain, which still is a kind of source domain \textit{representation}. In contrast, we propose an adaptation only relying on a few adaptation steps on few unlabelled target domain data, thereby solving a more constrained task in an efficient fashion.
\par 
\textbf{Domain Generalization}: 
Approaches for DG \cite{Dou2019, Li2019c, Seo2020, Yue2019} aim at an improvement in the target domain without having access to target domain data. Only few works exist for semantic segmentation \cite{Choi2021, Yue2019}. In contrast to DG, we assume the source data to be unavailable. Accordingly, we consider DG methods rather applicable during pre-training but not during adaptation of a given model and without access to source data, which we require in our task definition.
\par
\textbf{UDA w/o Source Domain Representations}: 
Improving the performance of a given model in the target domain without using source domain representations during adaptation is challenging and so far rarely addressed. Most techniques rely on training with pseudo-labels in the target domain, generated by the model pre-trained in the source domain \cite{Li2021, Yeh2021}. Additionally, alignment methods for the latent space distribution \cite{Li2020, Liang2020, Yeh2021} can be applied. Other approaches apply selection methods for ``good'' pseudo labels \cite{Li2021}. In a recent approach, Li~\etal~\cite{Li2020a} propose to train a class-conditional generator producing target-style data examples. Collaboration of this generator with the prediction model is shown to improve performance in the target domain without the use of source data. So far the mentioned frameworks are formulated only for image classification or object detection, while we address semantic segmentation. Also, our proposed adaptation method is potentially real-time capable and easier to generalize to other tasks as it requires only a few update steps of the BN statistics parameters and does thereby not rely on gradient optimization.
\par
\textbf{UDA via Normalization Layers}: Normalization techniques \cite{Ba2016, Huang2017b, Ioffe2017, Ioffe2015, Miyato2018, Nam2018, Ulyanov2016, Shao2019, Wu2018} are commonly used in many network architectures. For UDA often new domain-adaptation-specific normalization layers \cite{Chang2019, Mancini2018, Carlucci2017, Romijnders2019, Wang2019d} are proposed, \eg, Chang \etal~\cite{Chang2019} propose to use domain-specific BN statistics but share all other network parameters. Also, the approach of Xu~\etal~\cite{Xu2019b} combines adversarial learning with a BN statistics re-estimation in the target domain. In contrast to our method, these approaches all require training on labeled source data during adaptation.
\par
Other recent works from Li~\etal~\cite{Li2018d, Li2017b} and Zhang~\etal~\cite{Zhang2020a} show that BN statistics can be re-estimated in the target domain on the entire test set or over batches thereof for performance improvements which also can be seen as a kind of UDA without source domain representations. However, both approaches determine the BN statistics during inference in dependency of \textit{all test data}, making the performance on one sample depend on (the availability of) the other samples present in the test set or batch. On the other hand, we use a separate adaptation set, removing the inter-image dependency at test time. Also, our UBNA method shows that it is beneficial not to use statistics solely from the target domain (as in \cite{Li2018d, Li2017b, Zhang2020a}), but to mix source with target statistics by using an exponentially decaying BN momentum factor. Concurrently, Schneider~\etal~\cite{Schneider2020} found similar improvements by mixing statistics from perturbed and clean images for adversarial robustness, which we feel supports our novel finding on the defined UDA task.

\section{Revisiting Batch Normalization, Notations}
\label{sec:theoretical_background}

\begin{figure*}[t]
	\centering	
	\includegraphics[width=1.0\linewidth]{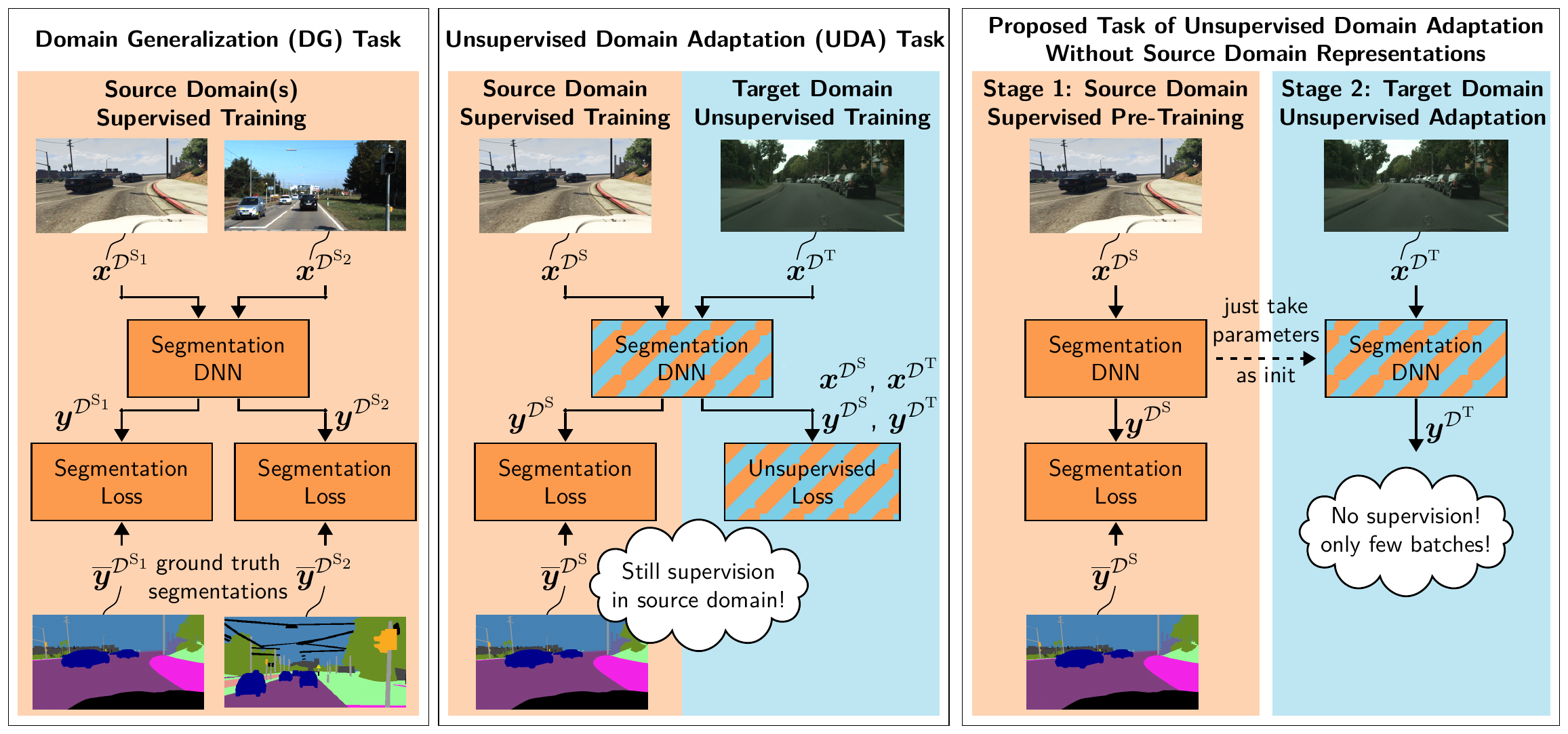}
	\caption{\textbf{Different task setups to improve performance on unseen data} for semantic segmentation: In \textbf{domain generalization} (left) the aim is to generalize to unseen data using diverse source domain data $\mathcal{D}^{\mathrm{S}_1}$, $\mathcal{D}^{\mathrm{S}_2}$. Meanwhile, in \textbf{unsupervised domain adaptation} (center), the model is trained simultaneously on the source domain $\mathcal{D}^{\mathrm{S}}$ and on the target domain $\mathcal{D}^{\mathrm{T}}$, while still using supervision from the source domain during the adaptation. In contrast, we propose a two-stage approach (right): A given semantic segmentation has been pre-trained in the source domain $\mathcal{D}^{\mathrm{S}}$, and in a second stage it is afterwards adapted to the target domain $\mathcal{D}^{\mathrm{T}}$ in an unsupervised fashion.}
	\label{fig:two_stage_training}
	\vspace{-0.12cm}
\end{figure*}

Before we will introduce our UBNA method in Section \ref{sec:uda_approach}, in this section we briefly revisit the batch normalization (BN) layer and thereby introduce necessary notations. Following the initial formulation from \cite{Ioffe2015}, we consider a batch of input feature maps $\bm{f}_\ell\in\mathbb{R}^{B\times H_\ell\times W_\ell\times C_\ell}$ for the BN layer $\ell$ with batch size $B$, height $H_\ell$, width $W_\ell$, and number of channels $C_\ell$. In a convolutional neural network (CNN) each feature $f_{\ell, b, i, c} \in \mathbb{R}$ with batch-internal sample index $b \in \mathcal{B} = \left\lbrace 1, ... , B \right\rbrace$, spatial index $i \in \mathcal{I}_\ell = \left\lbrace 1, ... , H_\ell \cdot W_\ell \right\rbrace$, and channel index $c \in \mathcal{C}_\ell = \left\lbrace 1, ... , C_\ell \right\rbrace$ is batch-normalized according to
\begin{equation}
	\hat{f}_{\ell, b, i, c} = \gamma_{\ell, c} \cdot \left(f_{\ell, b, i, c} - \mu_{\ell, c}\right)\cdot\left(\sigma_{\ell, c}^2 + \epsilon\right)^{-\frac{1}{2}} + \beta_{\ell, c}\;.
	\label{eq:batch_norm}
\end{equation}
Here, $\bm{\gamma}_\ell = (\gamma_{\ell, c}) \in \mathbb{R}^{C_\ell}$ and $\bm{\beta}_\ell = (\beta_{\ell, c}) \in \mathbb{R}^{C_\ell}$ are learnable scaling and shifting parameters, respectively, $\epsilon\in \mathbb{R}$ is a small number, and $\bm{\mu}_\ell = (\mu_{\ell, c}) \in \mathbb{R}^{C_\ell}$ and $\bm{\sigma}_\ell = (\sigma_{\ell, c}) \in \mathbb{R}^{C_\ell}$ are the computed mean and standard deviation, respectively, over the batch samples $b\in\mathcal{B}$ in layer $\ell$ and feature map $c$. Note that in a CNN the statistics are also calculated over the invariant spatial dimension $i$.
\par
At each training step $k$ a batch $\mathcal{B}$ is taken from the set of training samples. Then, mean $\hat{\bm{\mu}}^{(k)}_\ell$ and standard deviation $\hat{\bm{\sigma}}^{(k)}_\ell$ of the features in the current batch $\mathcal{B}$ are computed as
\begin{align}
	\hat{\mu}^{(k)}_{\ell, c} &= \frac{1}{B H_\ell W_\ell}\sum_{b\in \mathcal{B}} \sum_{i\in \mathcal{I}} f_{\ell, b, i, c},
	\label{eq:mean}\\
	\left(\hat{\sigma}^{(k)}_{\ell, c}\right)^2 &= \frac{1}{B H_\ell W_\ell}\sum_{b\in \mathcal{B}} \sum_{i\in \mathcal{I}} \left( f_{\ell, b, i, c} - \hat{\mu}^{(k)}_{\ell, c}\right)^2.
	\label{eq:variance}
\end{align}
During training, the values from (\ref{eq:mean}) and (\ref{eq:variance}) are directly used in (\ref{eq:batch_norm}), meaning $\bm{\mu}_{\ell} = \hat{\bm{\mu}}^{(k)}_{\ell}$ and $\bm{\sigma}_{\ell} = \hat{\bm{\sigma}}^{(k)}_{\ell}$. As preparation for inference, however, mean and variance over the training dataset are tracked progressively using (\ref{eq:mean}) and (\ref{eq:variance}) as 
\begin{align}
  \check{\mu}_{\ell,c}^{(k)} &= \left( 1- \eta \right)\cdot \check{\mu}_{\ell,c}^{(k-1)} + \eta \cdot \hat{\mu}_{\ell,c}^{(k)}, \label{eq:running_mean}\\
  \left(\check{\sigma}_{\ell,c}^{(k)}\right)^2 &= \left( 1- \eta \right)\cdot \left(\check{\sigma}_{\ell,c}^{(k-1)}\right)^2 + \eta \cdot \left(\hat{\sigma}_{\ell,c}^{(k)}\right)^2, \label{eq:running_variance}
\end{align}
where $\eta\in \left[0,1\right]$ is a momentum parameter. After training for $K$ steps, the final values from (\ref{eq:running_mean}) and (\ref{eq:running_variance}) are used for inference, \ie, $\bm{\mu}_{\ell} = \check{\bm{\mu}}^{(K)}_{\ell}$ and $\bm{\sigma}_{\ell} = \check{\bm{\sigma}}^{(K)}_{\ell}$ for the normalization in (\ref{eq:batch_norm}). Note that updating only the BN statistics parameters is computationally not very expensive, as no gradients have to be backpropagated through the network.

\section{Task and Method}
\label{sec:uda_approach}

Here, we define the task of ``unsupervised domain adaptation for semantic segmentation without using source domain representations'', and describe how we solve it by our Unsupervised BatchNorm Adaptation (UBNA) method.

\subsection{UDA w/o Source Domain Representations}
\label{sec:task_definition}

The aim of our task is to improve performance of a semantic segmentation model, trained in a source domain $\mathcal{D}^{\mathrm{S}}$, on a target domain $\mathcal{D}^{\mathrm{T}}$. As shown in Fig.~\ref{fig:two_stage_training} on the right hand side in contrast to previous task definitions (left and middle parts), we introduce a two-stage approach consisting of a supervised pre-training stage and an unsupervised adaptation stage.
\par
\textbf{Stage 1---Supervised Pre-Training}:
In the first stage, the model is trained using image-label pairs $(\bm{x}^{\mathcal{D}^{\mathrm{S}}}, \overline{\bm{y}}^{\mathcal{D}^{\mathrm{S}}})$ from the source domain $\mathcal{D}^{\mathrm{S}}$, yielding the model's output $\bm{y}^{\mathcal{D}^{\mathrm{S}}}$. The model is optimized in a supervised fashion using an arbitrary loss between the model's output $\bm{y}^{\mathcal{D}^{\mathrm{S}}}$ and the ground truth $\overline{\bm{y}}^{\mathcal{D}^{\mathrm{S}}}$. DG methods could also be applied in this stage. After pre-training is finished, the model parameters are passed on to the second stage, the adaptation stage. In particular, neither further explicit information, such as data or labels is passed on, nor any implicit information about the source domain, such as a generator networks.
\par
\textbf{Stage 2---Unsupervised Domain Adaptation}:
In the second stage only image samples $\bm{x}^{\mathcal{D}^{\mathrm{T}}}$ from the target domain $\mathcal{D}^{\mathrm{T}}$ are allowed to be used for adaptation of a given pre-trained model, either through some update rule for the pre-trained weights or some unsupervised loss function. Due to the catastrophic forgetting problem in neural networks, this presents a difficult task, as the semantic segmentation performance could be expected to decrease, if the model parameters are changed without keeping the supervision for the semantic segmentation \cite{Jung2016, Kirkpatrick2016}.
\par 
\textbf{Discussion}:
Following, we analyze this task in terms of (1) applicability, (2) online capability, and (3) comparability w.r.t.~the known tasks of domain generalization (DG) and unsupervised domain adaptation (UDA), as shown in Fig.~\ref{fig:two_stage_training}.
\par 
Our task is \textit{applicable} to a given pre-trained model, whenever one has access to target data. This is in contrast to UDA methods requiring simultaneous access to both source and target data (cannot always be granted due to data privacy issues). DG methods on the other hand require access to source data and are therefore applied during pre-training, which is something we consider ``out of our hands'' for the scope of our adaptation task. This also means that if a model has already been pre-trained and the source data is not available, DG methods cannot be applied, while methods solving our task are still applicable.
\par
Regarding \textit{online capability}, our task provides the general possibility of online adaptation of a model, whereas solutions to DG as of their definition are offline methods applied in pre-training. UDA, on the other hand, relies on the source data and labels, the model was initially trained on, which also limits their online capability, as usually the source data cannot be stored on the target device.
\par
Regarding \textit{comparability}, a fair comparison and conclusive analysis w.r.t.~DG methods is hardly possible as they use additional source data (cf.~Fig.~\ref{fig:two_stage_training}), while solutions to our task use additional target data (the latter being available during inference/adaptation!). Therefore, experimental comparison to DG cannot be provided, due to the strong dependence on the chosen additional pre-training/adaptation data. We can, however, compare to UDA techniques using the same pre-training/adaptation data as required by our task. Such methods are of course expected to outperform solutions to our task due to the additional use of labelled source data during adaptation.

\subsection{UBNA Learning Approach}

Our UBNA approach implements the adaptation (second stage) of our task definition in Fig.~\ref{fig:two_stage_training} (right side) and therefore assumes a pre-trained semantic segmentation model, which is supposed to be adapted to the target domain $\mathcal{D}^{\mathrm{T}}$. Also, our approach assumes that the network contains batch normalization (BN) layers, which were trained as described in Sec.~\ref{sec:theoretical_background}. During our UBNA approach, all learnable parameters are kept constant, only the statistics $\bm{\mu}_\ell$ and $\bm{\sigma}_\ell$ of the BN layers $\ell$ are adapted.
\par 
\textbf{Offline Protocol}:
For the second (the adaptation) stage, we initialize all statistics parameters using (\ref{eq:running_mean}) and (\ref{eq:running_variance}) as
\begin{align}
  \bm{\mu}_\ell^{(\kappa=0)} = \check{\bm{\mu}}_\ell^{(K)}, \bm{\sigma}_\ell^{(\kappa=0)} = \check{\bm{\sigma}}_\ell^{(K)}, \label{eq:init_mean_variance}
\end{align}
with $K$ being the last step of the pre-training stage. Modifying the running mean and variance updates from (\ref{eq:running_mean}) and (\ref{eq:running_variance}), respectively, by a batch-dependent momentum factor $\eta^{(\kappa)}$, we update all BN statistics in each adaptation step $\kappa$ as
\begin{align}
  \mu_{\ell,c}^{(\kappa)} &= \left( 1\!-\! \eta^{(\kappa)} \right)\cdot \mu_{\ell,c}^{(\kappa-1)} + \eta^{(\kappa)} \cdot \hat{\mu}_{\ell,c}^{(\kappa)}, \label{eq:running_mean_adapted}\\
  \left(\sigma_{\ell,c}^{(\kappa)}\right)^2 \!\!\!&= \left( 1\!- \!\eta^{(\kappa)} \right)\cdot \left(\sigma_{\ell,c}^{(\kappa-1)}\right)^2 \!\!+ \eta^{(\kappa)} \cdot \left(\hat{\sigma}_{\ell,c}^{(\kappa)}\right)^2\!\!\!, \label{eq:running_variance_adapted}
\end{align}
with each batch $\mathcal{B}$ being sampled randomly from the target domain. The batch-dependent momentum factor $\eta^{(\kappa)}$ is supposed to achieve a compromise between two conflicting goals: On the one hand, we want to adapt the BN statistics to the target domain. On the other hand, the initial statistics were optimized to the weights performing a task in the source domain, which is why the mismatch between statistics parameters and network weights should not become too large. Conclusively, we choose a decay protocol
\begin{equation}
    \eta^{(\kappa)} = \eta^{(0)} \exp\left(-\kappa\cdot \alpha^{\mathrm{BATCH}} \right),
    \label{eq:momentum_decay}
\end{equation}
determined by a batch-wise decay factor $\alpha^{\mathrm{BATCH}}$, which adapts the statistics to the target domain in a progressively decaying amount, while still keeping the task-specific knowledge learned in the source domain. When using a decay factor of $\alpha^{\mathrm{BATCH}} > 0$, we dub our method \textbf{UBNA}, and if using $\alpha^{\mathrm{BATCH}} = 0$ (no decay), we dub it \textbf{UBNA}$^\mathbf{0}$.
\par
\textbf{Online Protocol}:
Introducing a constraint into our offline protocol, we can derive an online-capable version of our approach. Here, we impose that the batches contain temporally ordered images, such that the adaptation step index $\kappa$ is equal to a time index $t\;\left[\mathrm{sec}\right]$ up to a scale factor. Assuming a batch size $B$ and a frame rate $\frac{1}{\Delta t}\;\left[\mathrm{frames/sec}\right]$, this constraint can be defined as $\frac{t}{\Delta t\cdot B} = \kappa \in \mathbb{N}$.
\par
\textbf{Layer-Wise Weighting}:
Previous approaches (\eg, \cite{Pan2018}) indicate that the domain mismatch occurs more strongly in the initial layers. This can be considered by weighting the batch-dependent momentum factor $\eta^{(\kappa)}$ in (\ref{eq:momentum_decay}) differently for each BN layer $\ell\in\left\lbrace1,.., L\right\rbrace$, with the total number of BN layers $L$, as 
\begin{equation}
    \eta_{\ell}^{(\kappa)}\!\! = \!\eta^{(\kappa)} \exp\!\left(-\ell\cdot\alpha^{\mathrm{LAYER}}\right).
    \label{eq:layer_decay}
\end{equation}
The weighting is determined through the factor $\alpha^{\mathrm{LAYER}}$.
\par
If $\alpha^{\mathrm{LAYER}}, \alpha^{\mathrm{BATCH}} > 0$, we dub our method \textbf{UBNA}$^{\myplus}$, meaning that the statistics parameters in the initial layers are adapted more rapidly than in the later layers. This also follows the idea to further improve the trade-off between adapting to the target domain (with the initial layers) and keeping the task-specific knowledge learned during pre-training in the source domain (in the later layers).

\section{Implementation Details}
\label{sec:implementation_details}

Following, we describe the experimental details of our framework which is implemented in \texttt{PyTorch} \cite{Paszke2019}.
\par
\begin{table}[t]
  \centering
  \small
  \setlength{\tabcolsep}{6pt}
  \caption{\textbf{Number of used images} in the databases for pre-training (source domain $\mathcal{D}^{\mathrm{S}}$) and adaptation (target domain $\mathcal{D}^{\mathrm{T}}$).}
  \begin{tabular}{|l|c|rrr|}
  \hline
  \multirow{2}{*}{Dataset} & \multirow{2}{*}{Domain} & train\;\;\; & \multirow{2}{*}{validate} & \multirow{2}{*}{test}\\[-0.1cm]
   & & or adapt & & \\
  \hline
  \rule{0pt}{2.2ex}\!
  GTA-5 \cite{Richter2016} & $\mathcal{D}^{\mathrm{S}}$ & 24,966 & - & -\\
  SYNTHIA \cite{Ros2016} & $\mathcal{D}^{\mathrm{S}}$ & 9,400 & - & -\\
  Cityscapes \cite{Cordts2016} & $\mathcal{D}^{\mathrm{S}}$ & 2,975 & - & -\\
  \hline
  \rule{0pt}{2.2ex}\!
  Cityscapes \cite{Cordts2016} & $\mathcal{D}^{\mathrm{T}}$ & 89,250 & 500 & 1,525\\
  KITTI \cite{Geiger2013, Menze2015} & $\mathcal{D}^{\mathrm{T}}$ & 29,000 & 200 & 200\\
  \hline
  \end{tabular}
  \label{tab:dataset_overview}
\end{table}
\textbf{Datasets}:
To pre-train the semantic segmentation models we use the synthetic datasets GTA-5 \cite{Richter2016} and SYNTHIA \cite{Ros2016}, as well as the real dataset Cityscapes \cite{Cordts2016} as source domains $\mathcal{D}^{\mathrm{S}}$ as defined in Tab.~\ref{tab:dataset_overview}. For adaptation, we use the training splits of Cityscapes and KITTI as target domains $\mathcal{D}^{\mathrm{T}}$. We evaluate on the respective validation sets.
\par
\textbf{Network Architecture}:
We use an encoder-decoder architecture with skip connections as in \cite{Godard2019, Klingner2020a}, where we modify the last layer to have $19$ feature maps, which are converted to pixel-wise class probabilities using a softmax function. For comparability to previous works \cite{Kim2020, Wang2020, Yang2020}, the encoder is based on the commonly used ResNet \cite{He2016} and VGG-16 \cite{Simonyan2015} architectures, (pre-)pre-trained on ImageNet \cite{Russakovsky2015}. If not mentioned otherwise, we use VGG-16.
\par
\textbf{Training Details}: For pre-training, we resize all images to a resolution of $1024 \times 576$, $1024 \times 608$, and $1024 \times 512$ for images from GTA-5, SYNTHIA, and Cityscapes, respectively. Afterwards, we randomly crop to a resolution of $640 \times 192$. We apply horizontal flipping, random brightness ($\pm 0.2$), contrast ($\pm 0.2$), saturation ($\pm 0.2$) and hue ($\pm 0.1$) transformations as input augmentations. We pre-train our segmentation models for $20$ epochs (1 epoch corresponds to approximately $10,000$ training steps) with the ADAM optimizer~\cite{Kingma2015}, a batch size of $12$, and a learning rate of $10^{-4}$, which is reduced to $10^{-5}$ after $15$ epochs. For adaptation we use resolutions of $1024 \times 512$ and $1024 \times 320$ for Cityscapes and KITTI, respectively. Here, we use a batch size of $6$ but only adapt for $50$ steps. Our first setting introduces a constant momentum factor $\eta^{(0)}\!=\!0.1$, which we dub \textbf{UBNA}$^{\mathbf{0}}$ ($\eta^{(0)}\!=\!0.1$, $\alpha^{\mathrm{BATCH}}\! =\!\alpha^{\mathrm{LAYER}}\! =\! 0$). For our \textbf{UBNA} model we use a batch-wise decay factor of $\alpha^{\mathrm{BATCH}} \!=\! 0.08$ ($\eta^{(0)}\!=\!0.1$, $\alpha^{\mathrm{BATCH}}\! =\! 0.08$, $\alpha^{\mathrm{LAYER}}\! =\! 0$). Additionally, we apply the layer-wise weighting of our \textbf{UBNA}$^{\myplus}$ method, which, however, requires target-domain-specific tuning of hyperparameters: $\eta^{(0)}\!=\!0.1$, $\alpha^{\mathrm{BATCH}}\! =\! 0.08$, $\alpha^{\mathrm{LAYER}}\! =\! 0.03$ when adapting to Cityscapes, and  $\eta^{(0)}\!=\!0.1$, $\alpha^{\mathrm{BATCH}}\! =\! 0.08$, $\alpha^{\mathrm{LAYER}}\! =\! 0.3$ when adapting to KITTI. As common for UDA methods, we determined optimal hyperparameters on target domain validation sets. An analysis is given in the supplementary.
\par
\textbf{Evaluation Metrics}:
The semantic segmentation is evaluated using the mean intersection over union (mIoU) \cite{Everingham2015}, which is computed considering the classes defined in \cite{Cordts2016}. We evaluate over all 19 defined classes, except for models, which have been pre-trained on SYNTHIA. Here, we follow \cite{Lee2019d, Vu2019a} and evaluate over subsets of 13 and 16 classes. We also evaluate the per-class intersection over union (IoU).

\section{Experimental Evaluation}
\label{sec:experiments}

To evaluate our method, we start by comparing our approach to other approaches that make use of adapting the BN layers, and to standard UDA approaches using source data supervision during adaptation. Afterwards, we show the online and few-shot capability of UBNA and an extensive ablation study over different network architectures.

\subsection{Comparison to Normalization Approaches}

\begin{table}[t]
  \centering
  \small
  \setlength{\tabcolsep}{4pt}
  \caption{\textbf{Comparison to normalization approaches:} Performance on the \textbf{Cityscapes validation set} for the adaptation from \textbf{SYNTHIA} ($\mathcal{D}^{\mathrm{S}}$, third column) or \textbf{GTA-5} ($\mathcal{D}^{\mathrm{S}}$, fourth column) \textbf{to Cityscapes} ($\mathcal{D}^{\mathrm{T}}$) in \textbf{comparison to other normalization approaches} (batch size $B=6$). The methods from \cite{Li2018d} and \cite{Zhang2020a} have been reimplemented by us for comparability.}
  \begin{tabular}{|l|c|c|c|}
  \hline
  & & & \\[-2.3ex]
  Method & \shortstack{Source data\\ during\\[-0.07cm] adaptation} & \shortstack{$\mathcal{D}^{\mathrm{S}}$:\,SYNTHIA\\ mIoU (\%)\\ (16 classes)} &  \shortstack{$\mathcal{D}^{\mathrm{S}}$:\,GTA-5\\ mIoU (\%)\\ (19 classes)} \\
  \hline
  No adaptation & - & 30.0 & 31.5 \\
  \hline
  Li~\etal~\cite{Li2018d} & no & 31.2 & 34.7 \\ 
  Zhang~\etal~\cite{Zhang2020a} & no & 31.5 & 34.6 \\ 
  \hline
  \rowcolor[gray]{.90} & & & \\[-2.3ex]
  \rowcolor[gray]{.90} \textbf{UBNA}$^{\mathbf{0}}$ & no & 32.6 & 33.9\\
  \rowcolor[gray]{.90} \textbf{UBNA} & no &  34.4 & 36.1\\
  \rowcolor[gray]{.90} \textbf{UBNA}$^{\myplus}$ & no & \textbf{34.6} & \textbf{36.5}\\
  \hline
  \end{tabular}
  \label{tab:baseline_comparison}
\end{table}

\begin{figure}[t]
	\centering	
\begin{tikzpicture}

\definecolor{color0}{rgb}{0.690196078431373,0,0.274509803921569}

\begin{axis}[
height=6.6cm,
legend cell align={left},
legend columns=3,
legend style={at={(0.03,0.03)}, anchor=south west, draw=white!80.0!black},
tick align=outside,
tick pos=left,
width=10cm,
x grid style={white!69.01960784313725!black},
xlabel={adaptation step \(\displaystyle \kappa\)},
xmajorgrids,
xmin=0, xmax=50,
xtick style={color=black},
y grid style={white!69.01960784313725!black},
ylabel={mIoU (\%)},
ymajorgrids,
ymin=28.2, ymax=35,
ytick style={color=black}
]
\addplot [ultra thick, black]
table {%
0 31.2
1 31.2
2 31.2
3 31.2
4 31.2
5 31.2
6 31.2
7 31.2
8 31.2
9 31.2
10 31.2
11 31.2
12 31.2
13 31.2
14 31.2
15 31.2
16 31.2
17 31.2
18 31.2
19 31.2
20 31.2
21 31.2
22 31.2
23 31.2
24 31.2
25 31.2
26 31.2
27 31.2
28 31.2
29 31.2
30 31.2
31 31.2
32 31.2
33 31.2
34 31.2
35 31.2
36 31.2
37 31.2
38 31.2
39 31.2
40 31.2
41 31.2
42 31.2
43 31.2
44 31.2
45 31.2
46 31.2
47 31.2
48 31.2
49 31.2
50 31.2
51 31.2
52 31.2
53 31.2
54 31.2
55 31.2
56 31.2
57 31.2
58 31.2
59 31.2
60 31.2
61 31.2
62 31.2
63 31.2
64 31.2
65 31.2
66 31.2
67 31.2
68 31.2
69 31.2
70 31.2
71 31.2
72 31.2
73 31.2
74 31.2
75 31.2
76 31.2
77 31.2
78 31.2
79 31.2
80 31.2
81 31.2
82 31.2
83 31.2
84 31.2
85 31.2
86 31.2
87 31.2
88 31.2
89 31.2
90 31.2
91 31.2
92 31.2
93 31.2
94 31.2
95 31.2
96 31.2
97 31.2
98 31.2
99 31.2
100 31.2
};
\addlegendentry{AdaBN}
\addplot [ultra thick, color0]
table {%
0 29.9826242058554
1 31.1780935317302
2 31.9781178591381
3 32.60204796299
4 33.1283648621121
5 33.5724418579193
6 33.7534830360398
7 33.9627028622632
8 34.0846248827983
9 34.21475591195
10 34.3967549264806
11 34.4757596849604
12 34.52420749847
13 34.5082458796502
14 34.5111091814618
15 34.5247864434412
16 34.4975628846184
17 34.4947270118576
18 34.510408701305
19 34.486713528441
20 34.4944122571556
21 34.4916822070275
22 34.4794352049936
23 34.4696380650237
24 34.4732482991839
25 34.4613707505843
26 34.455880507918
27 34.4418636304911
28 34.439713201092
29 34.4337956934822
30 34.4365733120212
31 34.432631474411
32 34.4316834300631
33 34.421826587311
34 34.4191789535374
35 34.4111803225829
36 34.4054035113264
37 34.4027864111449
38 34.4011884134684
39 34.3999409688869
40 34.394232707142
41 34.3922358802364
42 34.3906973718101
43 34.3894219052916
44 34.3829927978964
45 34.3815177188954
46 34.3761600206689
47 34.3726957225817
48 34.3717790025313
49 34.3724544779075
50 34.3678348565632
51 34.3662788980631
52 34.3658449172908
53 34.3632675538867
54 34.3626643659783
55 34.3632598191545
56 34.3627270006196
57 34.362426198524
58 34.36198566994
59 34.3614839355083
60 34.3618225710663
61 34.3599664118497
62 34.3604630917977
63 34.3598262099318
64 34.3600675080781
65 34.3587728620061
66 34.3583392411855
67 34.3573626471362
68 34.3566134259948
69 34.3565709055056
70 34.3564376245388
71 34.3560979077033
72 34.3560767441238
73 34.3558986373926
74 34.3557436128877
75 34.3553352446354
76 34.3554555079457
77 34.3550423043908
78 34.3546393164448
79 34.3544758196376
80 34.3543629541668
81 34.3543198491348
82 34.3541269765342
83 34.3539703568151
84 34.3539566098044
85 34.3538559947608
86 34.3538717129108
87 34.3539605151668
88 34.35384691278
89 34.3538787511111
90 34.3538778883835
91 34.3538802953479
92 34.353873368103
93 34.3537950166301
94 34.3537586397896
95 34.3536851902317
96 34.3536677672573
97 34.3536418717456
98 34.3536929279815
99 34.3536907367624
100 34.3536382963486
};
\addlegendentry{\textbf{UBNA}}
\addplot [ultra thick, color0, mark=+, mark size=2, mark options={solid}]
table {%
0 29.9826242058554
1 31.1407711476842
2 31.8453316515388
3 32.4179772668105
4 32.6251785188274
5 33.0144412335723
6 33.2882669082197
7 33.4472131482184
8 33.647806544281
9 33.8019733360534
10 33.9380311216941
11 34.0534802406778
12 34.1194573674833
13 34.1837860715235
14 34.2361575899173
15 34.2835041957207
16 34.3569216607059
17 34.425310771898
18 34.4687458159367
19 34.477834230626
20 34.4663122539987
21 34.4893092160889
22 34.4990176009812
23 34.5297951872081
24 34.5326151205781
25 34.5391153171739
26 34.5576596422795
27 34.5693995588554
28 34.5840922975311
29 34.593331676425
30 34.5989664583622
31 34.6026363748213
32 34.6129084619417
33 34.6163425989863
34 34.6178284622294
35 34.6177894916154
36 34.6178195052045
37 34.6161373060985
38 34.617399274686
39 34.6168628863642
40 34.6189672915814
41 34.6166426679086
42 34.6189284757364
43 34.6192191970046
44 34.6217363628328
45 34.6253613333187
46 34.6273923462549
47 34.6267312663895
48 34.628279499657
49 34.6302915334833
50 34.6308142399583
51 34.629941818123
52 34.6295071810859
53 34.6298585070389
54 34.6316353394729
55 34.6319506894942
56 34.6320444077033
57 34.6318542187447
58 34.6335695016539
59 34.6339365292623
60 34.6334950765207
61 34.6334560985674
62 34.6335873285009
63 34.6334111948185
64 34.6332997453571
65 34.6340358090861
66 34.6340730049986
67 34.6343739181529
68 34.6343536803536
69 34.6346124266745
70 34.6347270138858
71 34.6345067154829
72 34.6346829144803
73 34.6342939365299
74 34.6344949203831
75 34.6345076706173
76 34.6344543044269
77 34.6344239076126
78 34.6345460968198
79 34.6345869692788
80 34.6347582654365
81 34.6346067260653
82 34.634504412942
83 34.634462895664
84 34.6343488339641
85 34.6343697079179
86 34.6345555606762
87 34.6344113527881
88 34.6344620683592
89 34.6344777417385
90 34.634563259328
91 34.6347164625608
92 34.634687067623
93 34.6347571678084
94 34.6347506640113
95 34.6346128543165
96 34.6346918480316
97 34.6347114822481
98 34.6346554223674
99 34.6346862537933
100 34.6346838441529
};
\addlegendentry{\textbf{UBNA}$^\mathbf{+}$}
\addplot [ultra thick, black, dashed]
table {%
0 30
1 30
2 30
3 30
4 30
5 30
6 30
7 30
8 30
9 30
10 30
11 30
12 30
13 30
14 30
15 30
16 30
17 30
18 30
19 30
20 30
21 30
22 30
23 30
24 30
25 30
26 30
27 30
28 30
29 30
30 30
31 30
32 30
33 30
34 30
35 30
36 30
37 30
38 30
39 30
40 30
41 30
42 30
43 30
44 30
45 30
46 30
47 30
48 30
49 30
50 30
51 30
52 30
53 30
54 30
55 30
56 30
57 30
58 30
59 30
60 30
61 30
62 30
63 30
64 30
65 30
66 30
67 30
68 30
69 30
70 30
71 30
72 30
73 30
74 30
75 30
76 30
77 30
78 30
79 30
80 30
81 30
82 30
83 30
84 30
85 30
86 30
87 30
88 30
89 30
90 30
91 30
92 30
93 30
94 30
95 30
96 30
97 30
98 30
99 30
100 30
};
\addlegendentry{No Adaptation}
\addplot [ultra thick, color0, dashed]
table {%
0 29.9826242058554
1 31.2620157485482
2 32.197168756559
3 32.8218895388373
4 33.3318682418114
5 33.7828859626637
6 34.0599876432565
7 34.3219013977919
8 34.3490790897911
9 34.4135002370201
10 34.4112598714818
11 34.3017224746242
12 34.2506091522749
13 34.1976558178821
14 34.2863291003467
15 34.2215678532053
16 34.1145898156467
17 34.0423319926071
18 33.8202849500409
19 33.6836920778888
20 33.6040892362431
21 33.4233646921032
22 33.4040228100871
23 33.3255220934777
24 33.2514768972883
25 33.1284343210386
26 32.9866884996768
27 33.0461041060335
28 33.0636377225689
29 32.9028084253772
30 32.8466020368038
31 32.874996022055
32 32.8104427605819
33 32.7233670139594
34 32.7761548754831
35 32.8288577593931
36 32.7817339164116
37 32.6836351103661
38 32.7009556266779
39 32.5836399526638
40 32.7534438005015
41 32.8151897417498
42 32.7472599442101
43 32.5831760089069
44 32.6001053048032
45 32.4675305640635
46 32.4226613279153
47 32.477967586635
48 32.6040192866218
49 32.6968126618246
50 32.6254593452372
51 32.4909342385982
52 32.4356679963425
53 32.2661749135889
54 32.3221387499379
55 32.3695180399295
56 32.3236045958249
57 32.4135689621668
58 32.3313513299739
59 32.3992477946524
60 32.2860804592834
61 32.2599062007025
62 32.2606048151469
63 32.3672676929013
64 32.3844203951184
65 32.3912951592114
66 32.4152328354487
67 32.399910398877
68 32.4392155002839
69 32.3553531105657
70 32.2556737872675
71 32.2432979870068
72 32.3529299139968
73 32.3086256937705
74 32.3104294531082
75 32.2735478542085
76 32.3365349948267
77 32.3342653655457
78 32.3520573074558
79 32.3244490584857
80 32.3005387074125
81 32.2484733930384
82 32.1627623000403
83 32.2156974014983
84 32.2361427457128
85 32.2621160361098
86 32.4651920705307
87 32.3235382323986
88 32.3723176851941
89 32.360476126142
90 32.3348323151862
91 32.3147175674119
92 32.3522256403074
93 32.2572434465683
94 32.2623321834927
95 32.3121815894685
96 32.3471727596828
97 32.3665098167439
98 32.4239483592762
99 32.3276828668505
100 32.4647373717176
};
\addlegendentry{\textbf{UBNA}$^\mathbf{0}$}
\end{axis}

\end{tikzpicture}
	\put(-152.5,48){\footnotesize \cite{Li2018d}}
	\caption{\textbf{Comparison to normalization approaches}: Performance on the \textbf{Cityscapes validation set} for the adaptation from \textbf{SYNTHIA} ($\mathcal{D}^{\mathrm{S}}$) \textbf{to Cityscapes} ($\mathcal{D}^{\mathrm{T}}$) in \textbf{dependence of the adaptation step} $\kappa$ (batch size $B=6$). We also show results when using the statistics from the source domain (no adaptation) and from the target domain (AdaBN by Li~\etal~\cite{Li2018d}).}
	\label{fig:baseline_comparison}
\end{figure}
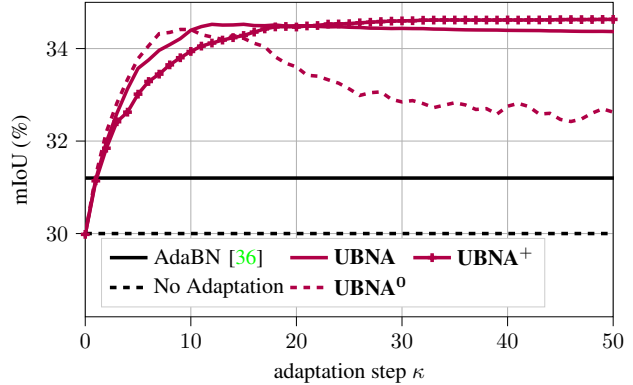

While for semantic segmentation no baseline approaches exist that we could report on our task, we still want to facilitate a comparison. Therefore, we reimplemented the approaches of Li~\etal~(AdaBN,~\cite{Li2018d}), who effectively recalculate the statistics of the BN layers on the target domain, and Zhang~\etal~\cite{Zhang2020a}, who evaluate using the statistics from just a single batch of images during testing. Due to fairness for AdaBN,~\cite{Li2018d} we also used the adaptation dataset to determine the target domain statistics, while for Zhang~\etal~\cite{Zhang2020a} this was not possible due to the batch-wise normalization. As can be seen in Tab.~\ref{tab:baseline_comparison} and Fig.~\ref{fig:baseline_comparison}, both baseline approaches are able to improve the model performance in the target domain. On the other hand, a simple experiment with constant BN momentum (UBNA$^{\mathbf{0}}$) in Fig.~\ref{fig:baseline_comparison} shows that after a few batches there is a maximum in performance, which is a typical behaviour we observed on all considered datasets (cf.~supplementary material). UBNA$^{\mathbf{0}}$ exceeds baseline performance on the SYNTHIA to Cityscapes but not on the GTA-5 to Cityscapes adaptation, yet indicating that it is beneficial not to discard the source domain statistics entirely. As UBNA$^{\mathbf{0}}$ does not show stable convergence, we propose our UBNA and UBNA$^{+}$ methods both achieving a stable convergence to some maximum performance after $\kappa=50$  steps by using a decaying BN momentum factor as described in (\ref{eq:momentum_decay}). Interestingly, the hyperparameter $\alpha^{\mathrm{BATCH}}=0.08$ can be determined independently of the used dataset combination which shows the general applicability of UBNA. We outperform all baseline approaches on all used dataset combinations as can be seen in Tab.~\ref{tab:baseline_comparison}.

\subsection{Comparison to UDA Approaches}

\begin{table*}[t]
  \centering
  \caption{\textbf{Comparison to UDA}: Performance of \textbf{UBNA} and \textbf{UBNA}$^{\myplus}$ (right task, Fig.~\ref{fig:two_stage_training}) in \textbf{comparison to UDA methods} (center task, Fig.~\ref{fig:two_stage_training}) on the \textbf{Cityscapes validation set} for the adaptation from \textbf{GTA-5} ($\mathcal{D}^{\mathrm{S}}$) \textbf{to Cityscapes} ($\mathcal{D}^{\mathrm{T}}$) (upper part) and \textbf{SYNTHIA} ($\mathcal{D}^{\mathrm{S}}$) \textbf{to Cityscapes} ($\mathcal{D}^{\mathrm{T}}$) (lower part). We evaluate \textbf{VGG-16}-based models, compare different methods (second column, values from the respective papers), show if they use labeled source data during UDA (third column), evaluate the class-wise IoU performance (\%), and give an mIoU (\%) over 19 classes. For SYNTHIA we evaluate over 16 classes (excluding classes marked with $*$) and over 13 classes (excluding classes marked with $**$) as is common in the field \cite{Lee2019d, Vu2019}. Best results for UDA with and w/o source data, respectively, are printed in boldface.}
  \footnotesize
  \setlength{\tabcolsep}{1.35pt}
  \begin{tabular}{c|l|c|ccccccccccccccccccc|c|c}
   \rotatebox{90}{Source data} & Method  & \shortstack{Source\\ data \\ during \\[-0.07cm] adaptation} & \rotatebox{90}{road} & \rotatebox{90}{sidewalk} & \rotatebox{90}{building} & \rotatebox{90}{wall$^{**}$} & \rotatebox{90}{fence$^{**}$} & \rotatebox{90}{pole$^{**}$} & \rotatebox{90}{traffic light} & \rotatebox{90}{traffic sign} & \rotatebox{90}{vegetation} & \rotatebox{90}{terrain$^{*}$} & \rotatebox{90}{sky} & \rotatebox{90}{person} & \rotatebox{90}{rider} & \rotatebox{90}{car} & \rotatebox{90}{truck$^{*}$} & \rotatebox{90}{bus} & \rotatebox{90}{on rails$^{*}$} & \rotatebox{90}{motorbike} & \rotatebox{90}{bike} & \multicolumn{2}{c}{\shortstack{$\mathrm{mIoU}$ (\%) \\ (19 classes)}}\\
  \hline
  \hline
  \multirow{6}{*}{\rotatebox{90}{GTA-5}} & Vu~\etal~\cite{Vu2019} & yes & 
  86.9 & 28.7 & 78.7 & 28.5 & 25.2 & 17.1 & 20.3 & 10.9 & 80.0 & 26.4 & 70.2 & 47.1 & 8.4 & 81.5 & 26.0 & 17.2 & \textbf{18.9} & 11.7 & 1.6 & \multicolumn{2}{c}{36.1} \\
   & Dong~\etal~\cite{Dong2020} & yes & 
  89.8 & \textbf{46.1} & 75.2 & 30.1 & \textbf{27.9} & 15.0 & 20.4 & 18.9 & 82.6 & \textbf{39.1} & 77.6 & 47.8 & 17.4 & 76.2 & 28.5 & \textbf{33.4} & 0.5 & 29.4 & \textbf{30.8} & \multicolumn{2}{c}{41.4} \\
   & Yang~\etal\cite{Yang2020a} & yes &
  \textbf{90.1} & 41.2 & \textbf{82.2} & \textbf{30.3} & 21.3 & \textbf{18.3} & \textbf{33.5} & \textbf{23.0} & \textbf{84.1} & 37.5 & \textbf{81.4} & \textbf{54.2} & \textbf{24.3} & \textbf{83.0} & \textbf{27.6} & 32.0 & 8.1 & \textbf{29.7} & 26.9 & \multicolumn{2}{c}{\textbf{43.6}} \\
  \cline{2-24}
   & \cellcolor[gray]{.90} No adaptation & \cellcolor[gray]{.90} - &\cellcolor[gray]{.90} 55.8 &\cellcolor[gray]{.90} 21.9 &\cellcolor[gray]{.90} 65.9 &\cellcolor[gray]{.90} 15.2 &\cellcolor[gray]{.90} 14.7 &\cellcolor[gray]{.90} 27.5 &\cellcolor[gray]{.90} \textbf{31.0} &\cellcolor[gray]{.90} 17.9 &\cellcolor[gray]{.90} \textbf{77.8} &\cellcolor[gray]{.90} \textbf{19.5} &\cellcolor[gray]{.90} 74.4 &\cellcolor[gray]{.90} 55.2 &\cellcolor[gray]{.90} 12.1 &\cellcolor[gray]{.90} 71.7 &\cellcolor[gray]{.90} 11.9 &\cellcolor[gray]{.90} 3.3 &\cellcolor[gray]{.90} 0.5 &\cellcolor[gray]{.90} 13.2 &\cellcolor[gray]{.90} 9.6 & \multicolumn{2}{|c}{\cellcolor[gray]{.90} 31.5}\\
  \cline{2-24}
   & \cellcolor[gray]{.90} \textbf{UBNA} & \cellcolor[gray]{.90} no &\cellcolor[gray]{.90} \textbf{80.8} &\cellcolor[gray]{.90} 29.4 &\cellcolor[gray]{.90} 77.6 &\cellcolor[gray]{.90} 19.8 &\cellcolor[gray]{.90} \textbf{17.1} &\cellcolor[gray]{.90} \textbf{33.9} &\cellcolor[gray]{.90} 29.3 &\cellcolor[gray]{.90} 20.5 &\cellcolor[gray]{.90} 73.9 &\cellcolor[gray]{.90} 16.8 &\cellcolor[gray]{.90} 76.7 &\cellcolor[gray]{.90} 58.3 &\cellcolor[gray]{.90} \textbf{15.2} &\cellcolor[gray]{.90} 79.1 &\cellcolor[gray]{.90} 13.6 &\cellcolor[gray]{.90} 12.5 &\cellcolor[gray]{.90} 5.7 &\cellcolor[gray]{.90} 14.1 &\cellcolor[gray]{.90} \textbf{10.8} & \multicolumn{2}{|c}{\cellcolor[gray]{.90} 36.1}\\
    & \cellcolor[gray]{.90} \textbf{UBNA}$^{\myplus}$ & \cellcolor[gray]{.90} no &\cellcolor[gray]{.90} 79.9 &\cellcolor[gray]{.90} \textbf{29.9} &\cellcolor[gray]{.90} \textbf{78.1} &\cellcolor[gray]{.90} \textbf{21.1} &\cellcolor[gray]{.90} 16.5 &\cellcolor[gray]{.90} 33.8 &\cellcolor[gray]{.90} 29.7 &\cellcolor[gray]{.90} \textbf{20.6} &\cellcolor[gray]{.90} 75.6 &\cellcolor[gray]{.90} 18.4 &\cellcolor[gray]{.90} \textbf{78.0} &\cellcolor[gray]{.90} \textbf{58.4} &\cellcolor[gray]{.90} 14.6 &\cellcolor[gray]{.90} \textbf{79.4} &\cellcolor[gray]{.90} \textbf{14.8} &\cellcolor[gray]{.90} \textbf{13.0} &\cellcolor[gray]{.90} \textbf{5.8} &\cellcolor[gray]{.90} \textbf{14.6} &\cellcolor[gray]{.90} 10.6 & \multicolumn{2}{|c}{\cellcolor[gray]{.90} \textbf{36.5}}\\
   \hline
   \hline
   & & & & & & & & & & & & & & & & & & & & & & 16 cl. & 13 cl.\\
  \hline
  \hline
  \multirow{6}{*}{\rotatebox{90}{SYNTHIA}}  & Lee~\etal~\cite{Lee2019d} & yes & 
   71.1 & 29.8 & 71.4 & 3.7 & 0.3 & \textbf{33.2} & 6.4 & 15.6 & \textbf{81.2} & - & 78.9 & 52.7 & 13.1 & 75.9 & - & 25.5 & - & 10.0 & 20.5 & 36.8 & \textbf{42.4}\\
   & Dong~\etal~\cite{Dong2020} & yes & 
  70.9 & \textbf{30.5} & \textbf{77.8} & \textbf{9.0} & \textbf{0.6} & 27.3 & 8.8 & 12.9 & 74.8 & - & 81.1 & 43.0 & \textbf{25.1} & 73.4 & - & \textbf{34.5} & - & \textbf{19.5} & 38.2 & 39.2 & - \\
   & Yang~\etal~\cite{Yang2020a} & yes & 
  \textbf{73.7} & 29.6 & 77.6 & 1.0 & 0.4 & 26.0 & \textbf{14.7} & \textbf{26.6} & 80.6 & - & \textbf{81.8} & \textbf{57.2} & 24.5 & \textbf{76.1} & - & 27.6 & - & 13.6 & \textbf{46.6} & \textbf{41.1} & - \\
  \cline{2-24}
   & \cellcolor[gray]{.90} No adaptation &\cellcolor[gray]{.90} - &\cellcolor[gray]{.90} 49.4 &\cellcolor[gray]{.90} 20.8 &\cellcolor[gray]{.90} 61.5 &\cellcolor[gray]{.90} \textbf{3.6} &\cellcolor[gray]{.90} 0.1 &\cellcolor[gray]{.90} 30.5 &\cellcolor[gray]{.90} \textbf{13.6} &\cellcolor[gray]{.90} 14.1 &\cellcolor[gray]{.90} 74.4 &\cellcolor[gray]{.90}  - & \cellcolor[gray]{.90} 75.5 &\cellcolor[gray]{.90} \textbf{53.5} &\cellcolor[gray]{.90} 10.6 &\cellcolor[gray]{.90} 47.2 &\cellcolor[gray]{.90}  - & \cellcolor[gray]{.90} 4.8 &\cellcolor[gray]{.90}  - & \cellcolor[gray]{.90} 3.0 &\cellcolor[gray]{.90} 17.1 &\cellcolor[gray]{.90} 30.0 &\cellcolor[gray]{.90} 35.2 \\
  \cline{2-24}
   & \cellcolor[gray]{.90} \textbf{UBNA} &\cellcolor[gray]{.90} no &\cellcolor[gray]{.90} \textbf{72.3} &\cellcolor[gray]{.90} 26.6 &\cellcolor[gray]{.90} \textbf{73.0} &\cellcolor[gray]{.90} 2.3 &\cellcolor[gray]{.90} \textbf{0.3} &\cellcolor[gray]{.90} 31.5 &\cellcolor[gray]{.90} 12.1 &\cellcolor[gray]{.90} 16.6 &\cellcolor[gray]{.90} 72.1 &\cellcolor[gray]{.90}  - & \cellcolor[gray]{.90} \textbf{75.6} &\cellcolor[gray]{.90} 45.4 &\cellcolor[gray]{.90} \textbf{13.6} &\cellcolor[gray]{.90} 61.2 &\cellcolor[gray]{.90}  - & \cellcolor[gray]{.90} \textbf{8.5} &\cellcolor[gray]{.90}  - & \cellcolor[gray]{.90} \textbf{8.5} &\cellcolor[gray]{.90} \textbf{30.1} &\cellcolor[gray]{.90} 34.4 &\cellcolor[gray]{.90} 40.7 \\
   & \cellcolor[gray]{.90} \textbf{UBNA}$^{\myplus}$ &\cellcolor[gray]{.90} no &\cellcolor[gray]{.90} 71.5 &\cellcolor[gray]{.90} \textbf{27.3} &\cellcolor[gray]{.90} 72.9 &\cellcolor[gray]{.90} 2.5 &\cellcolor[gray]{.90} \textbf{0.3} &\cellcolor[gray]{.90} \textbf{32.0} &\cellcolor[gray]{.90} 12.7 &\cellcolor[gray]{.90} \textbf{16.7} &\cellcolor[gray]{.90} \textbf{74.6} &\cellcolor[gray]{.90} - & \cellcolor[gray]{.90} 75.4 &\cellcolor[gray]{.90} 47.1 &\cellcolor[gray]{.90} \textbf{13.6} &\cellcolor[gray]{.90} \textbf{61.4} &\cellcolor[gray]{.90}  - & \cellcolor[gray]{.90} \textbf{8.5} &\cellcolor[gray]{.90}  - & \cellcolor[gray]{.90} 8.3 &\cellcolor[gray]{.90} 29.2 &\cellcolor[gray]{.90} \textbf{34.6} &\cellcolor[gray]{.90} \textbf{41.0} \\
  \end{tabular}
  \label{tab:gta5_to_cityscapes_val}
\end{table*}


Another type of comparison can be drawn to standard UDA approaches, which we present for the widely accepted standard benchmarks GTA-5 to Cityscapes and SYNTHIA to Cityscapes in Table~\ref{tab:gta5_to_cityscapes_val}. We would like to emphasize here that the white and gray parts in these tables are not completely comparable, as standard UDA approaches (white parts) make use of source data (cf.~middle task in Fig.~\ref{fig:two_stage_training}) during the adaptation, while our approach does not make use of this supervision (cf.~right task in Fig.~\ref{fig:two_stage_training}). Still, some conclusions can be drawn: Firstly, the improvement and final performance of our model is not as high as for standard UDA approaches, which is to be expected due to our much more constrained task of an adaptation without source data, where UDA approaches are not even applicable. Secondly, Tab.~\ref{tab:gta5_to_cityscapes_val} shows that we improve the baselines without adaptation by $5.0\%$ and $4.6\%/5.8\%$ absolute on the two adaptations from GTA-5 to Cityscapes and SYNTHIA to Cityscapes, respectively. Moreover, our UBNA$^+$ model improves the baseline in the respective single classes in 16 out of 19 and 12 out of 16 cases in terms of IoU, which shows a pretty consistent improvement for the majority of classes. Accordingly, our method still significantly improves a baseline trained without any adaptation. For a more detailed qualitative discussion, we refer to the supplementary. In conclusion, these observations show a trade-off between final performance and usage of labelled source data during adaptation compared to UDA approaches.

\subsection{Few-Shot Capability}

\begin{table}[t]
  \centering
  \small
  \setlength{\tabcolsep}{3.5pt}
  \caption{\textbf{Few-shot adaptation}: Performance of \textbf{UBNA}$^{\myplus}$ on the \textbf{Cityscapes validation set} for the adaptation from \textbf{SYNTHIA} ($\mathcal{D}^{\mathrm{S}}$, top row) or \textbf{GTA-5} ($\mathcal{D}^{\mathrm{S}}$, bottom row) \textbf{to Cityscapes} ($\mathcal{D}^{\mathrm{T}}$) for \textbf{different numbers of total images used for adaptation} (average over 10 different experiments). Results after the $50$th adaptation step, where in a single experiment the same batch is used in all adaptation steps, showing the few-shot capability of our approach.}
  \begin{tabular}{|l|c|ccccc|}
  \hline
  \# of images $B$ per batch & - & 1 & 2 & 3 & 5 & 10\\
  \hline
  mIoU (\%) (16 classes) & 30.0 & 33.1 & 33.3 & 34.2 & \textbf{34.3} & \textbf{34.3}\\
  mIoU (\%) (19 classes) & 31.0 & 34.9 & 35.6 & \textbf{36.3} & 35.9 & 36.2\\
  \hline
  \end{tabular}
  \label{tab:few_shot_applicability}
\end{table}

To investigate how many target domain images are necessary for a successful adaptation, we use only \textit{a single image batch} for $50$ steps of adaptation and thereby show that our method is indeed few-shot capable. For these experiments we chose the best performing method UBNA$^+$. In Tab.~\ref{tab:few_shot_applicability} we show results for a network adapted with only $B=1$, $2$, $3$, $5$, or $10$ images in the batch. As we suspected that the choice of images might have a significant impact on the final result, the results in Tab.~\ref{tab:few_shot_applicability} are averaged over 10 different experiments. Here, \textit{on average}, we can observe performance improvements even for the adaptation \textit{with a single image} (see supplementary for more detailed results). Indeed, one can observe a significant and consistent improvement in \textit{any} of the 10 experiments with an average improvement from $3\%$ to about $5\%$ absolute, which shows our method's few-shot capability.

\subsection{Online (i.e.,\ Video Adaptation) Capability}

As our method does not require the source dataset to be available during adaptation, we are able to adapt the segmentation network in an online setting using \textit{sequential} images from a video. As our target domain adaptation data we use the video from Stuttgart (images in sequential order), which is provided as part of the Cityscapes dataset \cite{Cordts2016}. Tab.\ \ref{tab:online_applicability} gives UBNA$^+$ results when using different numbers of sequential images per batch, meaning that the total number of sequential images is proportional to the batch size, while we adapt for 50 steps in all experiments. We observe that in all experiments we can improve the baseline by about $3\%$ to $4\%$ absolute. Also, using the VGG-16 backbone, inference of the segmentation network with or without UBNA is realtime-capable with \SI{23}{fps} on an \texttt{NVIDIA GeForce GTX 1080 Ti} graphics card, since the computational effort to update of the BN statistics using (\ref{eq:running_mean_adapted}) and (\ref{eq:running_variance_adapted}) is negligible. The frame rate of Cityscapes is about \SI{16.66}{fps}, accordingly a UBNA model utilizing a VGG-16 backbone would be real time-capable consuming only $\frac{16.66}{23}= 73\%$ of the available GPU computation power.

\begin{table}[t]
  \centering
  \small
  \setlength{\tabcolsep}{2.9pt}
  \caption{\textbf{Online adaptation}: Performance of \textbf{UBNA}$^{\myplus}$ on the \textbf{Cityscapes validation set} for the adaptation from \textbf{SYNTHIA} ($\mathcal{D}^{\mathrm{S}}$, top row) or \textbf{GTA-5} ($\mathcal{D}^{\mathrm{S}}$, bottom row) \textbf{to Cityscapes} ($\mathcal{D}^{\mathrm{T}}$), where the \textbf{adaptation images are sequentially taken from a video}. Results after the $50$th adaptation step, whereby the adaptation batches contain temporally ordered images from a video sequence, showing the online applicability of our approach.}
  \begin{tabular}{|l|c|ccccc|}
  \hline
  \# of images $B$ per batch & - & 1 & 2 & 3 & 5 & 10 \\
  \hline
  mIoU (\%) (16 classes) & 30.0 & \textbf{33.7} & 33.5 & 33.5 & 33.5 & \textbf{33.7}\\
  mIoU (\%) (19 classes) & 31.0 & \textbf{35.6} & 35.5 & 35.5 & 35.3 & 35.5\\
  \hline
  \end{tabular}
  \label{tab:online_applicability}
\end{table}

\subsection{Ablation Studies}

\begin{table}[t]
  \centering
  \small
  \setlength{\tabcolsep}{5pt}
  \caption{\textbf{Network topology ablation}: Performance of \textbf{UBNA}$^{\mathbf{0}}$, \textbf{UBNA}, and \textbf{UBNA}$^{\myplus}$ on the \textbf{KITTI validation set} for the adaptation from \textbf{Cityscapes} ($\mathcal{D}^{\mathrm{S}}$) \textbf{to KITTI} ($\mathcal{D}^{\mathrm{T}}$) for \textbf{various network topologies}; best results for each network topology in boldface.}
  \begin{tabular}{|l|c|c|c|c|}
  \hline
  Network & No adaptation & \textbf{UBNA}$^{\mathbf{0}}$ & \textbf{UBNA} & \textbf{UBNA}$^{\myplus}$\\
  \hline
  VGG-16 & 51.1 & 55.9 & 57.1 & \textbf{58.4} \\
  VGG-19 & 53.7 & 54.2 & 58.6 & \textbf{60.2} \\
  \hline
  ResNet-18 & 49.9 & 52.0 & 55.5 & \textbf{58.1} \\
  ResNet-34 & 54.7 & 56.7 & 58.2 & \textbf{60.3} \\
  ResNet-50 & 46.9 & 56.1 & 56.4 & \textbf{56.9} \\
  ResNet-101 & 51.9 & 54.5 & 57.1 & \textbf{58.2} \\
  \hline
  \end{tabular}
  \vspace{-0.15cm}
  \label{tab:network_architectures}
\end{table}

\begin{table}[t]
  \centering
  \small
  \setlength{\tabcolsep}{5pt}
  \caption{\textbf{Test set generalization}: Performance of \textbf{UBNA}$^{\mathbf{0}}$, \textbf{UBNA}, and \textbf{UBNA}$^{\myplus}$ on the \textbf{KITTI validation and test set} when adapting from \textbf{Cityscapes} ($\mathcal{D}^{\mathrm{S}}$) \textbf{to KITTI} ($\mathcal{D}^{\mathrm{T}}$); \textbf{ResNet-34}.}
  \begin{tabular}{|l|c|c|c|}
  \hline
  & & & \\[-2.3ex]
  Method & \shortstack{Source data\\ during\\[-0.07cm] adaptation} & \shortstack{mIoU (\%)\\[-0.07cm] on val. set\\ (19 classes)} & \shortstack{mIoU (\%)\\[-0.03cm] on test set\\ (19 classes)} \\
  \hline
  No adaptation & - & 54.7 & 49.8\\
  \hline
   & & & \\[-2.3ex]
  \textbf{UBNA}$^{\mathbf{0}}$ & no & 56.7 & 52.3 \\
  \textbf{UBNA} & no & 58.2 & 53.5\\
  \textbf{UBNA}$^{\myplus}$ & no & \textbf{60.3} & \textbf{55.0}\\
  \hline
  \end{tabular}
  \vspace{-0.15cm}
  \label{tab:kitti_benchmark}
\end{table}

While our main experiments were carried out in the commonly used synthetic-to-real settings, we also want to show the generalizability of our method to a real-to-real setting from Cityscapes ($\mathcal{D}^{\mathrm{S}}$) to KITTI ($\mathcal{D}^{\mathrm{T}}$) and across different architectures, which we showcase in Tab.~\ref{tab:network_architectures}. We compare our non-adapted models with our adapted versions for several VGG and ResNet architectures and achieve significant improvements when applying our UBNA and UBNA$^+$ methods. It is important to note that we used the same hyperparameters for all adaptations, which underlines generalizability of our methods across different network architectures (all UBNA methods) and dataset choices (UBNA$^{\mathbf{0}}$, UBNA). Moreover, UBNA$^{+}$ yields improved results compared to UBNA for all network architectures by using the layer-wise weighting from (\ref{eq:layer_decay}). Thereby, our best model utilizing a ResNet-34 backbone achieves an impressive mIoU of $60.3\%$ without using any source data during adaptation. While the UBNA$^{+}$ works consistently better in the real-to-real setting, the performance gains in the synthetic-to-real settings are rather small (cf.~supplementary material). We suspect that with a larger domain gap, an adaptation focusing only on initial layers is not sufficient.
\par
Finally, we evaluate and compare our best model (ResNet-34) on the KITTI benchmark in Tab.~\ref{tab:kitti_benchmark}. We observe that the improvements of the different variants of our UBNA method over the no adaptation baseline are reproducible on the test set, which we did not use during any of the other experiments, again underlining the generalizability of our hyperparameter setting.

\subsection{Sequential Domain Adaptation}

\begin{figure}[t]
    \centering
	\includegraphics[width=1.0\linewidth]{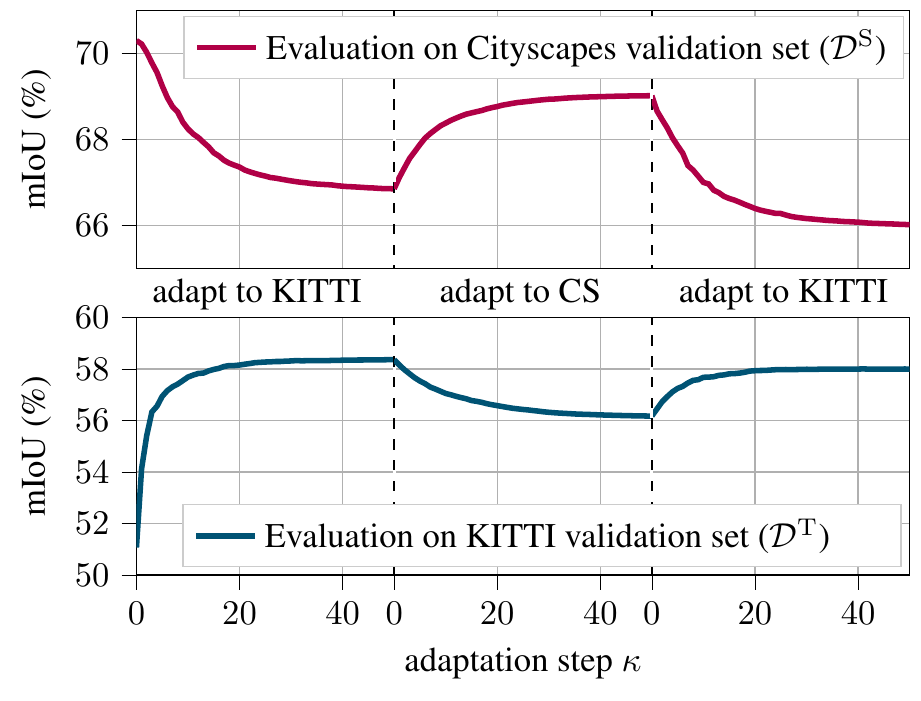}
	\caption{\textbf{Sequential Adaptation}: Performance of \textbf{UBNA}$^{\myplus}$ on the \textbf{Cityscapes (upper graph) and KITTI (lower graph) validation sets} for the adaptation from \textbf{Cityscapes} ($\mathcal{D}^{\mathrm{S}}$) \textbf{to KITTI} ($\mathcal{D}^{\mathrm{T}}$) (left), and after 50 adaptation steps \textbf{back to Cityscapes} ($\mathcal{D}^{\mathrm{T}}$) (middle), then \textbf{back to KITTI} ($\mathcal{D}^{\mathrm{T}}$) (right), in dependence on the number of adaptation batches; \textbf{VGG-16}; mIoU values in \%.}
	\vspace{-0.1cm}
	\label{fig:source_target_tradeoff_plus}
\end{figure} 

While in UDA the primary focus is mostly on a single adaptation to the target domain, it is also of interest to investigate, to what extent this process can be reversed as, \eg, the model should be adapted from day to night and back to day during deployment. As for GTA-5 and SYNTHIA usually the entire dataset is used for training, for this study we use an adaptation setting from Cityscapes to KITTI, where we have available validation sets in both domains and evaluate our VGG-16-based model (cf.~Tab.~\ref{tab:network_architectures}) when adapting with UBNA$^{+}$. Fig.~\ref{fig:source_target_tradeoff_plus} shows that while the performance on KITTI ($\mathcal{D}^{\mathrm{T}}$) improves by about $7\%$, the performance on Cityscapes ($\mathcal{D}^{\mathrm{S}}$) decreases by only $3\%$. Interestingly, this effect can even be reversed to a large extent when adapting back to Cityscapes, and even another time, when adapting back to KITTI. This indicates that our approach can be used in a sequential fashion to continuously adapt between two domains as long as the time of domain switch is known (this is needed to re-initialize the momentum factor in (\ref{eq:layer_decay})).

\section{Conclusions}
\label{sec:conclusion}

In this work we present the task of ``unsupervised domain adaptation (UDA) for semantic segmentation without using source domain representations'' and propose our novel Unsupervised BatchNorm Adaptation (UBNA) method as a solution. We show that UBNA works in various settings for UDA, specifically GTA-5 to Cityscapes, SYNTHIA to Cityscapes, and Cityscapes to KITTI, where we improve the VGG-16-based baseline by $5.0\%$, $4.6\%/5.8\%$, and $7.3\%$ absolute, respectively, without the use of any source domain representations for adaptation. We also outperform various baseline approaches, making use of adapting the normalization layers and show that our approach is applicable in an online setting and in a few-shot fashion. Our approach can be beneficial as long as the considered network architecture utilizes batch normalization layers. Under this condition, our UBNA method generalizes over 3 dataset combinations and 6 network architectures by using the same set of hyperparameters, while results can be further improved with UBNA$^+$ and dataset-specific hyperparameters. Also, we expect that our method is potentially generalizable to other computer vision tasks and that a combination of our method with domain generalization helps improving further.

\clearpage
{\small
\bibliographystyle{ieee_fullname}
\bibliography{bib/ifn_spaml_bibliography}

\begin{thebibliography}{10}\itemsep=-1pt

\bibitem{Ba2016}
Jimmy~Lei Ba, Jamie~Ryan Kiros, and Geoffrey~E. Hinton.
\newblock {Layer Normalization}.
\newblock {\em arXiv}, July 2016.

\bibitem{Bateson2020}
Mathilde Bateson, Hoel Kervadec, Jose Dolz, Herv{\'e} Lombaert, and Ismail~Ben
  Ayed.
\newblock {Source-Relaxed Domain Adaptation for Image Segmentation}.
\newblock In {\em Proc. of MICCAI}, pages 490--499, Lima, Peru, Oct. 2020.

\bibitem{Bolte2019a}
Jan-Aike Bolte, Markus Kamp, Antonia Breuer, Silviu Homoceanu, Peter Schlicht,
  Fabian {H{\"u}ger}, Daniel Lipinski, and Tim Fing\-scheidt.
\newblock {Unsupervised Domain Adaptation to Improve Image Segmentation Quality
  Both in the Source and Target Domain}.
\newblock In {\em Proc. of CVPR - Workshops}, pages 1404--1413, Long Beach, CA,
  USA, June 2019.

\bibitem{Chang2019}
Woong-Gi Chang, Tackgeun You, Seonguk Seo, Suha Kwak, and Bohyung Han.
\newblock {Domain-Specific Batch Normalization for Unsupervised Domain
  Adaptation}.
\newblock In {\em Proc. of CVPR}, pages 7354--7362, Long Beach, CA, USA, June
  2019.

\bibitem{Chen2019c}
Minghao Chen, Hongyang Xue, and Deng Cai.
\newblock {Domain Adaptation for Semantic Segmentation With Maximum Squares
  Loss}.
\newblock In {\em Proc. of ICCV}, pages 2090--2099, Seoul, Korea, Oct. 2019.

\bibitem{Choi2019}
Jaehoon Choi, Taekyung Kim, and Changick Kim.
\newblock {Self-Ensembling With GAN-Based Data Augmentation for Domain
  Adaptation in Semantic Segmentation}.
\newblock In {\em Proc. of ICCV}, pages 6830--6840, Seoul, Korea, Oct. 2019.

\bibitem{Choi2021}
Sungha Choi, Sanghun Jung, Huiwon Yun, Joanne~T. Kim, Seungryong Kim, and
  Jaegul Choo.
\newblock {RobustNet: Improving Domain Generalization in Urban-Scene
  Segmentation via Instance Selective Whitening}.
\newblock In {\em Proc. of CVPR}, pages 11580--11590, Virtual, June 2021.

\bibitem{Cordts2016}
Marius Cordts, Mohamed Omran, Sebastian Ramos, Timo Rehfeld, Markus Enzweiler,
  Rodrigo Benenson, Uwe Franke, Stefan Roth, and Bernt Schiele.
\newblock {The Cityscapes Dataset for Semantic Urban Scene Understanding}.
\newblock In {\em Proc. of CVPR}, pages 3213--3223, Las Vegas, NV, USA, June
  2016.

\bibitem{Dong2020}
Jiahua Dong, Yang Cong, Gan Sun, Yuyang Liu, and Xiaowei Xu.
\newblock {CSCL: Critical Semantic-Consistent Learning for Unsupervised Domain
  Adaptation}.
\newblock In {\em Proc. of ECCV}, pages 745--762, Glasgow, UK, Aug. 2020.

\bibitem{Dou2019}
Qi Dou, Daniel Coelho~de Castro, Konstantinos Kamnitsas, and Ben Glocker.
\newblock {Domain Generalization via Model-Agnostic Learning of Semantic
  Features}.
\newblock In {\em Proc. of NeurIPS}, pages 6447--6458, Vancouver, Canada, Dec.
  2019.

\bibitem{Du2019}
Liang Du, Jingang Tan, Hongye Yang, Jianfeng Feng, Xiangyang Xue, Qibao Zheng,
  Xiaoqing Ye, and Xiaolin Zhang.
\newblock {SSF-DAN: Separated Semantic Feature Based Domain Adaptation Network
  for Semantic Segmentation}.
\newblock In {\em Proc. of ICCV}, pages 982--991, Seoul, Korea, Oct. 2019.

\bibitem{Everingham2015}
Mark Everingham, Luc Van~Gool, Christopher K.~I. Williams, John Winn, and
  Andrew Zisserman.
\newblock {The Pascal Visual Object Classes Challenge: A Retrospective}.
\newblock {\em International Journal of Computer Vision (IJCV)},
  111(1):98--136, Jan. 2015.

\bibitem{Ganin2015}
Yaroslav Ganin and Victor Lempitsky.
\newblock {Unsupervised Domain Adaptation by Backpropagation}.
\newblock In {\em Proc. of ICML}, pages 1180--1189, Lille, France, July 2015.

\bibitem{Geiger2013}
Andreas Geiger, Philip Lenz, Christoph Stiller, and Raquel Urtasun.
\newblock {Vision Meets Robotics: The KITTI Dataset}.
\newblock {\em International Journal of Robotics Research (IJRR)},
  32(11):1231--1237, Aug. 2013.

\bibitem{Godard2019}
Cl{\'e}ment Godard, Oisin Mac~Aodha, Michael Firman, and Gabriel~J. Brostow.
\newblock {Digging Into Self-Supervised Monocular Depth Estimation}.
\newblock In {\em Proc. of ICCV}, pages 3828--3838, Seoul, Korea, Oct. 2019.

\bibitem{Gong2019}
Rui Gong, Wen Li, Yuhua Chen, and Luc~Van Gool.
\newblock {DLOW: Domain Flow for Adaptation and Generalization}.
\newblock In {\em Proc. of CVPR}, pages 2477--2486, Long Beach, CA, USA, June
  2019.

\bibitem{He2016}
Kaiming He, Xiangyu Zhang, Shaoqing Ren, and Jian Sun.
\newblock {Deep Residual Learning for Image Recognition}.
\newblock In {\em Proc. of CVPR}, pages 770--778, Las Vegas, NV, USA, June
  2016.

\bibitem{Hoffman2018}
Judy Hoffman, Eric Tzeng, Taesung Park, Jun-Yan Zhu, Philip Isola, Kate Saenko,
  Alexei~A. Efros, and Trevor Darrell.
\newblock {{CyCADA}: Cycle-Consistent Adversarial Domain Adaptation}.
\newblock In {\em Proc. of ICML}, pages 1989--1998, Stockholm, Sweden, July
  2018.

\bibitem{Hoffman2016}
Judy Hoffman, Dequan Wang, Fischer Yu, and Trevor Darrell.
\newblock {FCNs in the Wild: Pixel-Level Adversarial and Constraint-Based
  Adaptation}.
\newblock {\em arXiv}, (1612.02649), Dec. 2016.

\bibitem{Huang2020}
Jiaxing Huang, Shijian Lu, Dayan Guan, and Xiaobing Zhang.
\newblock {Contextual-Relation Consistent Domain Adaptation for Semantic
  Segmentation}.
\newblock In {\em Proc of ECCV}, pages 705--722, Glasow, UK, Aug. 2020.

\bibitem{Huang2017b}
Xun Huang and Serge Belongie.
\newblock {Arbitrary Style Transfer in Real-Time With Adaptive Instance
  Normalization}.
\newblock In {\em Proc. of ICCV}, pages 1501--1510, Venice, Italy, Oct. 2017.

\bibitem{Ioffe2017}
Sergey Ioffe.
\newblock {Batch Renormalization: Towards Reducing Minibatch Dependence in
  Batch-Normalized Models}.
\newblock In {\em Proc. of NIPS}, pages 1945--1953, Long Beach, CA, USA, Dec.
  2017.

\bibitem{Ioffe2015}
Sergey Ioffe and Christian Szegedy.
\newblock {Batch Normalization: Accelerating Deep Network Training by Reducing
  Internal Covariate Shift}.
\newblock In {\em Proc. of ICML}, pages 448--456, Lille, France, July 2015.

\bibitem{Jung2016}
H. Jung, J. Ju, M. Jung, and J. Kim.
\newblock {Less-Forgetting Learning in Deep Neural Networks}.
\newblock {\em arXiv}, (1607.00122), 2016.

\bibitem{Kim2020}
Myeongjin Kim and Hyeran Byun.
\newblock {Learning Texture Invariant Representation for Domain Adaptation of
  Semantic Segmentation}.
\newblock In {\em Proc. of CVPR}, pages 12975--12984, Seattle, WA, USA, June
  2020.

\bibitem{Kingma2015}
Diederik~P. Kingma and Jimmy Ba.
\newblock {Adam: A Method for Stochastic Optimization}.
\newblock In {\em Proc. of ICLR}, pages 1--15, San Diego, CA, USA, May 2015.

\bibitem{Kirkpatrick2016}
J. Kirkpatrick, R. Pascanu, N. Rabinowitz, J. Veness, G. Desjardins, A.~A.
  Rusu, K. Milan, J. Quan, T. Ramalho, A. Grabska-Barwinska, D. Hassabis, C.
  Clopath, D. Kumarana, and R. Hadsella.
\newblock {Overcoming Catastrophic Forgetting in Neural Networks}.
\newblock {\em Proceedings of the National Academy of Sciences},
  114(13):3521--3526, 2016.

\bibitem{Klingner2020}
Marvin Klingner, Andreas B\"{a}r, and Tim Fing\-scheidt.
\newblock {Improved Noise and Attack Robustness for Semantic Segmentation by
  Using Multi-Task Training with Self-Supervised Depth Estimation}.
\newblock In {\em Proc. of CVPR - Workshops}, pages 1299--1309, Seattle, WA,
  USA, June 2020.

\bibitem{Klingner2020a}
Marvin Klingner, Jan-Aike Term\"{o}hlen, Jonas Mikolajczyk, and Tim
  Fing\-scheidt.
\newblock {Self-Supervised Monocular Depth Estimation: Solving the Dynamic
  Object Problem by Semantic Guidance}.
\newblock In {\em Proc. of ECCV}, pages 582--600, Glasgow, UK, Aug. 2020.

\bibitem{Kurmi2021}
Vinod~K. Kurmi, Venkatesh~K. Subramanian, and Vinay~P. Namboodiri.
\newblock {Domain Impression: A Source Data Free Domain Adaptation Method}.
\newblock In {\em Proc. of WACV}, pages 615--625, Virtual Conference, Jan.
  2021.

\bibitem{Lee2019d}
Kuan-Hui Lee, German Ros, Jie Li, and Adrien Gaidon.
\newblock {SPIGAN: Privileged Adversarial Learning from Simulation}.
\newblock In {\em Proc. of ICLR}, pages 1--14, New Orleans, LA, USA, Apr. 2019.

\bibitem{Li2019c}
Da Li, Jianshu Zhang, Yongxin Yang, Cong Liu, Yi-Zhe Song, and Timothy~M.
  Hospedales.
\newblock {Episodic Training for Domain Generalization}.
\newblock In {\em Proc. of ICCV}, pages 1446--1455, Seoul, Korea, Oct. 2019.

\bibitem{Li2020a}
Rui Li, Qianfen Jiao, Wenming Cao, Hau-San Wong, and Si Wu.
\newblock {Model Adaptation: Unsupervised Domain Adaptation Without Source
  Data}.
\newblock In {\em Proc. of CVPR}, pages 9641--9650, Seattle, WA, USA, June
  2020.

\bibitem{Li2020}
Shunkai Li, Xin Wang, Yingdian Cao, Fei Xue, Zike Yan, and Hongbin Zha.
\newblock {Self-Supervised Deep Visual Odometry With Online Adaptation}.
\newblock In {\em Proc. of CVPR}, pages 6339--6348, Seattle, WA, USA, June
  2020.

\bibitem{Li2021}
Xianfeng Li, Weijie Chen, Di Xie, Shicai Yang, Peng Yuan, Shiliang Pu, and
  Yueting Zhuang.
\newblock {A Free Lunch for Unsupervised Domain Adaptive Object Detection
  Without Source Data}.
\newblock In {\em Proc. of AAAI}, pages xxxx--xxxx, Virtual Conference, Feb.
  2021.

\bibitem{Li2018d}
Yanghao Li, Naiyan Wang, Jianping Shi, Xiaodi Hou, and Jiaying Liu.
\newblock {Adaptive Batch Normalization for Practical Domain Adaptation}.
\newblock {\em Pattern Recognition}, 80:109--117, Aug. 2018.

\bibitem{Li2017b}
Yanghao Li, Naiyan Wang, Jianping Shi, Jiaying Liu, and Xiaodi Hou.
\newblock {Revisiting Batch Normalization for Practical Domain Adaptation}.
\newblock In {\em Proc. of ICLR - Workshops}, pages 1--10, Toulon, France, Apr.
  2017.

\bibitem{Li2019b}
Yunsheng Li, Lu Yuan, and Nuno Vasconcelos.
\newblock {Bidirectional Learning for Domain Adaptation of Semantic
  Segmentation}.
\newblock In {\em Proc. of CVPR}, pages 6936--6945, Long Beach, CA, USA, June
  2019.

\bibitem{Liang2020}
Jian Liang, Dapeng Hu, and Jiashi Feng.
\newblock {Do We Really Need to Access the Source Data? Source Hypothesis
  Transfer for Unsupervised Domain Adaptation}.
\newblock In {\em Proc. of ICML}, pages 6028--6039, Virtual Conference, July
  2020.

\bibitem{lin2014microsoft}
Tsung-Yi Lin, Michael Maire, Serge Belongie, James Hays, Pietro Perona, Deva
  Ramanan, Piotr Doll{\'a}r, and C~Lawrence Zitnick.
\newblock {Microsoft COCO: Common Objects in Context}.
\newblock In {\em Proc. of ECCV}, pages 740--755, Zurich, Switzerland, Sept.
  2014.

\bibitem{Liu2021}
Yuang Liu, Wei Zhang, and Jun Wang.
\newblock {Source-Free Domain Adaptation for Semantic Segmentation}.
\newblock In {\em Proc. of CVPR}, pages 1215--1224, Virtual, June 2021.

\bibitem{Mancini2018}
Massimiliano Mancini, Lorenzo Porzi, Samuel~Rota Bulò, Barbara Caputo, and
  Elisa Ricci.
\newblock {Boosting Domain Adaptation by Discovering Latent Domains}.
\newblock In {\em Proc. of CVPR}, pages 3771--3780, Salt Lake City, UT, USA,
  June 2018.

\bibitem{Carlucci2017}
Fabio Maria~Carlucci, Lorenzo Porzi, Barbara Caputo, Elisa Ricci, and Samuel
  Rota~Bulo.
\newblock {AutoDIAL: Automatic DomaIn Alignment Layers}.
\newblock In {\em Proc. of ICCV}, pages 5067--5075, Venice, Italy, Oct. 2017.

\bibitem{Mei2020}
Ke Mei, Chuang Zhu, Jiaqi Zou, and Shanghang Zhang.
\newblock {Instance Adaptive Self-Training for Unsupervised Domain Adaptation}.
\newblock In {\em Proc. of ECCV}, pages 415--430, Glasgow, UK, Aug. 2020.

\bibitem{Menze2015}
Moritz Menze and Andreas Geiger.
\newblock {Object Scene Flow for Autonomous Vehicles}.
\newblock In {\em Proc. of CVPR}, pages 3061--3070, Boston, MA, USA, June 2015.

\bibitem{Miyato2018}
Takeru Miyato, Toshiki Kataoka, Masanori Koyama, and Yuichi Yoshida.
\newblock {Spectral Normalization for Generative Adversarial Networks}.
\newblock In {\em {Proc. of ICLR}}, pages 1--12, Vancouver, BC, Canada, Apr.
  2018.

\bibitem{Nam2018}
Hyeonseob Nam and Hyo-Eun Kim.
\newblock {Batch-Instance Normalization for Adaptively Style-Invariant Neural
  Network}.
\newblock In {\em Proc. of NIPS}, pages 2558--2567, Montr\' {e}al, QC, Canada,
  Dec. 2018.

\bibitem{Neuhold2017}
Gerhard Neuhold, Tobias Ollmann, Samuel~Rota Bulò, and Peter Kontschieder.
\newblock {The Mapillary Vistas Dataset for Semantic Understanding of Street
  Scenes}.
\newblock In {\em Proc. of ICCV}, pages 4990--4999, Venice, Italy, Oct. 2017.

\bibitem{Pan2018}
Xingang Pan, Ping Luo, Jianping Shi, and Xiaoou Tang.
\newblock {Two at Once: Enhancing Learning and Generalization Capacities via
  IBN-Net}.
\newblock In {\em Proc. of ECCV}, pages 464--479, Munich, Germany, Sept. 2018.

\bibitem{Paszke2019}
Adam Paszke, Sam Gross, Francisco Massa, Adam Lerer, James Bradbury, Gregory
  Chanan, Trevor Killeen, Zeming Lin, Natalia Gimelshein, Luca Antiga, Alban
  Desmaison, et~al.
\newblock {PyTorch: An Imperative Style, High-Performance Deep Learn\-ing
  Library}.
\newblock In {\em Proc. of NeurIPS}, pages 8024--8035, Vancouver, BC, Canada,
  Dec. 2019.

\bibitem{RaviKumar2021a}
Varun Ravi~Kumar, Marvin Klingner, Senthil Yogamani, Markus Bach, Stefan Milz,
  Tim Fingscheidt, and Patrick M\"{a}der.
\newblock {SVDistNet: Self-Supervised Near-Field Distance Estimation on
  Surround View Fisheye Cameras}.
\newblock {\em IEEE Transactions on Intelligent Transportation Systems}, pages
  1--10, 2021.

\bibitem{RaviKumar2021}
Varun Ravi~Kumar, Marvin Klingner, Senthil Yogamani, Stefan Milz, Tim
  Fingscheidt, and Patrick M\"{a}der.
\newblock {SynDistNet: Self-Supervised Monocular Fisheye Camera Distance
  Estimation Synergized with Semantic Segmentation for Autonomous Driving}.
\newblock In {\em Proc. of WACV}, pages 61--71, Waikoloa, HI, USA, Jan. 2021.

\bibitem{Richter2016}
Stephan Richter, Vibhav Vineet, Stefan Roth, and Vladlen Koltun.
\newblock {Playing for Data: Ground Truth from Computer Games}.
\newblock In {\em Proc. of ECCV}, pages 102--118, Amsterdam, Netherlands, Oct.
  2016.

\bibitem{Romijnders2019}
Rob Romijnders, Panagiotis Meletis, and Gijs Dubbelman.
\newblock {A Domain Agnostic Normalization Layer for Unsupervised Adversarial
  Domain Adaptation}.
\newblock In {\em Proc. of WACV}, pages 1866--1875, Waikoloa, Hawaii, Jan.
  2019.

\bibitem{Ros2016}
German Ros, Laura Sellart, Joanna Materzynska, David Vazquez, and Antonio~M.
  Lopez.
\newblock {The SYNTHIA Dataset: A Large Collection of Synthetic Images for
  Semantic Segmentation of Urban Scenes}.
\newblock In {\em Proc. of CVPR}, pages 3234--3243, Las Vegas, NV, USA, June
  2016.

\bibitem{Russakovsky2015}
Olga Russakovsky, Jia Deng, Hao Su, Jonathan Krause, Sanjeev Satheesh, Sean Ma,
  Zhiheng Huang, Andrej Karpathy, Aditya Khosla, Michael Bernstein,
  Alexander~C. Berg, and Li Fei-Fei.
\newblock {ImageNet Large Scale Visual Recognition Challenge}.
\newblock {\em International Journal of Computer Vision (IJCV)},
  115(3):211--252, Dec. 2015.

\bibitem{Schneider2020}
Steffen Schneider, Evgenia Rusak, Luisa Eck, Oliver Bringmann, Wieland Brendel,
  and Matthias Bethge.
\newblock {Improving Robustness Against Common Corruptions by Covariate Shift
  Adaptation}.
\newblock In {\em Proc. of NeurIPS}, pages 1--13, Virtual Conference, Dec.
  2020.

\bibitem{Seo2020}
Seonguk Seo, Yumin Suh, Dongwan Kim, Geeho Kim, Jongwoo Han, and Bohyung Han.
\newblock {Learning to Optimize Domain Specific Normalization for Domain
  Generalization}.
\newblock In {\em Proc. of ECCV}, pages 68--83, Glasgow, UK, Aug. 2020.

\bibitem{Shao2019}
Wenqi Shao, Tianjian Meng, Jingyu Li, Ruimao Zhang, Yudian Li, Xiaogang Wang,
  and Ping Luo.
\newblock {SSN: Learning Sparse Switchable Normalization via SparsestMax}.
\newblock In {\em {Proc. of CVPR}}, pages 443--451, Long Beach, CA, USA, June
  2019.

\bibitem{Simonyan2015}
Karen Simonyan and Andrew Zisserman.
\newblock {Very Deep Convolutional Networks for Large-Scale Image Recognition}.
\newblock In {\em Proc. of ICLR}, pages 1--27, San Diego, CA, USA, May 2015.

\bibitem{Subhani2020}
Muhammad~Naseer Subhani and Mohsen Ali.
\newblock {Learning from Scale-Invariant Examples for Domain Adaptation in
  Semantic Segmentation}.
\newblock In {\em Proc. of ECCV}, pages 290--306, Glasgow, UK, Aug. 2020.

\bibitem{Teja2021}
Prabhu Teja~S and Francois Fleuret.
\newblock {Uncertainty Reduction for Model Adaptation in Semantic
  Segmentation}.
\newblock In {\em Proc. of CVPR}, pages 9613--9623, Virtual, June 2021.

\bibitem{Termoehlen2021}
Jan-Aike Term\"{o}hlen, Marvin Klingner, Leon~J. Brettin, Nico~M. Schmidt, and
  Tim Fingscheidt.
\newblock {Continual Unsupervised Domain Adaptation for Semantic Segmentation
  by Online Frequency Domain Style Transfer}.
\newblock In {\em Proc. of ITSC}, pages 2881--2888, Virtual, Sept. 2021.

\bibitem{Tranheden2021}
Wilhelm Tranheden, Viktor Olsson, Juliano Pinto, and Lennart Svensson.
\newblock {DACS: Domain Adaptation via Cross-Domain Mixed Sampling}.
\newblock In {\em Proc. of WACV}, pages 1379--1389, Waikoloa, HI, USA, Jan.
  2021.

\bibitem{Tsai2018}
Yi-Hsuan Tsai, Wei-Chih Hung, Samuel Schulter, Kihyuk Sohn, Ming-Hsuan Yang,
  and Manmohan Chandraker.
\newblock {Learning to Adapt Structured Output Space for Semantic
  Segmentation}.
\newblock In {\em Proc. of CVPR}, pages 7472--7481, Salt Lake City, UT, USA,
  June 2018.

\bibitem{Ulyanov2016}
Dmitry Ulyanov, Andrea Vedaldi, and Victor Lempitsky.
\newblock {Instance Normalization: The Missing Ingredient for Fast
  Stylization}.
\newblock {\em arXiv}, July 2016.
\newblock (1607.08022).

\bibitem{Vu2019}
Tuan-Hung Vu, Himalaya Jain, Maxime Bucher, Matthieu Cord, and Patrick Perez.
\newblock {ADVENT: Adversarial Entropy Minimization for Domain Adaptation in
  Semantic Segmentation}.
\newblock In {\em Proc. of CVPR}, pages 2517--2526, Long Beach, CA, USA, June
  2019.

\bibitem{Vu2019a}
Tuan-Hung Vu, Himalaya Jain, Maxime Bucher, Matthieu Cord, and Patrick
  P\'{e}rez.
\newblock {DADA: Depth-Aware Domain Adaptation in Semantic Segmentation}.
\newblock In {\em Proc. of ICCV}, pages 7364--7373, Seoul, Korea, Oct. 2019.

\bibitem{Wang2019d}
Ximei Wang, Ying Jin, Mingsheng Long, Jianmin Wang, and Michael~I. Jordan.
\newblock {Transferable Normalization: Towards Improving Transferability of
  Deep Neural Networks}.
\newblock In {\em Proc. of NIPS}, pages 1953--1963, Vancouver, BC, Canada, Dec.
  2019.

\bibitem{Wang2020}
Zhonghao Wang, Mo Yu, Yunchao Wei, Rogerio Feris, Jinjun Xiong, {Wen-Mei} Hwu,
  Thomas~S. Huang, and Honghui Shi.
\newblock {Differential Treatment for Stuff and Things: A Simple Unsupervised
  Domain Adaptation Method for Semantic Segmentation}.
\newblock In {\em Proc. of CVPR}, pages 12635--12644, Seattle, WA, USA, June
  2020.

\bibitem{Wu2018}
Yuxin Wu and Kaiming He.
\newblock {Group Normalization}.
\newblock In {\em Proc. of ECCV}, pages 3--19, Munich, Germany, Sept. 2018.

\bibitem{Wulfmeier2018}
Markus Wulfmeier, Alex Bewley, and Ingmar Posner.
\newblock {Incremental Adversarial Domain Adaptation for Continually Changing
  Environments}.
\newblock In {\em Proc. of ICRA}, pages 4489--4495, Brisbane, Australia, May
  2018.

\bibitem{Xu2019b}
Jiaolong Xu, Liang Xiao, and Antonio~M. López.
\newblock {Self-Supervised Domain Adaptation for Computer Vision Tasks}.
\newblock {\em IEEE Access}, 7:156694--156706, 2019.

\bibitem{Yang2020a}
Jinyu Yang, Weizhi An, Sheng Wang, Xinliang Zhu, Chaochao Yan, and Junzhou
  Huang.
\newblock {Label-Driven Reconstruction for Domain Adaptation in Semantic
  Segmentation}.
\newblock In {\em Proc. of ECCV}, pages 480--498, Glasgow, UK, Aug. 2020.

\bibitem{Yang2020}
Yanchao Yang and Stefano Soatto.
\newblock {FDA: Fourier Domain Adaptation for Semantic Segmentation}.
\newblock In {\em Proc. of CVPR}, pages 4085--4095, Seattle, WA, USA, June
  2020.

\bibitem{Yeh2021}
Hao-Wei Yeh, Baoyao Yang, Pong~C. Yuen, and Tatsuya Harada.
\newblock {SoFA: Source-Data-Free Feature Alignment for Unsupervised Domain
  Adaptation}.
\newblock In {\em Proc. of WACV}, pages 474--483, Virtual Conference, Jan.
  2021.

\bibitem{Yue2019}
Xiangyu Yue, Yang Zhang, Sicheng Zhao, Alberto Sangiovanni-Vincentelli, Kurt
  Keutzer, and Boqing Gong.
\newblock {Domain Randomization and Pyramid Consistency: Simulation-to-Real
  Generalization Without Accessing Target Domain Data}.
\newblock In {\em Proc. of ICCV}, pages 2100--2110, Seoul, Korea, Oct. 2019.

\bibitem{Zhang2020a}
Jian Zhang, Lei Qi, Yinghuan Shi, and Yang Gao.
\newblock {Generalizable Semantic Segmentation via Model-Agnostic Learning and
  Target-Specific Normalization}.
\newblock {\em arXiv}, (2003.12296), May 2020.

\bibitem{Zhang2019e}
Qiming Zhang, Jing Zhang, Wei Liu, and Dacheng Tao.
\newblock {Category Anchor-Guided Unsupervised Domain Adaptation for Semantic
  Segmentation}.
\newblock In {\em Proc. of NeurIPS}, pages 433--443, Vancouver, Canada, Dec.
  2019.

\bibitem{Zou2018}
Yang Zou, Zhiding Yu, B.~V.~K. Vijaya~Kumar, and Jinsong Wang.
\newblock {Unsupervised Domain Adaptation for Semantic Segmentation via
  Class-Balanced Self-Training}.
\newblock In {\em Proc. of ECCV}, pages 289--305, Munich, Germany, Sept. 2018.

\end{thebibliography}
}
\clearpage
 
\appendix

\section{Additional Experimental Evaluation}
\label{sec:quantitative_sup}

With this section our main aim is to provide a more complete overview on the results of our method. This in particular means that we provide a hyperparameter analysis, show our conducted experiments on other source/target domain combinations, provide results for more network architectures, and give a deeper analysis on the variance of different experiments with the same hyperparameter setting but different random seeds. Note that most subsection names are equal to the ones in Sec.~\textcolor{red}{6} of the main paper, and thereby correspond to these respective sections.

\subsection{Hyperparameter Analysis}

\begin{figure}[t]
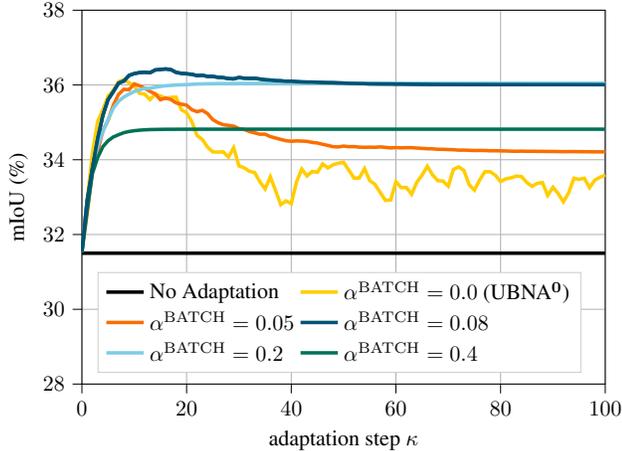

	\centering	
	\includestandalone[width=1.0\linewidth]{figs/ablation/momentum_decay_gta_to_cs}
	\caption{\textbf{Hyperparameter analysis}: Performance of \textbf{UBNA} on the \textbf{Cityscapes validation set} for the adaptation from \textbf{GTA-5} ($\mathcal{D}^{\mathrm{S}}$) \textbf{to Cityscapes} ($\mathcal{D}^{\mathrm{T}}$) in \textbf{dependence of the batch decay factor $\alpha^{\mathrm{BATCH}}$}. Here, $\alpha^{\mathrm{LAYER}}= 0.0$ for all UBNA experiments.}
	\label{fig:momentum_decay_gta_to_cs}
\end{figure} 

\begin{figure}[t]
	\centering	
	\includestandalone[width=1.0\linewidth]{figs/ablation/momentum_decay_synthia_to_cs}
	\caption{\textbf{Hyperparameter analysis}: Performance of \textbf{UBNA} on the \textbf{Cityscapes validation set} for the adaptation from \textbf{SYNTHIA} ($\mathcal{D}^{\mathrm{S}}$) \textbf{to Cityscapes} ($\mathcal{D}^{\mathrm{T}}$) in \textbf{dependence of the batch decay factor $\alpha^{\mathrm{BATCH}}$}. Here, $\alpha^{\mathrm{LAYER}}= 0.0$ for all UBNA experiments.}
	\label{fig:momentum_decay_synthia_to_cs}
\end{figure} 

\begin{figure}[t]
	\centering	
	\includestandalone[width=1.0\linewidth]{figs/ablation/momentum_decay_cs_to_kitti}
	\caption{\textbf{Hyperparameter analysis}: Performance of \textbf{UBNA} on the \textbf{KITTI validation set} for the adaptation from \textbf{Cityscapes} ($\mathcal{D}^{\mathrm{S}}$) \textbf{to KITTI} ($\mathcal{D}^{\mathrm{T}}$) in \textbf{dependence of the batch decay factor $\alpha^{\mathrm{BATCH}}$}. Here, $\alpha^{\mathrm{LAYER}}= 0.0$ for all UBNA experiments.}
	\label{fig:momentum_decay_cs_to_kitti}
\end{figure} 

\begin{figure}[t]
	\centering	
	\includestandalone[width=1.0\linewidth]{figs/ablation/layer_decay_gta_to_cs}
	\caption{\textbf{Hyperparameter analysis}: Performance of \textbf{UBNA}$^{\myplus}$ on the \textbf{Cityscapes validation set} for the adaptation from \textbf{GTA-5} ($\mathcal{D}^{\mathrm{S}}$) \textbf{to Cityscapes} ($\mathcal{D}^{\mathrm{T}}$) in \textbf{dependence of the layer decay factor $\alpha^{\mathrm{LAYER}}$}. Here, $\alpha^{\mathrm{BATCH}}= 0.08$ for all \textbf{UBNA}$^{\myplus}$ experiments.}
	\label{fig:layer_decay_gta_to_cs}
\end{figure} 

\begin{figure}[t]
	\centering	
	\includestandalone[width=1.0\linewidth]{figs/ablation/layer_decay_synthia_to_cs}
	\caption{\textbf{Hyperparameter analysis}: Performance of \textbf{UBNA}$^{\myplus}$ on the \textbf{Cityscapes validation set} for the adaptation from \textbf{SYNTHIA} ($\mathcal{D}^{\mathrm{S}}$) \textbf{to Cityscapes} ($\mathcal{D}^{\mathrm{T}}$) in \textbf{dependence of the layer decay factor $\alpha^{\mathrm{LAYER}}$}. Here, $\alpha^{\mathrm{BATCH}}= 0.08$ for all \textbf{UBNA}$^{\myplus}$ experiments.}
	\label{fig:layer_decay_synthia_to_cs}
\end{figure} 

\begin{figure}[t]
	\centering	
	\includestandalone[width=1.0\linewidth]{figs/ablation/layer_decay_cs_to_kitti}
	\caption{\textbf{Hyperparameter analysis}: Performance of \textbf{UBNA}$^{\myplus}$ on the \textbf{KITTI validation set} for the adaptation from \textbf{Cityscapes} ($\mathcal{D}^{\mathrm{S}}$) \textbf{to KITTI} ($\mathcal{D}^{\mathrm{T}}$) in \textbf{dependence of the layer decay factor $\alpha^{\mathrm{LAYER}}$}. Here, $\alpha^{\mathrm{BATCH}}= 0.08$ for all \textbf{UBNA}$^{\myplus}$ experiments.}
	\label{fig:layer_decay_cs_to_kitti}
\end{figure} 

To give a further insight into the hyperparameter selection for our UBNA method, we show for all three considered dataset settings in Figs.~\ref{fig:momentum_decay_gta_to_cs}, \ref{fig:momentum_decay_synthia_to_cs}, and \ref{fig:momentum_decay_cs_to_kitti} the performance (ordinate) in dependence of the number of adaptation steps (abscissa). Here, we can already see that the adaptation consistently converges after $\kappa=50$ adaptation steps and a further adaptation does not improve the performance anymore. Note, that Figs.~\ref{fig:momentum_decay_gta_to_cs}, \ref{fig:momentum_decay_synthia_to_cs}, and \ref{fig:momentum_decay_cs_to_kitti} show the adaptation until $\kappa=100$ adaptation steps, though we do not observe any change in performance after $\kappa=50$ adaptation steps. For $\alpha^{\mathrm{BATCH}}=0.08$, using less adaptation steps could potentially still improve the results (cf.~Fig.~\ref{fig:momentum_decay_gta_to_cs}), however, this optimal point does not generalize well across different datasets (cf.~Figs.~\ref{fig:momentum_decay_synthia_to_cs} and \ref{fig:baseline_comparison_cs_to_kitti}), which is why we use  $\kappa=50$ adaptation steps in the main paper, where we observe a stable convergence.
\par
We also show the influence of using different batch-wise decay factors $\alpha^{\mathrm{BATCH}}$. Interestingly, we observe that adapting the statistics to the target domain is beneficial regardless of the used value for $\alpha^{\mathrm{BATCH}}$. However, the convergence is quite optimal for values around $\alpha^{\mathrm{BATCH}}=0.08$. For GTA-5 to Cityscapes (Fig.~\ref{fig:momentum_decay_gta_to_cs}) and SYNTHIA to Cityscapes (Fig.~\ref{fig:momentum_decay_synthia_to_cs}), this yields the optimal performance, and for the Cityscapes to KITTI adaptation only a value of $\alpha^{\mathrm{BATCH}}=0.2$ yields a slightly better performance. Much smaller values of $\alpha^{\mathrm{BATCH}}$ lead to an unstable convergence and a lower performance gain (as observed in UBNA$^0$ with $\alpha^{\mathrm{BATCH}}=0$, yellow curves in Figs.~\ref{fig:momentum_decay_gta_to_cs}, \ref{fig:momentum_decay_synthia_to_cs}, and \ref{fig:momentum_decay_cs_to_kitti}). On the other hand the more rapidly decreasing BN momentum for larger values of $\alpha^{\mathrm{BATCH}}$ seem to stop the adaptation too rapidly and lead to a convergence much below the optimal point. Still, the approximate optimal hyperparameter values of $\alpha^{\mathrm{BATCH}}=0.08$ and $\kappa=50$ generalize well across different dataset settings, which is essential for practical applications of our UBNA method.
\par
Furthermore, we introduced a layer-wise weighting factor $\alpha^{\mathrm{LAYER}}$, which causes the statistics in the initial layers to be updated more rapidly than in the deeper layers. We show the performance for different values of $\alpha^{\mathrm{LAYER}}$ in Figs.~\ref{fig:layer_decay_gta_to_cs}, \ref{fig:layer_decay_synthia_to_cs}, and \ref{fig:layer_decay_cs_to_kitti}. We observe that for a suitable value of $\alpha^{\mathrm{LAYER}}$ this weighting again improves on top of the UBNA method ($\alpha^{\mathrm{LAYER}}=0$, yellow curves in Figs.~\ref{fig:layer_decay_gta_to_cs}, \ref{fig:layer_decay_synthia_to_cs}, and \ref{fig:layer_decay_cs_to_kitti}), resulting in the UBNA$^{+}$ approach. This could hint at the fact that the domain gap can be compensated to a large extent in the initial layers, which extract domain-specific knowledge, while the deeper layers already learned more domain-invariant features that are rather task-specific. The optimal $\alpha^{\mathrm{LAYER}}$ value of this method is, however, dataset-dependent. On the synthetic-to-real settings GTA-5 to Cityscapes (Fig.~\ref{fig:layer_decay_gta_to_cs}) and SYNTHIA to Cityscapes (Fig.~\ref{fig:layer_decay_synthia_to_cs}) we found an optimal value of $\alpha^{\mathrm{LAYER}}=0.03$, while for the real-to-real setting Cityscapes to KITTI (Fig.~\ref{fig:layer_decay_cs_to_kitti}) the optimal value was found to be $\alpha^{\mathrm{LAYER}}=0.3$. We suspect that due to the larger domain gap in the synthetic-to-real settings an adaptation only in the initial layers is not sufficient, thereby requiring a smaller value of $\alpha^{\mathrm{LAYER}}$. Conclusively, if one has access to a labelled validation set in the target domain, one can tune this hyperparameter for better performance.
\par

\subsection{Comparison to Normalization Approaches}

\begin{figure}[t]
	\centering	
\begin{tikzpicture}

\definecolor{color0}{rgb}{0.690196078431373,0,0.274509803921569}

\begin{axis}[
height=6.6cm,
legend cell align={left},
legend columns=3,
legend style={at={(0.03,0.03)}, anchor=south west, draw=white!80.0!black},
tick align=outside,
tick pos=left,
width=10cm,
x grid style={white!69.01960784313725!black},
xlabel={adaptation step \(\displaystyle \kappa\)},
xmajorgrids,
xmin=0, xmax=50,
xtick style={color=black},
y grid style={white!69.01960784313725!black},
ylabel={mIoU (\%)},
ymajorgrids,
ymin=29, ymax=37,
ytick style={color=black}
]
\addplot [ultra thick, black]
table {%
0 34.7
1 34.7
2 34.7
3 34.7
4 34.7
5 34.7
6 34.7
7 34.7
8 34.7
9 34.7
10 34.7
11 34.7
12 34.7
13 34.7
14 34.7
15 34.7
16 34.7
17 34.7
18 34.7
19 34.7
20 34.7
21 34.7
22 34.7
23 34.7
24 34.7
25 34.7
26 34.7
27 34.7
28 34.7
29 34.7
30 34.7
31 34.7
32 34.7
33 34.7
34 34.7
35 34.7
36 34.7
37 34.7
38 34.7
39 34.7
40 34.7
41 34.7
42 34.7
43 34.7
44 34.7
45 34.7
46 34.7
47 34.7
48 34.7
49 34.7
50 34.7
51 34.7
52 34.7
53 34.7
54 34.7
55 34.7
56 34.7
57 34.7
58 34.7
59 34.7
60 34.7
61 34.7
62 34.7
63 34.7
64 34.7
65 34.7
66 34.7
67 34.7
68 34.7
69 34.7
70 34.7
71 34.7
72 34.7
73 34.7
74 34.7
75 34.7
76 34.7
77 34.7
78 34.7
79 34.7
80 34.7
81 34.7
82 34.7
83 34.7
84 34.7
85 34.7
86 34.7
87 34.7
88 34.7
89 34.7
90 34.7
91 34.7
92 34.7
93 34.7
94 34.7
95 34.7
96 34.7
97 34.7
98 34.7
99 34.7
100 34.7
};
\addlegendentry{AdaBN}
\addplot [ultra thick, color0]
table {%
0 31.5373507411942
1 32.9079289885792
2 33.7669883441508
3 34.5538180361038
4 35.1839384072919
5 35.6100305118696
6 35.8215381777551
7 36.0478056826121
8 36.1046777938189
9 36.2546679738796
10 36.3045437701834
11 36.3306300536483
12 36.3400315647132
13 36.3268121377049
14 36.4015487910366
15 36.4153583556121
16 36.4282996509206
17 36.4106600358601
18 36.3359088742694
19 36.3109039751598
20 36.2991809355418
21 36.2853274400606
22 36.2505309923884
23 36.2590907572009
24 36.2293921353937
25 36.2286587788101
26 36.2056052396842
27 36.1961864512043
28 36.170168022008
29 36.1631212959528
30 36.1994418243896
31 36.1808735629596
32 36.1763781670125
33 36.177803557057
34 36.1593523250193
35 36.1387197399415
36 36.1311024027072
37 36.1146444789785
38 36.1111660559288
39 36.0981832657391
40 36.098594061566
41 36.0880166832975
42 36.0809884413867
43 36.077246258712
44 36.0773896688194
45 36.0695154714209
46 36.0643986420076
47 36.0605232130655
48 36.0595820079545
49 36.0542899137091
50 36.0510432580443
51 36.0441332326807
52 36.0408264004301
53 36.0376729854033
54 36.0347206258813
55 36.0292679046725
56 36.0288271306604
57 36.0272054939275
58 36.0266737887308
59 36.0260142785501
60 36.0240213299301
61 36.0240856680857
62 36.0221637272372
63 36.0208125907548
64 36.0189333358582
65 36.017619435275
66 36.016763182349
67 36.0155285776465
68 36.0145792150419
69 36.0139141003395
70 36.0138730911276
71 36.013502211754
72 36.0121971478996
73 36.0123410188013
74 36.0119813702186
75 36.0123599993771
76 36.0121729129298
77 36.0116954419088
78 36.0110541981405
79 36.0113227367778
80 36.0107103841193
81 36.0105272731853
82 36.01039728646
83 36.0102379545
84 36.0102297070343
85 36.0098737347226
86 36.0095773753953
87 36.009501887701
88 36.0093549021976
89 36.0093135884476
90 36.0091454125851
91 36.0090135731807
92 36.0089428633239
93 36.0086842823637
94 36.008564253389
95 36.0085012472693
96 36.008375382271
97 36.008231953746
98 36.0081426128835
99 36.0080449044361
100 36.0080394314754
};
\addlegendentry{\textbf{UBNA}}
\addplot [ultra thick, color0, mark=+, mark size=2, mark options={solid}]
table {%
0 31.5373507411942
1 32.6679711139742
2 33.5322462594423
3 34.2845772475272
4 34.7866159802578
5 35.2176164510253
6 35.5638523041265
7 35.800753979942
8 35.9504640546506
9 36.0520819971222
10 36.2152550985327
11 36.3501037542473
12 36.3743815850227
13 36.4226875930092
14 36.4392871886751
15 36.4448130932851
16 36.438141215708
17 36.4750216636536
18 36.5559872541902
19 36.5407004775213
20 36.5639279788241
21 36.5431249252929
22 36.5589364805322
23 36.5473764635685
24 36.5233142949429
25 36.507087851491
26 36.4886109888269
27 36.4906827175178
28 36.4841108317231
29 36.4795846395047
30 36.4773554654652
31 36.4942196954875
32 36.4929283527226
33 36.5046689127448
34 36.5175469805107
35 36.5072960253437
36 36.5101615072433
37 36.5012575554593
38 36.4907146453508
39 36.4927430054703
40 36.485988787491
41 36.4900346320229
42 36.4927762136645
43 36.4827607972736
44 36.4809144625533
45 36.4779239680108
46 36.4754593516548
47 36.4717682494855
48 36.4707430208836
49 36.4651766866876
50 36.4648376225235
51 36.4635722933919
52 36.461867524523
53 36.4613845704587
54 36.4593557183099
55 36.4585872529155
56 36.4580794167813
57 36.4553878393038
58 36.4546957884116
59 36.4546713252157
60 36.4541071158109
61 36.4531471558769
62 36.4524572131117
63 36.4515056931314
64 36.4508946322127
65 36.4516923224798
66 36.4518622970621
67 36.4512439680111
68 36.4509621441574
69 36.4508595279964
70 36.4503700616918
71 36.4497014033814
72 36.4495275020895
73 36.4493474924766
74 36.4491682999721
75 36.4487573174534
76 36.4485825071037
77 36.4489839239007
78 36.4488437073869
79 36.4483027121853
80 36.4483438617124
81 36.4482883169428
82 36.4484417051854
83 36.448334156802
84 36.4480700224277
85 36.4478773032583
86 36.4478750604803
87 36.4477060178312
88 36.447689619722
89 36.4477403001625
90 36.4477802857872
91 36.4477857201988
92 36.447811676843
93 36.4477126658382
94 36.4476344330748
95 36.4475562332487
96 36.4474698069198
97 36.4475630865739
98 36.4475824017602
99 36.4475771220242
100 36.4475901136734
};
\addlegendentry{\textbf{UBNA}$^\mathbf{+}$}
\addplot [ultra thick, black, dashed]
table {%
0 31.5
1 31.5
2 31.5
3 31.5
4 31.5
5 31.5
6 31.5
7 31.5
8 31.5
9 31.5
10 31.5
11 31.5
12 31.5
13 31.5
14 31.5
15 31.5
16 31.5
17 31.5
18 31.5
19 31.5
20 31.5
21 31.5
22 31.5
23 31.5
24 31.5
25 31.5
26 31.5
27 31.5
28 31.5
29 31.5
30 31.5
31 31.5
32 31.5
33 31.5
34 31.5
35 31.5
36 31.5
37 31.5
38 31.5
39 31.5
40 31.5
41 31.5
42 31.5
43 31.5
44 31.5
45 31.5
46 31.5
47 31.5
48 31.5
49 31.5
50 31.5
51 31.5
52 31.5
53 31.5
54 31.5
55 31.5
56 31.5
57 31.5
58 31.5
59 31.5
60 31.5
61 31.5
62 31.5
63 31.5
64 31.5
65 31.5
66 31.5
67 31.5
68 31.5
69 31.5
70 31.5
71 31.5
72 31.5
73 31.5
74 31.5
75 31.5
76 31.5
77 31.5
78 31.5
79 31.5
80 31.5
81 31.5
82 31.5
83 31.5
84 31.5
85 31.5
86 31.5
87 31.5
88 31.5
89 31.5
90 31.5
91 31.5
92 31.5
93 31.5
94 31.5
95 31.5
96 31.5
97 31.5
98 31.5
99 31.5
100 31.5
};
\addlegendentry{No Adaptation}
\addplot [ultra thick, color0, dashed]
table {%
0 31.5373507411942
1 32.8999090086923
2 34.1842817470743
3 35.0221449238558
4 35.3412027940937
5 35.7024937737477
6 35.8566244098308
7 36.08886763526
8 36.1393643085663
9 36.0590424630087
10 35.9727254067122
11 35.7742580289977
12 35.7557433558609
13 35.6081280060771
14 35.7578924968844
15 35.7085318296109
16 35.6378021303091
17 35.6670391462953
18 35.67360052381
19 35.3355568637425
20 35.2581227436284
21 35.0195625648776
22 34.6487321423972
23 34.5457622798508
24 34.3394717069808
25 34.2114482733921
26 34.2034019880969
27 34.2124353545125
28 34.0671966340247
29 34.3219969413537
30 33.836668329837
31 33.7836465432059
32 33.6904761580928
33 33.6586349838753
34 33.412689638463
35 33.219210429251
36 33.5449708867269
37 33.1755846200087
38 32.7978154421069
39 32.8995591096881
40 32.8228375020176
41 33.4038904567363
42 33.7516788301293
43 33.5663636291636
44 33.6865782703644
45 33.6655244571536
46 33.8880546837468
47 33.7813193829249
48 33.8674684403654
49 33.9060764422175
50 33.9319012921476
51 33.7220680889702
52 33.5047278968989
53 33.5108592208854
54 33.7613219112387
55 33.4418784346593
56 33.3413993311175
57 33.1225442595527
58 32.9118369766315
59 32.9242026002458
60 33.4095039707824
61 33.2553131782255
62 33.2596291505624
63 33.0776895636229
64 33.3995567833158
65 33.5886610246352
66 33.8260370507415
67 33.5962519188353
68 33.4388968851862
69 33.4309392941301
70 33.2917763655339
71 33.7421277691257
72 33.8086551857773
73 33.7116398975817
74 33.3294777068035
75 33.185071645592
76 33.4592633641586
77 33.680214617262
78 33.6798231978943
79 33.5858963766125
80 33.4147336982325
81 33.4913324910518
82 33.5154202758235
83 33.5628409906162
84 33.5497395345669
85 33.3943570555452
86 33.2883307108595
87 33.3680976961026
88 33.3888190839697
89 33.0466709555732
90 33.2647017318093
91 33.0973642670195
92 32.8747897112697
93 33.1118113399272
94 33.5045602290004
95 33.4752196577484
96 33.5043430328657
97 33.2555393944082
98 33.4533233385871
99 33.5334359662065
100 33.5859508642449
};
\addlegendentry{\textbf{UBNA}$^\mathbf{0}$}
\end{axis}

\end{tikzpicture}
	\put(-152.5,48){\footnotesize \cite{Li2018d}}
	\caption{\textbf{Comparison to normalization approaches}: Performance on the \textbf{Cityscapes validation set} for the adaptation from \textbf{GTA-5} ($\mathcal{D}^{\mathrm{S}}$) \textbf{to Cityscapes} ($\mathcal{D}^{\mathrm{T}}$) in \textbf{dependence of the adaptation step} $\kappa$ (batch size $B=6$). We also show results when using the statistics from the source domain (no adaptation) and from the target domain (AdaBN by Li~\etal~\cite{Li2018d}).}
	\label{fig:baseline_comparison_gta_to_cs}
\end{figure}
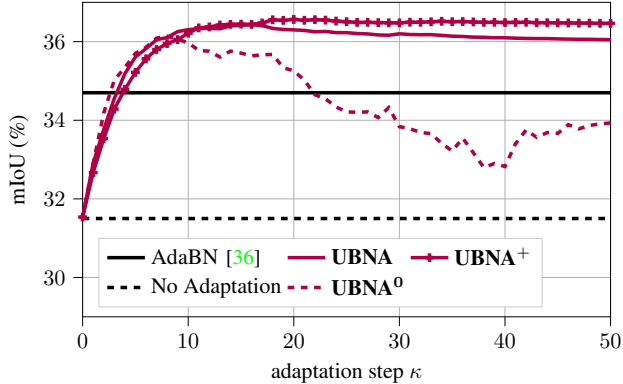

\begin{figure}[t]
	\centering	
\begin{tikzpicture}

\definecolor{color0}{rgb}{0,0.325490196078431,0.454901960784314}

\begin{axis}[
height=6.6cm,
legend cell align={left},
legend columns=3,
legend style={at={(0.03,0.03)}, anchor=south west, draw=white!80.0!black},
tick align=outside,
tick pos=left,
width=10cm,
x grid style={white!69.01960784313725!black},
xlabel={adaptation step \(\displaystyle \kappa\)},
xmajorgrids,
xmin=0, xmax=50,
xtick style={color=black},
y grid style={white!69.01960784313725!black},
ylabel={mIoU (\%)},
ymajorgrids,
ymin=47, ymax=60,
ytick style={color=black}
]
\addplot [ultra thick, black]
table {%
0 55.7
1 55.7
2 55.7
3 55.7
4 55.7
5 55.7
6 55.7
7 55.7
8 55.7
9 55.7
10 55.7
11 55.7
12 55.7
13 55.7
14 55.7
15 55.7
16 55.7
17 55.7
18 55.7
19 55.7
20 55.7
21 55.7
22 55.7
23 55.7
24 55.7
25 55.7
26 55.7
27 55.7
28 55.7
29 55.7
30 55.7
31 55.7
32 55.7
33 55.7
34 55.7
35 55.7
36 55.7
37 55.7
38 55.7
39 55.7
40 55.7
41 55.7
42 55.7
43 55.7
44 55.7
45 55.7
46 55.7
47 55.7
48 55.7
49 55.7
50 55.7
51 55.7
52 55.7
53 55.7
54 55.7
55 55.7
56 55.7
57 55.7
58 55.7
59 55.7
60 55.7
61 55.7
62 55.7
63 55.7
64 55.7
65 55.7
66 55.7
67 55.7
68 55.7
69 55.7
70 55.7
71 55.7
72 55.7
73 55.7
74 55.7
75 55.7
76 55.7
77 55.7
78 55.7
79 55.7
80 55.7
81 55.7
82 55.7
83 55.7
84 55.7
85 55.7
86 55.7
87 55.7
88 55.7
89 55.7
90 55.7
91 55.7
92 55.7
93 55.7
94 55.7
95 55.7
96 55.7
97 55.7
98 55.7
99 55.7
100 55.7
};
\addlegendentry{AdaBN}
\addplot [ultra thick, color0]
table {%
0 51.0734520751163
1 54.5847741691449
2 55.8617312507212
3 56.7780121399792
4 57.1949997084711
5 57.6314687232835
6 57.7939502762026
7 57.8368623449657
8 57.7956090658657
9 57.7527482515476
10 57.7332204789365
11 57.6876455114357
12 57.7295827605379
13 57.7273235994403
14 57.7398033578305
15 57.7234208564744
16 57.6503753370564
17 57.6045951207267
18 57.5567581264404
19 57.5029640973307
20 57.4789065013297
21 57.4236669146346
22 57.3598294182783
23 57.3214934821449
24 57.2927444144319
25 57.2945856510954
26 57.2890148180321
27 57.272981791032
28 57.2558212173682
29 57.2389568320318
30 57.2160441224881
31 57.2034610424259
32 57.1865294912947
33 57.1789937896921
34 57.1677518595826
35 57.1633576735871
36 57.1511253025981
37 57.147908377437
38 57.1421800186962
39 57.1316941072619
40 57.1299320881113
41 57.131104371974
42 57.1240402301111
43 57.1159310915188
44 57.1140794295021
45 57.1113629934264
46 57.1080983054508
47 57.1048221500469
48 57.1030444909522
49 57.1059165987036
50 57.1023776221064
51 57.0973101927404
52 57.0949445090516
53 57.0931454630125
54 57.0912377599579
55 57.091392987084
56 57.0915084618383
57 57.0906055508121
58 57.0920989991144
59 57.0908105168926
60 57.0902474912034
61 57.0904552978794
62 57.0900054034996
63 57.0877524684547
64 57.0864076390149
65 57.0862939391349
66 57.0853582234987
67 57.0851148773901
68 57.0854123767774
69 57.0845692319109
70 57.084039596718
71 57.0832972118155
72 57.0841037620979
73 57.0833550685828
74 57.083698732134
75 57.083706268282
76 57.083238353991
77 57.083287422193
78 57.0830668091054
79 57.0821702794237
80 57.0822223259216
81 57.0812351468314
82 57.0816128868644
83 57.081254070359
84 57.0815702968479
85 57.0816274916447
86 57.081679683311
87 57.0817722746082
88 57.0805277814549
89 57.0803834527256
90 57.080267542864
91 57.0805694132752
92 57.0804255908997
93 57.0803084822127
94 57.0800915230419
95 57.0800390120003
96 57.0799184160729
97 57.0800095181308
98 57.0799636633317
99 57.0800212222742
100 57.0798072431618
};
\addlegendentry{\textbf{UBNA}}
\addplot [ultra thick, color0, mark=+, mark size=2, mark options={solid}]
table {%
0 51.0734520751163
1 54.1294032996581
2 55.4196569211955
3 56.3334817597027
4 56.5575890953228
5 56.9395525693477
6 57.1606717112362
7 57.3095787219327
8 57.4125326488611
9 57.5486904767177
10 57.6866796988707
11 57.7632814861034
12 57.826096826411
13 57.8428111364713
14 57.9251359530192
15 57.9832226008952
16 58.0252672623476
17 58.0972619769569
18 58.133335498737
19 58.13369210019
20 58.1530068340212
21 58.1849987242926
22 58.2121917269998
23 58.2451959964688
24 58.2579751566487
25 58.2704841314711
26 58.2761693227955
27 58.2906531511419
28 58.2912372533191
29 58.3053030966511
30 58.3111621292547
31 58.325946263049
32 58.317579860449
33 58.321196389446
34 58.3265289131432
35 58.3242831382269
36 58.326466744348
37 58.3250059418466
38 58.3323006508066
39 58.3337417476555
40 58.3410730102719
41 58.3412427697282
42 58.3422922416649
43 58.34901027277
44 58.3493996235504
45 58.3543536615936
46 58.3552459301
47 58.3555469735397
48 58.3575654186899
49 58.3610716558936
50 58.3615691348853
51 58.3606636026076
52 58.3605122561595
53 58.3600526902271
54 58.3595906578087
55 58.3593368658259
56 58.3591406001538
57 58.3588989987806
58 58.3587187363614
59 58.3599819771992
60 58.3593859361841
61 58.3595956155542
62 58.3589985234233
63 58.3583583541477
64 58.3594166735525
65 58.3598343471654
66 58.3601932239935
67 58.3604980372196
68 58.360614411924
69 58.3614669864168
70 58.3613182661419
71 58.3612423112751
72 58.3608848419185
73 58.3613512125262
74 58.3607677079823
75 58.3612429344202
76 58.3612151901101
77 58.3611990873559
78 58.3612823238787
79 58.3620893652836
80 58.3619509700325
81 58.3621476766352
82 58.3624424549231
83 58.3631881389321
84 58.3630870089328
85 58.3634261038179
86 58.3640448413584
87 58.3641965482446
88 58.3643415481156
89 58.3648312344178
90 58.3647345177637
91 58.3646333087142
92 58.3644878800414
93 58.3645270141721
94 58.3645543737598
95 58.3645323934826
96 58.3645370724293
97 58.36453400853
98 58.3645329110385
99 58.3645884086634
100 58.3645802999635
};
\addlegendentry{\textbf{UBNA}$^\mathbf{+}$}
\addplot [ultra thick, black, dashed]
table {%
0 51.1
1 51.1
2 51.1
3 51.1
4 51.1
5 51.1
6 51.1
7 51.1
8 51.1
9 51.1
10 51.1
11 51.1
12 51.1
13 51.1
14 51.1
15 51.1
16 51.1
17 51.1
18 51.1
19 51.1
20 51.1
21 51.1
22 51.1
23 51.1
24 51.1
25 51.1
26 51.1
27 51.1
28 51.1
29 51.1
30 51.1
31 51.1
32 51.1
33 51.1
34 51.1
35 51.1
36 51.1
37 51.1
38 51.1
39 51.1
40 51.1
41 51.1
42 51.1
43 51.1
44 51.1
45 51.1
46 51.1
47 51.1
48 51.1
49 51.1
50 51.1
51 51.1
52 51.1
53 51.1
54 51.1
55 51.1
56 51.1
57 51.1
58 51.1
59 51.1
60 51.1
61 51.1
62 51.1
63 51.1
64 51.1
65 51.1
66 51.1
67 51.1
68 51.1
69 51.1
70 51.1
71 51.1
72 51.1
73 51.1
74 51.1
75 51.1
76 51.1
77 51.1
78 51.1
79 51.1
80 51.1
81 51.1
82 51.1
83 51.1
84 51.1
85 51.1
86 51.1
87 51.1
88 51.1
89 51.1
90 51.1
91 51.1
92 51.1
93 51.1
94 51.1
95 51.1
96 51.1
97 51.1
98 51.1
99 51.1
100 51.1
};
\addlegendentry{No Adaptation}
\addplot [ultra thick, color0, dashed]
table {%
0 51.0734520751163
1 54.083513520543
2 55.3241505982888
3 56.0442024652784
4 56.9768117610023
5 56.9192161688233
6 57.2396004963088
7 57.2240411324293
8 57.175407317607
9 57.3103859266542
10 57.3437876571377
11 57.0674883741668
12 56.9195938350325
13 56.809747085206
14 56.7319521405018
15 56.6719787143098
16 56.5775913087945
17 56.5218797230609
18 56.4213019402993
19 56.375750688446
20 56.3412699808934
21 56.2951142832118
22 56.1172983076614
23 56.2676357730423
24 56.340499748957
25 56.2054316573817
26 56.1681171059817
27 56.0654691379091
28 56.0626246894871
29 56.0071811333158
30 55.8711940558646
31 55.8578703662028
32 55.7898752316089
33 55.6029537945112
34 55.2509556469781
35 55.3050907163095
36 55.4287139895104
37 55.3806610997072
38 55.5027109782978
39 55.4786970231772
40 55.3865243625956
41 55.406397639123
42 55.7533544501852
43 55.8762052541976
44 55.9957757352877
45 55.8287350887308
46 55.8775130792692
47 55.5181041884948
48 55.3789916660521
49 55.4193632224985
50 55.6561916780731
51 55.5045708319022
52 55.608507302268
53 55.8285655380509
54 56.1569165296427
55 56.0097974035067
56 55.8367199349296
57 55.7362981052813
58 55.4787461249752
59 55.6445225074089
60 55.8853378347509
61 55.7352650952375
62 55.5845450544854
63 55.611022180174
64 55.67281378283
65 55.7546570337298
66 55.6829195585839
67 55.5052754965201
68 55.4839924487908
69 55.5581088812968
70 55.5156561866315
71 55.5639629270432
72 55.4460710245187
73 55.5054821198598
74 55.5174604748925
75 55.5175623998842
76 55.4866643023394
77 55.2980755245015
78 55.6114026074621
79 55.5706208806649
80 55.6786245902717
81 55.56375125571
82 55.6146233871203
83 55.7755598212222
84 55.8277993235333
85 55.77523095878
86 55.9589735730779
87 55.9920750995478
88 56.1128121125277
89 55.9649607591604
90 55.8053746170051
91 55.7662319171123
92 55.5230949384853
93 55.6539012464505
94 55.7222912622268
95 55.7400694710676
96 55.8768695472257
97 55.8037659377342
98 55.7878081727131
99 55.977356310222
100 55.8509483757865
};
\addlegendentry{\textbf{UBNA}$^\mathbf{0}$}
\end{axis}

\end{tikzpicture}
	\put(-152.5,48){\footnotesize \cite{Li2018d}}
	\caption{\textbf{Comparison to normalization approaches}: Performance on the \textbf{KITTI validation set} for the adaptation from \textbf{Cityscapes} ($\mathcal{D}^{\mathrm{S}}$) \textbf{to KITTI} ($\mathcal{D}^{\mathrm{T}}$) in \textbf{dependence of the adaptation step} $\kappa$ (batch size $B=6$). We also show results when using the statistics from the source domain (no adaptation) and from the target domain (AdaBN by Li~\etal~\cite{Li2018d}).}
	\label{fig:baseline_comparison_cs_to_kitti}
\end{figure}
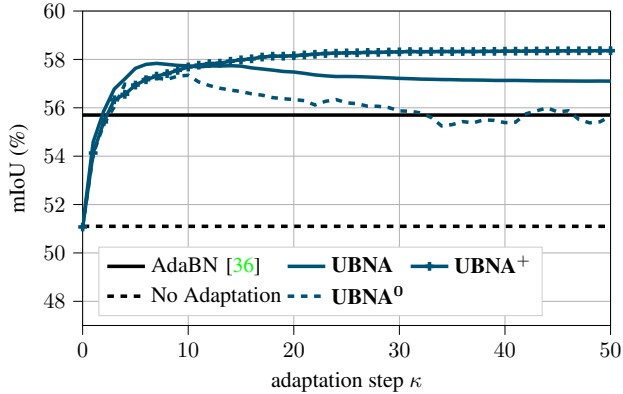

While our main paper's comparison regarding baseline approaches was on the adaptation setting from SYNTHIA ($\mathcal{D}^{\mathrm{S}}$) to Cityscapes ($\mathcal{D}^{\mathrm{T}}$), in this part we want to show the same results for the adaptation from GTA-5 ($\mathcal{D}^{\mathrm{S}}$) to Cityscapes ($\mathcal{D}^{\mathrm{T}}$) and from Cityscapes ($\mathcal{D}^{\mathrm{S}}$) to KITTI ($\mathcal{D}^{\mathrm{T}}$) in Figures~\ref{fig:baseline_comparison_gta_to_cs} and \ref{fig:baseline_comparison_cs_to_kitti}, respectively. Here, we also observe that the UBNA$^{\mathbf{0}}$ method has a peak after adapting with just a few batches, which outperforms the no adaptation and the AdaBN \cite{Li2018d} baselines. Using our UBNA method allows convergence of the performance close to or even above this maximum performance, although there might still be a little potential to optimize the hyperparameter $\alpha^{\mathrm{BATCH}}$ for a more stable convergence at the performance maximum. However, for the scope of this paper we rather wanted to show that our hyperparameter setting generalizes to a large degree over different dataset combinations. The more stable convergence of UBNA$^+$ also shows that with further hyperparameter tuning, the convergence behavior and maximum performance can even be imporved. In conclusion, this shows that our results observed on other datasets are consistent with the results on the SYNTHIA to Cityscapes adaptation setting.

\subsection{Comparison to UDA Approaches}

\begin{table*}[t]
  \centering
  \caption{\textbf{Comparison to UDA approaches}: Performance of \textbf{UBNA} and \textbf{UBNA}$^{\myplus}$ in \textbf{comparison to UDA methods} on the \textbf{Cityscapes validation set} for the adaptation from \textbf{GTA-5} ($\mathcal{D}^{\mathrm{S}}$) \textbf{to Cityscapes} ($\mathcal{D}^{\mathrm{T}}$). We evaluate different models (first column), compare different methods (second column, values from the respective papers), show if they use labeled source data during unsupervised domain adaptation (third column), evaluate the IoU performance (\%) on the single semantic classes, and finally give an mIoU (\%) over 19 classes. For ResNet results we use the \textbf{ResNet-50} topology, as this yielded better results at lower computational complexity compared to the ResNet-101 often used in UDA approaches. Best results for UDA and for UDA without source data are printed in boldface.}
  \footnotesize
  \setlength{\tabcolsep}{1.35pt}
  \begin{tabular}{c|l|c|ccccccccccccccccccc|c}
   \rotatebox{90}{Network} & Method  & \shortstack{Source\\ data \\ during \\[-0.07cm] adaptation} & \rotatebox{90}{road} & \rotatebox{90}{sidewalk} & \rotatebox{90}{building} & \rotatebox{90}{wall$^{**}$} & \rotatebox{90}{fence$^{**}$} & \rotatebox{90}{pole$^{**}$} & \rotatebox{90}{traffic light} & \rotatebox{90}{traffic sign} & \rotatebox{90}{vegetation} & \rotatebox{90}{terrain$^{*}$} & \rotatebox{90}{sky} & \rotatebox{90}{person} & \rotatebox{90}{rider} & \rotatebox{90}{car} & \rotatebox{90}{truck$^{*}$} & \rotatebox{90}{bus} & \rotatebox{90}{on rails$^{*}$} & \rotatebox{90}{motorbike} & \rotatebox{90}{bike} & \shortstack{$\mathrm{mIoU}$ (\%) \\ (19 classes)} \\
  \hline
  \hline
  \multirow{12}{*}{\rotatebox{90}{ResNet-50}} & Du~\etal~\cite{Du2019} & yes & 
  90.3 & 38.9 & 81.7 & 24.8 & 22.9 & 30.5 & 37.0 & 21.2 & 84.8 & 38.8 & 76.9 & 58.8 & 30.7 & 85.7 & 30.6 & 38.1 & 5.9 & 28.3 & 36.9 & 45.4 \\
   & Vu~\etal~\cite{Vu2019} & yes & 
  89.4 & 33.1 & 81.0 & 26.6 & 26.8 & 27.2 & 33.5 & 24.7 & 83.9 & 36.7 & 78.8 & 58.7 & 30.5 & 84.8 & 38.5 & 44.5 & 1.7 & 31.6 & 32.4 & 45.5 \\
   & Li~\etal~\cite{Li2019b} & yes & 
  91.0 & 44.7 & 84.2 & 34.6 & 27.6 & 30.2 & 36.0 & 36.0 & 85.0 & \textbf{43.6} & 83.0 & 58.6 & 31.6 & 83.3 & 35.3 & 49.7 & 3.3 & 28.8 & 35.6 & 48.5 \\
   & Dong~\etal~\cite{Dong2020} & yes & 
  89.6 & 50.4 & 83.0 & 35.6 & 26.9 & 31.1 & 37.3 & 35.1 & 83.5 & 40.6 & 84.0 & 60.6 & 34.3 & 80.9 & 35.1 & 47.3 & 0.5 & 34.5 & 33.7 & 48.6 \\
   & Wang~\etal~\cite{Wang2020} & yes & 
  90.6 & 44.7 & 84.8 & 34.3 & 28.7 & 31.6 & 35.0 & 37.6 & 84.7 & 43.3 & 85.3 & 57.0 & 31.5 & 83.8 & 42.6 & 48.5 & 1.9 & 30.4 & 39.0 & 49.2 \\
   & Yang~\etal\cite{Yang2020a} & yes &
  90.8 & 41.4 & 84.7 & 35.1 & 27.5 & 31.2 & 38.0 & 32.8 & \textbf{85.6} & 42.1 & 84.9 & 59.6 & 34.4 & 85.0 & \textbf{42.8} & 52.7 & 3.4 & 30.9 & 38.1 & 49.5 \\
   & Zhang~\etal~\cite{Zhang2019e} & yes & 
  90.4 & 51.6 & 83.8 & 34.2 & 27.8 & \textbf{38.4} & 25.3 & \textbf{48.4} & 85.4 & 38.2 & 78.1 & 58.6 & \textbf{34.6} & 84.7 & 21.9 & 42.7 & \textbf{41.1} & 29.3 & 37.2 & 50.2 \\
   & Kim~\etal~\cite{Kim2020} & yes & 
  92.9 & 55.0 &  85.3 & 34.2 & \textbf{31.1} & 34.9 & 40.7 & 34.0 & 85.2 & 40.1 & 87.1 & 61.0 & 31.1 & 82.5 & 32.3 & 42.9 & 0.3 & \textbf{36.4} & 46.1 & 50.2 \\
   & Yang~\etal~\cite{Yang2020} & yes & 
  92.5 & 53.3 & 82.4 & 26.5 & 27.6 & 36.4 & 40.6 & 38.9 & 82.3 & 39.8 & 78.0 & 62.6 & 34.4 & 84.9 & 34.1 & \textbf{53.1} & 16.9 & 27.7 & \textbf{46.4} & 50.5 \\
   & Mei~\etal~\cite{Mei2020} & yes & 
  \textbf{94.1} & \textbf{58.8} & \textbf{85.4} & \textbf{39.7} & 29.2 & 25.1 & \textbf{43.1} & 34.2 & 84.8 & 34.6 & \textbf{88.7} & \textbf{62.7} & 30.3 & \textbf{87.6} & 42.3 & 50.3 & 24.7 & 35.2 & 40.2 & \textbf{52.2} \\
  \cline{2-23}
   & \cellcolor[gray]{.90} No adaptation & \cellcolor[gray]{.90} - &\cellcolor[gray]{.90} 58.1 &\cellcolor[gray]{.90} 23.8 &\cellcolor[gray]{.90} 70.5 &\cellcolor[gray]{.90} 14.8 &\cellcolor[gray]{.90} 19.2 &\cellcolor[gray]{.90} 30.5 &\cellcolor[gray]{.90} 29.0 &\cellcolor[gray]{.90} 17.7 &\cellcolor[gray]{.90} \textbf{79.1} &\cellcolor[gray]{.90} \textbf{21.8} &\cellcolor[gray]{.90} \textbf{83.1} &\cellcolor[gray]{.90} 56.4 &\cellcolor[gray]{.90} \textbf{14.8} &\cellcolor[gray]{.90} 72.3 &\cellcolor[gray]{.90} \textbf{19.5} &\cellcolor[gray]{.90} 4.5 &\cellcolor[gray]{.90} 0.9 &\cellcolor[gray]{.90} 16.5 &\cellcolor[gray]{.90} 5.6 &\cellcolor[gray]{.90} 33.6 \\
  \cline{2-21}
   & \cellcolor[gray]{.90} \textbf{UBNA} & \cellcolor[gray]{.90} no &\cellcolor[gray]{.90} \textbf{81.8} &\cellcolor[gray]{.90} \textbf{32.3} &\cellcolor[gray]{.90} \textbf{79.5} &\cellcolor[gray]{.90} \textbf{18.2} &\cellcolor[gray]{.90} \textbf{23.8} &\cellcolor[gray]{.90} \textbf{34.9} &\cellcolor[gray]{.90} \textbf{29.5} &\cellcolor[gray]{.90} \textbf{19.8} &\cellcolor[gray]{.90} 74.2 &\cellcolor[gray]{.90} 17.9 &\cellcolor[gray]{.90} 82.4 &\cellcolor[gray]{.90} \textbf{57.5} &\cellcolor[gray]{.90} 11.1 &\cellcolor[gray]{.90} \textbf{81.6} &\cellcolor[gray]{.90} 16.1 &\cellcolor[gray]{.90} \textbf{19.0} &\cellcolor[gray]{.90} \textbf{2.5} &\cellcolor[gray]{.90} \textbf{21.3} &\cellcolor[gray]{.90} \textbf{9.8} &\cellcolor[gray]{.90} \textbf{37.5}\\
   & \cellcolor[gray]{.90} \textbf{UBNA}$^{\myplus}$ & \cellcolor[gray]{.90} no &\cellcolor[gray]{.90} 70.6 &\cellcolor[gray]{.90} 25.8 &\cellcolor[gray]{.90} 78.5 &\cellcolor[gray]{.90} 17.7 &\cellcolor[gray]{.90} 23.7 &\cellcolor[gray]{.90} 34.2 &\cellcolor[gray]{.90} 28.9 &\cellcolor[gray]{.90} 19.0 &\cellcolor[gray]{.90} 77.8 &\cellcolor[gray]{.90} 19.6 &\cellcolor[gray]{.90} 82.6 &\cellcolor[gray]{.90} 57.4 &\cellcolor[gray]{.90} 11.5 &\cellcolor[gray]{.90} 81.5 &\cellcolor[gray]{.90} 16.9 &\cellcolor[gray]{.90} 17.1 &\cellcolor[gray]{.90} 1.0 &\cellcolor[gray]{.90} 20.6 &\cellcolor[gray]{.90} 9.1 &\cellcolor[gray]{.90} 36.5\\
  \hline
  \hline
  \multirow{11}{*}{\rotatebox{90}{VGG-16}} & Vu~\etal~\cite{Vu2019} & yes & 
  86.9 & 28.7 & 78.7 & 28.5 & 25.2 & 17.1 & 20.3 & 10.9 & 80.0 & 26.4 & 70.2 & 47.1 & 8.4 & 81.5 & 26.0 & 17.2 & 18.9 & 11.7 & 1.6 & 36.1 \\
   & Du~\etal~\cite{Du2019} & yes & 
  88.7 & 32.1 & 79.5 & 29.9 & 22.0 & 23.8 & 21.7 & 10.7 & 80.8 & 29.8 & 72.5 & 49.5 & 16.1 & 82.1 & 23.2 & 18.1 & 3.5 & 24.4 & 8.1 & 37.7 \\
   & Li~\etal~\cite{Li2019b} & yes & 
  89.2 & 40.9 & 81.2 & 29.1 & 19.2 & 14.2 & 29.0 & 19.6 & 83.7 & 35.9 & 80.7 & 54.7 & 23.3 & 82.7 & 25.8 & 28.0 & 2.3 & 25.7 & 19.9 & 41.3 \\
   & Dong~\etal~\cite{Dong2020} & yes & 
  89.8 & 46.1 & 75.2 & 30.1 & \textbf{27.9} & 15.0 & 20.4 & 18.9 & 82.6 & 39.1 & 77.6 & 47.8 & 17.4 & 76.2 & 28.5 & 33.4 & 0.5 & 29.4 & \textbf{30.8} & 41.4 \\
   & Yang~\etal~\cite{Yang2020} & yes & 
  86.1 & 35.1 & 80.6 & 30.8 & 20.4 & 27.5 & 30.0 & 26.0 & 82.1 & 30.3 & 73.6 & 52.5 & 21.7 & 81.7 & 24.0 & 30.5 & \textbf{29.9} & 14.6 & 24.0 & 42.2 \\
   & Kim~\etal~\cite{Kim2020} & yes & 
  \textbf{92.5} & \textbf{54.5} & \textbf{83.9} & \textbf{34.5} & 25.5 & 31.0 & 30.4 & 18.0 & \textbf{84.1} & \textbf{39.6} & \textbf{83.9} & 53.6 & 19.3 & 81.7 & 21.1 & 13.6 & 17.7 & 12.3 & 6.5 & 42.3 \\
   & Wang~\etal~\cite{Wang2020} & yes & 
  88.1 & 35.8 & 83.1 & 25.8 & 23.9 & 29.2 & 28.8 & 28.6 & 83.0 & 36.7 & 82.3 & 53.7 & 22.8 & 82.3 & 26.4 & \textbf{38.6} & 0.0 & 19.6 & 17.1 & 42.4 \\
   & Choi~\etal~\cite{Choi2019} & yes & 
  90.2 & 51.5 & 81.1 & 15.0 & 10.7 & \textbf{37.5} & \textbf{35.2} & \textbf{28.9} & \textbf{84.1} & 32.7 & 75.9 & \textbf{62.7} & 19.9 & 82.6 & 22.9 & 28.3 & 0.0  & 23.0 & 25.4 & 42.5 \\
   & Yang~\etal\cite{Yang2020a} & yes &
  90.1 & 41.2 & 82.2 & 30.3 & 21.3 & 18.3 & 33.5 & 23.0 & \textbf{84.1} & 37.5 & 81.4 & 54.2 & \textbf{24.3} & \textbf{83.0} & \textbf{27.6} & 32.0 & 8.1 & \textbf{29.7} & 26.9 & \textbf{43.6} \\
  \cline{2-23}
   & \cellcolor[gray]{.90} No adaptation & \cellcolor[gray]{.90} - &\cellcolor[gray]{.90} 55.8 &\cellcolor[gray]{.90} 21.9 &\cellcolor[gray]{.90} 65.9 &\cellcolor[gray]{.90} 15.2 &\cellcolor[gray]{.90} 14.7 &\cellcolor[gray]{.90} 27.5 &\cellcolor[gray]{.90} \textbf{31.0} &\cellcolor[gray]{.90} 17.9 &\cellcolor[gray]{.90} \textbf{77.8} &\cellcolor[gray]{.90} \textbf{19.5} &\cellcolor[gray]{.90} 74.4 &\cellcolor[gray]{.90} 55.2 &\cellcolor[gray]{.90} 12.1 &\cellcolor[gray]{.90} 71.7 &\cellcolor[gray]{.90} 11.9 &\cellcolor[gray]{.90} 3.3 &\cellcolor[gray]{.90} 0.5 &\cellcolor[gray]{.90} 13.2 &\cellcolor[gray]{.90} 9.6 & \cellcolor[gray]{.90} 31.5\\
  \cline{2-23}
   & \cellcolor[gray]{.90} \textbf{UBNA} & \cellcolor[gray]{.90} no &\cellcolor[gray]{.90} \textbf{80.8} &\cellcolor[gray]{.90} 29.4 &\cellcolor[gray]{.90} 77.6 &\cellcolor[gray]{.90} 19.8 &\cellcolor[gray]{.90} \textbf{17.1} &\cellcolor[gray]{.90} \textbf{33.9} &\cellcolor[gray]{.90} 29.3 &\cellcolor[gray]{.90} 20.5 &\cellcolor[gray]{.90} 73.9 &\cellcolor[gray]{.90} 16.8 &\cellcolor[gray]{.90} 76.7 &\cellcolor[gray]{.90} 58.3 &\cellcolor[gray]{.90} \textbf{15.2} &\cellcolor[gray]{.90} 79.1 &\cellcolor[gray]{.90} 13.6 &\cellcolor[gray]{.90} 12.5 &\cellcolor[gray]{.90} 5.7 &\cellcolor[gray]{.90} 14.1 &\cellcolor[gray]{.90} \textbf{10.8} & \cellcolor[gray]{.90} 36.1\\
    & \cellcolor[gray]{.90} \textbf{UBNA}$^{\myplus}$ & \cellcolor[gray]{.90} no &\cellcolor[gray]{.90} 79.9 &\cellcolor[gray]{.90} \textbf{29.9} &\cellcolor[gray]{.90} \textbf{78.1} &\cellcolor[gray]{.90} \textbf{21.1} &\cellcolor[gray]{.90} 16.5 &\cellcolor[gray]{.90} 33.8 &\cellcolor[gray]{.90} 29.7 &\cellcolor[gray]{.90} \textbf{20.6} &\cellcolor[gray]{.90} 75.6 &\cellcolor[gray]{.90} 18.4 &\cellcolor[gray]{.90} \textbf{78.0} &\cellcolor[gray]{.90} \textbf{58.4} &\cellcolor[gray]{.90} 14.6 &\cellcolor[gray]{.90} \textbf{79.4} &\cellcolor[gray]{.90} \textbf{14.8} &\cellcolor[gray]{.90} \textbf{13.0} &\cellcolor[gray]{.90} \textbf{5.8} &\cellcolor[gray]{.90} \textbf{14.6} &\cellcolor[gray]{.90} 10.6 & \cellcolor[gray]{.90} \textbf{36.5}\\
  \end{tabular}
  \label{tab:gta5_to_cityscapes_val_sup}
\end{table*}

\begin{table*}[t]
  \centering
  \caption{\textbf{Comparison to UDA approaches}: Performance of \textbf{UBNA} and \textbf{UBNA}$^{\myplus}$ in \textbf{comparison to UDA methods} on the \textbf{Cityscapes validation set} for the adaptation from \textbf{SYNTHIA} ($\mathcal{D}^{\mathrm{S}}$) \textbf{to Cityscapes} ($\mathcal{D}^{\mathrm{T}}$). We evaluate different models (first column), compare different methods (second columns, values from the respective papers), show if they use labeled source data during unsupervised domain adaptation (third column), evaluate the IoU performance (\%) on the single semantic classes, and finally give an mIoU (\%) over 13 classes and over 16 classes (the latter including also wall, fence and pole) as in \cite{Lee2019d, Vu2019}. For ResNet results we use the \textbf{ResNet-50} topology, as this yielded better results at lower computational complexity compared to the ResNet-101 commonly used in other approaches. Best UDA results and best UDA without source data results are printed in boldface. }
  \footnotesize
  \setlength{\tabcolsep}{1.0pt}
  \begin{tabular}{c|l|c|ccccccccccccccccccc|c|c}
   \rotatebox{90}{Network} & Method & \shortstack{Source\\ data \\ during \\[-0.07cm] adaptation}  & \rotatebox{90}{road} & \rotatebox{90}{sidewalk} & \rotatebox{90}{building} & \rotatebox{90}{wall$^{**}$} & \rotatebox{90}{fence$^{**}$} & \rotatebox{90}{pole$^{**}$} & \rotatebox{90}{traffic light} & \rotatebox{90}{traffic sign} & \rotatebox{90}{vegetation} &  \rotatebox{90}{terrain$^{*}$} & \rotatebox{90}{sky} & \rotatebox{90}{person} & \rotatebox{90}{rider} & \rotatebox{90}{car} & \rotatebox{90}{truck$^{*}$} & \rotatebox{90}{bus} & \rotatebox{90}{on rails$^{*}$} & \rotatebox{90}{motorbike} & \rotatebox{90}{bike} & \shortstack{$\mathrm{mIoU}$ (\%) \\ (16 classes)} & \shortstack{$\mathrm{mIoU}$ (\%) \\ (13 classes)}\\
  \hline
  \hline
  \multirow{9}{*}{\rotatebox{90}{ResNet-50}} & Vu~\etal~\cite{Vu2019} & yes &
  85.6 & 42.2 & 79.7 & 8.7 & 0.4 & 25.9 & 5.4 & 8.1 & 80.4 & - & 84.1 & 57.9 & 23.8 & 73.3 & - & 36.4 & - & 14.2 & 33.0 & 41.2 & 48.0
  \\
   & Du~\etal~\cite{Du2019} & yes & 
  84.6 & 41.7 & 80.8 & - & - & - & 11.5 & 14.7 & 80.8 & - & \textbf{85.3} & 57.5 & 21.6 & 82.0 & - & 36.0 & - & 19.3 & 34.5 & - & 50.0\\
   & Li~\etal~\cite{Li2019b} & yes & 
  \textbf{86.0} & \textbf{46.7} & 80.3 & - & - & - & 14.1 & 11.6 & 79.2 & - & 81.3 & 54.1 & 27.9 & 73.7 & - & \textbf{42.2} & - & 25.7 & 45.3 & - & 51.4\\
   & Wang~\etal~\cite{Wang2020} & yes & 
  83.0 & 44.0 & 80.3 & - & - & - & 17.1 & 15.8 & 80.5 & - & 81.8 & 59.9 & \textbf{33.1} & 70.2 & - & 37.3 & - & 28.5 & 45.8 & - & 52.1 \\
   & Yang~\etal~\cite{Yang2020a} & yes & 
  85.1 & 44.5 & 81.0 & - & - & - & 16.4 & 15.2 & 80.1 & - & 84.8 & 59.4 & 31.9 & 73.2 & - & 41.0 & - & 32.6 & 44.7 & - & 53.1 \\
   & Dong~\etal~\cite{Dong2020} & yes & 
  80.2 & 41.1 & 78.9 & \textbf{23.6} & 0.6 & 31.0 & 27.1 & \textbf{29.5} & 82.5 & - & 83.2 & 62.1 & 26.8 & 81.5 & - & 37.2 & - & 27.3 & 42.9 & 47.2 & - \\
   & Mei~\etal~\cite{Mei2020} & yes & 
  81.9 & 41.5 & \textbf{83.3} & 17.7 & \textbf{4.6} & \textbf{32.3} & \textbf{30.9} & 28.8 & \textbf{83.4} & - & 85.0 & \textbf{65.5} & 30.8 & \textbf{86.5} & - & 38.2 & - & \textbf{33.1} & \textbf{52.7} & \textbf{49.8} & \textbf{57.0} \\
  \cline{2-24}
   & \cellcolor[gray]{.90} No adaptation &\cellcolor[gray]{.90} - &\cellcolor[gray]{.90} 36.5 &\cellcolor[gray]{.90} 18.6 &\cellcolor[gray]{.90} 68.3 &\cellcolor[gray]{.90} 2.0 &\cellcolor[gray]{.90} 0.2 &\cellcolor[gray]{.90} 30.3 &\cellcolor[gray]{.90} 6.0 &\cellcolor[gray]{.90} 10.2 &\cellcolor[gray]{.90} 74.5 & \cellcolor[gray]{.90} - &\cellcolor[gray]{.90} 81.6 &\cellcolor[gray]{.90} \textbf{51.9} &\cellcolor[gray]{.90} 10.6 &\cellcolor[gray]{.90} 41.3 & \cellcolor[gray]{.90} - &\cellcolor[gray]{.90} 9.5 & \cellcolor[gray]{.90} - & \cellcolor[gray]{.90} 2.2 &\cellcolor[gray]{.90} 22.6 &\cellcolor[gray]{.90} 29.1 &\cellcolor[gray]{.90} 34.1 \\
  \cline{2-24}
   & \cellcolor[gray]{.90} \textbf{UBNA} &\cellcolor[gray]{.90} no &\cellcolor[gray]{.90} \textbf{62.5} &\cellcolor[gray]{.90} \textbf{22.8} &\cellcolor[gray]{.90} \textbf{75.6} &\cellcolor[gray]{.90} 3.1 &\cellcolor[gray]{.90} \textbf{0.5} &\cellcolor[gray]{.90} 32.5 &\cellcolor[gray]{.90} \textbf{8.6} &\cellcolor[gray]{.90} \textbf{11.3} &\cellcolor[gray]{.90} 73.0 &\cellcolor[gray]{.90} - &\cellcolor[gray]{.90} \textbf{82.7} &\cellcolor[gray]{.90} 42.5 &\cellcolor[gray]{.90} 12.5 &\cellcolor[gray]{.90} \textbf{67.1} & \cellcolor[gray]{.90} -  &\cellcolor[gray]{.90} 12.5 & \cellcolor[gray]{.90} - &\cellcolor[gray]{.90} \textbf{5.7} &\cellcolor[gray]{.90} 27.8 &\cellcolor[gray]{.90} \textbf{33.8} &\cellcolor[gray]{.90} \textbf{39.7} \\
   & \cellcolor[gray]{.90} \textbf{UBNA}$^{\myplus}$ &\cellcolor[gray]{.90} no &\cellcolor[gray]{.90} 57.4 &\cellcolor[gray]{.90} 22.3 &\cellcolor[gray]{.90} 75.0 &\cellcolor[gray]{.90} \textbf{3.6} &\cellcolor[gray]{.90} 0.4 &\cellcolor[gray]{.90} \textbf{33.8} &\cellcolor[gray]{.90} 8.5 &\cellcolor[gray]{.90} 11.1 &\cellcolor[gray]{.90} \textbf{76.2} &\cellcolor[gray]{.90} - &\cellcolor[gray]{.90} 82.6 &\cellcolor[gray]{.90} 46.8 &\cellcolor[gray]{.90} \textbf{12.8} &\cellcolor[gray]{.90} 61.2 & \cellcolor[gray]{.90} -  &\cellcolor[gray]{.90} \textbf{12.8} & \cellcolor[gray]{.90} - &\cellcolor[gray]{.90} 4.8 &\cellcolor[gray]{.90} \textbf{28.5} &\cellcolor[gray]{.90} 33.6 &\cellcolor[gray]{.90} 39.4 \\
  \hline
  \hline
  \multirow{6}{*}{\rotatebox{90}{VGG-16}} & Vu~\etal~\cite{Vu2019} & yes & 
  67.9 & 29.4 & 71.9 & 6.3 & 0.3 & 19.9 & 0.6 & 2.6 & - & 74.9 & 74.9 & 35.4 & 9.6 & 67.8 & - & 21.4 & - & 4.1 & 15.5 & 31.4 & 36.6 \\
   & Lee~\etal~\cite{Lee2019d} & yes & 
  71.1 & 29.8 & 71.4 & 3.7 & 0.3 & \textbf{33.2} & 6.4 & 15.6 & \textbf{81.2} & - & 78.9 & 52.7 & 13.1 & 75.9 & - & 25.5 & - & 10.0 & 20.5 & 36.8 & \textbf{42.4}\\
   & Dong~\etal~\cite{Dong2020} & yes & 
  70.9 & \textbf{30.5} & \textbf{77.8} & \textbf{9.0} & \textbf{0.6} & 27.3 & 8.8 & 12.9 & 74.8 & - & 81.1 & 43.0 & \textbf{25.1} & 73.4 & - & \textbf{34.5} & - & \textbf{19.5} & 38.2 & 39.2 & - \\
   & Yang~\etal~\cite{Yang2020a} & yes & 
  \textbf{73.7} & 29.6 & 77.6 & 1.0 & 0.4 & 26.0 & \textbf{14.7} & \textbf{26.6} & 80.6 & - & \textbf{81.8} & \textbf{57.2} & 24.5 & \textbf{76.1} & - & 27.6 & - & 13.6 & \textbf{46.6} & \textbf{41.1} & - \\
  \cline{2-24}
   & \cellcolor[gray]{.90} No adaptation &\cellcolor[gray]{.90} - &\cellcolor[gray]{.90} 49.4 &\cellcolor[gray]{.90} 20.8 &\cellcolor[gray]{.90} 61.5 &\cellcolor[gray]{.90} \textbf{3.6} &\cellcolor[gray]{.90} 0.1 &\cellcolor[gray]{.90} 30.5 &\cellcolor[gray]{.90} \textbf{13.6} &\cellcolor[gray]{.90} 14.1 &\cellcolor[gray]{.90} 74.4 &\cellcolor[gray]{.90}  - & \cellcolor[gray]{.90} 75.5 &\cellcolor[gray]{.90} \textbf{53.5} &\cellcolor[gray]{.90} 10.6 &\cellcolor[gray]{.90} 47.2 &\cellcolor[gray]{.90}  - & \cellcolor[gray]{.90} 4.8 &\cellcolor[gray]{.90}  - & \cellcolor[gray]{.90} 3.0 &\cellcolor[gray]{.90} 17.1 &\cellcolor[gray]{.90} 30.0 &\cellcolor[gray]{.90} 35.2 \\
  \cline{2-24}
   & \cellcolor[gray]{.90} \textbf{UBNA} &\cellcolor[gray]{.90} no &\cellcolor[gray]{.90} \textbf{72.3} &\cellcolor[gray]{.90} 26.6 &\cellcolor[gray]{.90} \textbf{73.0} &\cellcolor[gray]{.90} 2.3 &\cellcolor[gray]{.90} \textbf{0.3} &\cellcolor[gray]{.90} 31.5 &\cellcolor[gray]{.90} 12.1 &\cellcolor[gray]{.90} 16.6 &\cellcolor[gray]{.90} 72.1 &\cellcolor[gray]{.90}  - & \cellcolor[gray]{.90} \textbf{75.6} &\cellcolor[gray]{.90} 45.4 &\cellcolor[gray]{.90} \textbf{13.6} &\cellcolor[gray]{.90} 61.2 &\cellcolor[gray]{.90}  - & \cellcolor[gray]{.90} \textbf{8.5} &\cellcolor[gray]{.90}  - & \cellcolor[gray]{.90} \textbf{8.5} &\cellcolor[gray]{.90} \textbf{30.1} &\cellcolor[gray]{.90} 34.4 &\cellcolor[gray]{.90} 40.7 \\
   & \cellcolor[gray]{.90} \textbf{UBNA}$^{\myplus}$ &\cellcolor[gray]{.90} no &\cellcolor[gray]{.90} 71.5 &\cellcolor[gray]{.90} \textbf{27.3} &\cellcolor[gray]{.90} 72.9 &\cellcolor[gray]{.90} 2.5 &\cellcolor[gray]{.90} \textbf{0.3} &\cellcolor[gray]{.90} \textbf{32.0} &\cellcolor[gray]{.90} 12.7 &\cellcolor[gray]{.90} \textbf{16.7} &\cellcolor[gray]{.90} \textbf{74.6} &\cellcolor[gray]{.90} - & \cellcolor[gray]{.90} 75.4 &\cellcolor[gray]{.90} 47.1 &\cellcolor[gray]{.90} \textbf{13.6} &\cellcolor[gray]{.90} \textbf{61.4} &\cellcolor[gray]{.90}  - & \cellcolor[gray]{.90} \textbf{8.5} &\cellcolor[gray]{.90}  - & \cellcolor[gray]{.90} 8.3 &\cellcolor[gray]{.90} 29.2 &\cellcolor[gray]{.90} \textbf{34.6} &\cellcolor[gray]{.90} \textbf{41.0} \\
  \end{tabular}
  \label{tab:synthia_to_cityscapes_val_sup}
\end{table*}
\label{sec:benchmark_results_sup}

To facilitate a more extensive comparison with respect to past and future works, we extend our comparison to UDA baseline approaches also to models using a ResNet-based network architecture, which we demonstrate in Tables~\ref{tab:gta5_to_cityscapes_val_sup} and \ref{tab:synthia_to_cityscapes_val_sup}. For our experiments we use a ResNet-50 backbone, which provides a good trade-off between computational complexity and performance. When analyzing the results we observe that UBNA improves the final model performance by 3.9\% and 4.7/5.6\% absolute for the adaptations from GTA-5 to Cityscapes and SYNTHIA to Cityscapes, respectively. Additionally, the performance on the single classes improves for the large majority of classes. This is in consistency with the results obtained using a VGG-16-based model. Interestingly, for ResNet-50, the UBNA$^+$ method does not give an improvement over UBNA, as the hyperparameters were optimized for the VGG-16 backbone. Accordingly, we can see that UBNA has a higher degree of generalization over different network architectures, while UBNA$^+$ is a bit more hyperparameter-sensitive.

\subsection{Few-Shot Capability}

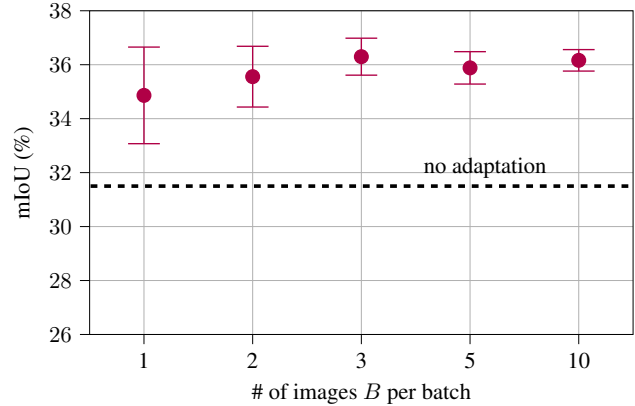
\begin{figure}[t]
	\centering	
\begin{tikzpicture}

\definecolor{color0}{rgb}{0.690196078431373,0,0.274509803921569}

\begin{axis}[
height=6.6cm,
tick align=outside,
tick pos=left,
width=10cm,
x grid style={white!69.01960784313725!black},
xlabel={\# of images \(\displaystyle B\) per batch},
xmajorgrids,
xmin=-0.5, xmax=4.5,
xtick style={color=black},
xtick={-1,0,1,2,3,4,5},
xticklabels={0,1,2,3,5,10,},
y grid style={white!69.01960784313725!black},
ylabel={mIoU (\%)},
ymajorgrids,
ymin=26, ymax=38,
ytick style={color=black}
]
\path [draw=color0, semithick]
(axis cs:0,33.0725716061152)
--(axis cs:0,36.6534194347156);

\path [draw=color0, semithick]
(axis cs:1,34.4309710053722)
--(axis cs:1,36.6829690766727);

\path [draw=color0, semithick]
(axis cs:2,35.6125151672996)
--(axis cs:2,36.9830400997796);

\path [draw=color0, semithick]
(axis cs:3,35.2826789590811)
--(axis cs:3,36.4835105539879);

\path [draw=color0, semithick]
(axis cs:4,35.7640046525327)
--(axis cs:4,36.5602008993169);

\addplot [semithick, color0, mark=-, mark size=7, mark options={solid}, only marks]
table {%
0 33.0725716061152
1 34.4309710053722
2 35.6125151672996
3 35.2826789590811
4 35.7640046525327
};
\addplot [semithick, color0, mark=-, mark size=7, mark options={solid}, only marks]
table {%
0 36.6534194347156
1 36.6829690766727
2 36.9830400997796
3 36.4835105539879
4 36.5602008993169
};
\addplot [ultra thick, black, dashed]
table {%
-5 31.5
-4.84848484848485 31.5
-4.6969696969697 31.5
-4.54545454545454 31.5
-4.39393939393939 31.5
-4.24242424242424 31.5
-4.09090909090909 31.5
-3.93939393939394 31.5
-3.78787878787879 31.5
-3.63636363636364 31.5
-3.48484848484848 31.5
-3.33333333333333 31.5
-3.18181818181818 31.5
-3.03030303030303 31.5
-2.87878787878788 31.5
-2.72727272727273 31.5
-2.57575757575758 31.5
-2.42424242424242 31.5
-2.27272727272727 31.5
-2.12121212121212 31.5
-1.96969696969697 31.5
-1.81818181818182 31.5
-1.66666666666667 31.5
-1.51515151515152 31.5
-1.36363636363636 31.5
-1.21212121212121 31.5
-1.06060606060606 31.5
-0.909090909090909 31.5
-0.757575757575758 31.5
-0.606060606060606 31.5
-0.454545454545454 31.5
-0.303030303030303 31.5
-0.151515151515151 31.5
0 31.5
0.151515151515151 31.5
0.303030303030303 31.5
0.454545454545455 31.5
0.606060606060606 31.5
0.757575757575758 31.5
0.909090909090909 31.5
1.06060606060606 31.5
1.21212121212121 31.5
1.36363636363636 31.5
1.51515151515152 31.5
1.66666666666667 31.5
1.81818181818182 31.5
1.96969696969697 31.5
2.12121212121212 31.5
2.27272727272727 31.5
2.42424242424242 31.5
2.57575757575758 31.5
2.72727272727273 31.5
2.87878787878788 31.5
3.03030303030303 31.5
3.18181818181818 31.5
3.33333333333333 31.5
3.48484848484848 31.5
3.63636363636364 31.5
3.78787878787879 31.5
3.93939393939394 31.5
4.09090909090909 31.5
4.24242424242424 31.5
4.39393939393939 31.5
4.54545454545454 31.5
4.6969696969697 31.5
4.84848484848485 31.5
5 31.5
5.15151515151515 31.5
5.3030303030303 31.5
5.45454545454546 31.5
5.60606060606061 31.5
5.75757575757576 31.5
5.90909090909091 31.5
6.06060606060606 31.5
6.21212121212121 31.5
6.36363636363636 31.5
6.51515151515152 31.5
6.66666666666667 31.5
6.81818181818182 31.5
6.96969696969697 31.5
7.12121212121212 31.5
7.27272727272727 31.5
7.42424242424242 31.5
7.57575757575758 31.5
7.72727272727273 31.5
7.87878787878788 31.5
8.03030303030303 31.5
8.18181818181818 31.5
8.33333333333333 31.5
8.48484848484848 31.5
8.63636363636364 31.5
8.78787878787879 31.5
8.93939393939394 31.5
9.09090909090909 31.5
9.24242424242424 31.5
9.39393939393939 31.5
9.54545454545455 31.5
9.6969696969697 31.5
9.84848484848485 31.5
10 31.5
};
\addplot [semithick, color0, mark=*, mark size=3, mark options={solid}, only marks]
table {%
0 34.8629955204154
1 35.5569700410225
2 36.2977776335396
3 35.8830947565345
4 36.1621027759248
};
\node at (axis cs:2.5,32)[
  anchor=base west,
  text=black,
  rotate=0.0
]{no adaptation};
\end{axis}

\end{tikzpicture}
	\caption{\textbf{Few-shot adaptation}: Performance of \textbf{UBNA}$^{\myplus}$ on the \textbf{Cityscapes validation set} for the adaptation from \textbf{GTA-5} ($\mathcal{D}^{\mathrm{S}}$) \textbf{to Cityscapes} ($\mathcal{D}^{\mathrm{T}}$) for \textbf{different numbers of total images used for adaptation}. We show the mean and the standard deviation over 10 different experiments after the $\kappa=50$th step, whereby in each experiment all batches contain the same image(s).}
	\label{fig:few_shot_applicability_gta}
\end{figure}

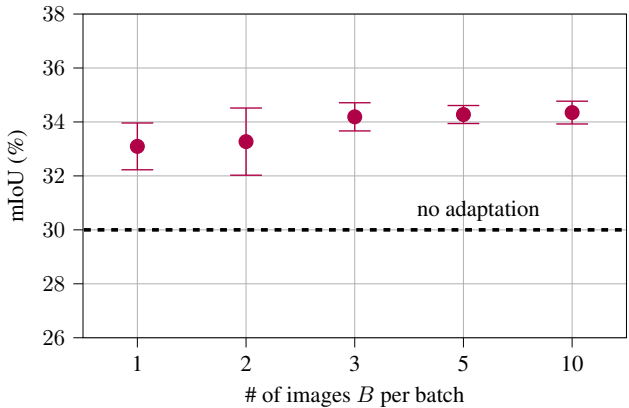
\begin{figure}[t]
	\centering	
\begin{tikzpicture}

\definecolor{color0}{rgb}{0.690196078431373,0,0.274509803921569}

\begin{axis}[
height=6.6cm,
tick align=outside,
tick pos=left,
width=10cm,
x grid style={white!69.01960784313725!black},
xlabel={\# of images \(\displaystyle B\) per batch},
xmajorgrids,
xmin=-0.5, xmax=4.5,
xtick style={color=black},
xtick={-1,0,1,2,3,4,5},
xticklabels={0,1,2,3,5,10,},
y grid style={white!69.01960784313725!black},
ylabel={mIoU (\%)},
ymajorgrids,
ymin=26, ymax=38,
ytick style={color=black}
]
\path [draw=color0, semithick]
(axis cs:0,32.2268671348046)
--(axis cs:0,33.9608164361408);

\path [draw=color0, semithick]
(axis cs:1,32.0257761995327)
--(axis cs:1,34.5158210902014);

\path [draw=color0, semithick]
(axis cs:2,33.6670920880873)
--(axis cs:2,34.7113481460751);

\path [draw=color0, semithick]
(axis cs:3,33.9411918788112)
--(axis cs:3,34.6086470196984);

\path [draw=color0, semithick]
(axis cs:4,33.9249650641159)
--(axis cs:4,34.7695125950518);

\addplot [semithick, color0, mark=-, mark size=7, mark options={solid}, only marks]
table {%
0 32.2268671348046
1 32.0257761995327
2 33.6670920880873
3 33.9411918788112
4 33.9249650641159
};
\addplot [semithick, color0, mark=-, mark size=7, mark options={solid}, only marks]
table {%
0 33.9608164361408
1 34.5158210902014
2 34.7113481460751
3 34.6086470196984
4 34.7695125950518
};
\addplot [ultra thick, black, dashed]
table {%
-5 30
-4.84848484848485 30
-4.6969696969697 30
-4.54545454545454 30
-4.39393939393939 30
-4.24242424242424 30
-4.09090909090909 30
-3.93939393939394 30
-3.78787878787879 30
-3.63636363636364 30
-3.48484848484848 30
-3.33333333333333 30
-3.18181818181818 30
-3.03030303030303 30
-2.87878787878788 30
-2.72727272727273 30
-2.57575757575758 30
-2.42424242424242 30
-2.27272727272727 30
-2.12121212121212 30
-1.96969696969697 30
-1.81818181818182 30
-1.66666666666667 30
-1.51515151515152 30
-1.36363636363636 30
-1.21212121212121 30
-1.06060606060606 30
-0.909090909090909 30
-0.757575757575758 30
-0.606060606060606 30
-0.454545454545454 30
-0.303030303030303 30
-0.151515151515151 30
0 30
0.151515151515151 30
0.303030303030303 30
0.454545454545455 30
0.606060606060606 30
0.757575757575758 30
0.909090909090909 30
1.06060606060606 30
1.21212121212121 30
1.36363636363636 30
1.51515151515152 30
1.66666666666667 30
1.81818181818182 30
1.96969696969697 30
2.12121212121212 30
2.27272727272727 30
2.42424242424242 30
2.57575757575758 30
2.72727272727273 30
2.87878787878788 30
3.03030303030303 30
3.18181818181818 30
3.33333333333333 30
3.48484848484848 30
3.63636363636364 30
3.78787878787879 30
3.93939393939394 30
4.09090909090909 30
4.24242424242424 30
4.39393939393939 30
4.54545454545454 30
4.6969696969697 30
4.84848484848485 30
5 30
5.15151515151515 30
5.3030303030303 30
5.45454545454546 30
5.60606060606061 30
5.75757575757576 30
5.90909090909091 30
6.06060606060606 30
6.21212121212121 30
6.36363636363636 30
6.51515151515152 30
6.66666666666667 30
6.81818181818182 30
6.96969696969697 30
7.12121212121212 30
7.27272727272727 30
7.42424242424242 30
7.57575757575758 30
7.72727272727273 30
7.87878787878788 30
8.03030303030303 30
8.18181818181818 30
8.33333333333333 30
8.48484848484848 30
8.63636363636364 30
8.78787878787879 30
8.93939393939394 30
9.09090909090909 30
9.24242424242424 30
9.39393939393939 30
9.54545454545455 30
9.6969696969697 30
9.84848484848485 30
10 30
};
\addplot [semithick, color0, mark=*, mark size=3, mark options={solid}, only marks]
table {%
0 33.0938417854727
1 33.2707986448671
2 34.1892201170812
3 34.2749194492548
4 34.3472388295838
};
\node at (axis cs:2.5,30.5)[
  anchor=base west,
  text=black,
  rotate=0.0
]{no adaptation};
\end{axis}

\end{tikzpicture}
	\caption{\textbf{Few-shot adaptation}: Performance of \textbf{UBNA}$^{\myplus}$ on the \textbf{Cityscapes validation set} for the adaptation from \textbf{SYNTHIA} ($\mathcal{D}^{\mathrm{S}}$) \textbf{to Cityscapes} ($\mathcal{D}^{\mathrm{T}}$) for \textbf{different numbers of total images used for adaptation}. We show the mean and the standard deviation over 10 different experiments after the $\kappa=50$th step, whereby in each experiment all batches contain the same image(s).}
	\label{fig:few_shot_applicability_synthia}
\end{figure}

To give further insight into the number of images necessary to obtain a stable adaptation using just a few images as shown in Tab.~\textcolor{red}{4}, in Figures~\ref{fig:few_shot_applicability_gta} and \ref{fig:few_shot_applicability_synthia} we show the mean value as well as the standard deviation over the results from ten experiments. We observe that the standard deviation is quite high for only one or two images per batch. The performance slightly increases for three or more images per adaptation batch. To reduce the dependency on a single image, at least three images should be chosen for adaptation, as from this number on, the standard deviation of different experiments is significantly lower and the final performance significantly higher as can be seen in Figures~\ref{fig:few_shot_applicability_gta} and \ref{fig:few_shot_applicability_synthia}.

\subsection{Ablation Studies}

\begin{table}[t]
  \centering
  \setlength{\tabcolsep}{3.5pt}
  \caption{\textbf{Network topology ablation}: Performance of \textbf{UBNA}$^{\mathbf{0}}$, \textbf{UBNA}, and \textbf{UBNA}$^{\myplus}$ on the \textbf{Cityscapes validation set} for the adaptation from \textbf{GTA-5} ($\mathcal{D}^{\mathrm{S}}$) \textbf{to Cityscapes} ($\mathcal{D}^{\mathrm{T}}$)  for \textbf{various network topologies}; best results for each network topology in boldface; mIoU values in \%; batch size $B=6$; $\alpha^{\mathrm{BATCH}} = 0.08$ (\textbf{UBNA}, \textbf{UBNA}$^{\myplus}$); $\alpha^{\mathrm{LAYER}} = 0.03$ (\textbf{UBNA}$^{\myplus}$).}
  \begin{tabular}{|l|c|c|c|c|}
  \hline
  Network & No adaptation & \textbf{UBNA}$^{\mathbf{0}}$ & \textbf{UBNA} & \textbf{UBNA}$^{\myplus}$\\
  \hline
  VGG-16 & 31.5 & 33.9 & 36.1 & \textbf{36.5}\\
  VGG-19 & 31.2 & 35.3 & 37.4 & \textbf{37.5}\\
  \hline
  ResNet-18 & 34.1 & 29.4 & 34.8 & \textbf{35.5}\\
  ResNet-34 & 33.6 & 34.3 & \textbf{36.0} & 35.5\\
  ResNet-50 & 33.6 & 35.6 & \textbf{37.5} & 36.5\\
  ResNet-101 & 30.0 & 33.4 & \textbf{34.6} & 32.6\\
  \hline
  \end{tabular}
  \label{tab:network_architectures_gta_to_cs}
\end{table}

\begin{table}[t]
  \centering
  \setlength{\tabcolsep}{3.5pt}
  \caption{\textbf{Network topology ablation}: Performance of \textbf{UBNA}$^{\mathbf{0}}$, \textbf{UBNA}, and \textbf{UBNA}$^{\myplus}$ on the \textbf{Cityscapes validation set} for the adaptation from \textbf{SYNTHIA} ($\mathcal{D}^{\mathbf{S}}$) \textbf{to Cityscapes} ($\mathcal{D}^{\mathrm{T}}$)  for \textbf{various network topologies}; best results for each network topology in boldface; mIoU values in \%; batch size $B=6$; $\alpha^{\mathrm{BATCH}} = 0.08$ (\textbf{UBNA}, \textbf{UBNA}$^{\myplus}$); $\alpha^{\mathrm{LAYER}} = 0.03$ (\textbf{UBNA}$^{\myplus}$).}
  \begin{tabular}{|l|c|c|c|c|}
  \hline
  Network & No adaptation & \textbf{UBNA}$^{\mathbf{0}}$ & \textbf{UBNA} & \textbf{UBNA}$^{\myplus}$\\
  \hline
  VGG-16 & 30.0 & 32.6 & 34.4 & \textbf{34.6}\\
  VGG-19 & 30.9 & 29.1 & 30.8 & \textbf{31.7}\\
  \hline
  ResNet-18 & 28.6 & 31.3 & \textbf{31.9} & 31.7\\
  ResNet-34 & 27.4 & 30.1 & 30.0 & \textbf{30.3}\\
  ResNet-50 & 29.1 & 32.9 & \textbf{33.8} & 33.6\\
  ResNet-101 & 30.8 & 30.0 & 31.8 & \textbf{34.3}\\
  \hline
  \end{tabular}
  \label{tab:network_architectures_synthia_to_cs}
\end{table}

While the layer-wise weighting (UBNA$^{+}$) worked particularly well in the real-to-real adaptation setting (cf.~Tab.~\textcolor{red}{7}), the success on settings with a larger domain gap is a little ambiguous, as can be seen in Tables~\ref{tab:network_architectures_gta_to_cs} and \ref{tab:network_architectures_synthia_to_cs}. While the UBNA$^{+}$ method always improves on top of the non-adapted baseline, we observe that the performance gains over the UBNA method are rather small and in some cases, we even observe a better performance with UBNA (\eg, the last three rows in Tab.~\ref{tab:network_architectures_gta_to_cs}). Also, we had to reduce the layer-wise weighting factor by a factor of $10$ to a value $\alpha^{\mathrm{LAYER}}=\mathrm{0.03}$ to obtain decent results, which shows the higher hyperparameter dependency compared to the plain UBNA method. Therefore, on a new setting we would recommend to start with the UBNA method and (if possible) tune afterwards with the layer-wise weighting characterizing our UBNA$^{+}$ method.

\section{Additional Qualitative Results}
\label{sec:qualitative_sup}

\begin{figure*}[t] 
	\centering
	\includegraphics{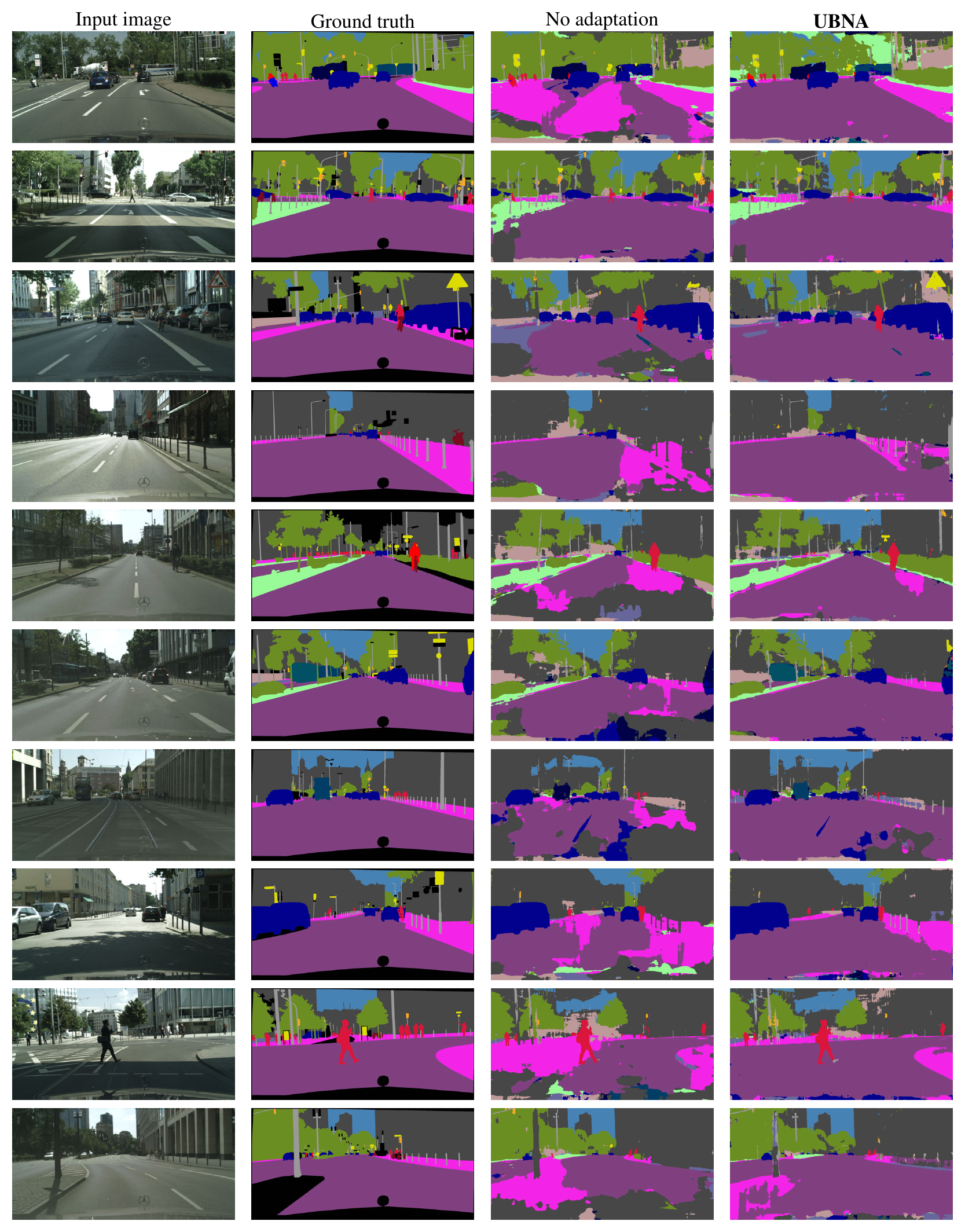}
	\caption{\textbf{Qualitative comparison}: Qualitative results of \textbf{UBNA} in \textbf{comparison to no adaptation and the ground truth} on the \textbf{Cityscapes validation set} for the adaptation from \textbf{GTA-5} ($\mathcal{D}^{\mathrm{S}}$) \textbf{to Cityscapes} ($\mathcal{D}^{\mathrm{T}}$). The figure is best viewed on screen and in color.}
	\label{fig:qualitative_gta_to_cs}
\end{figure*}

\begin{figure*}[t] 
	\centering
	\includegraphics{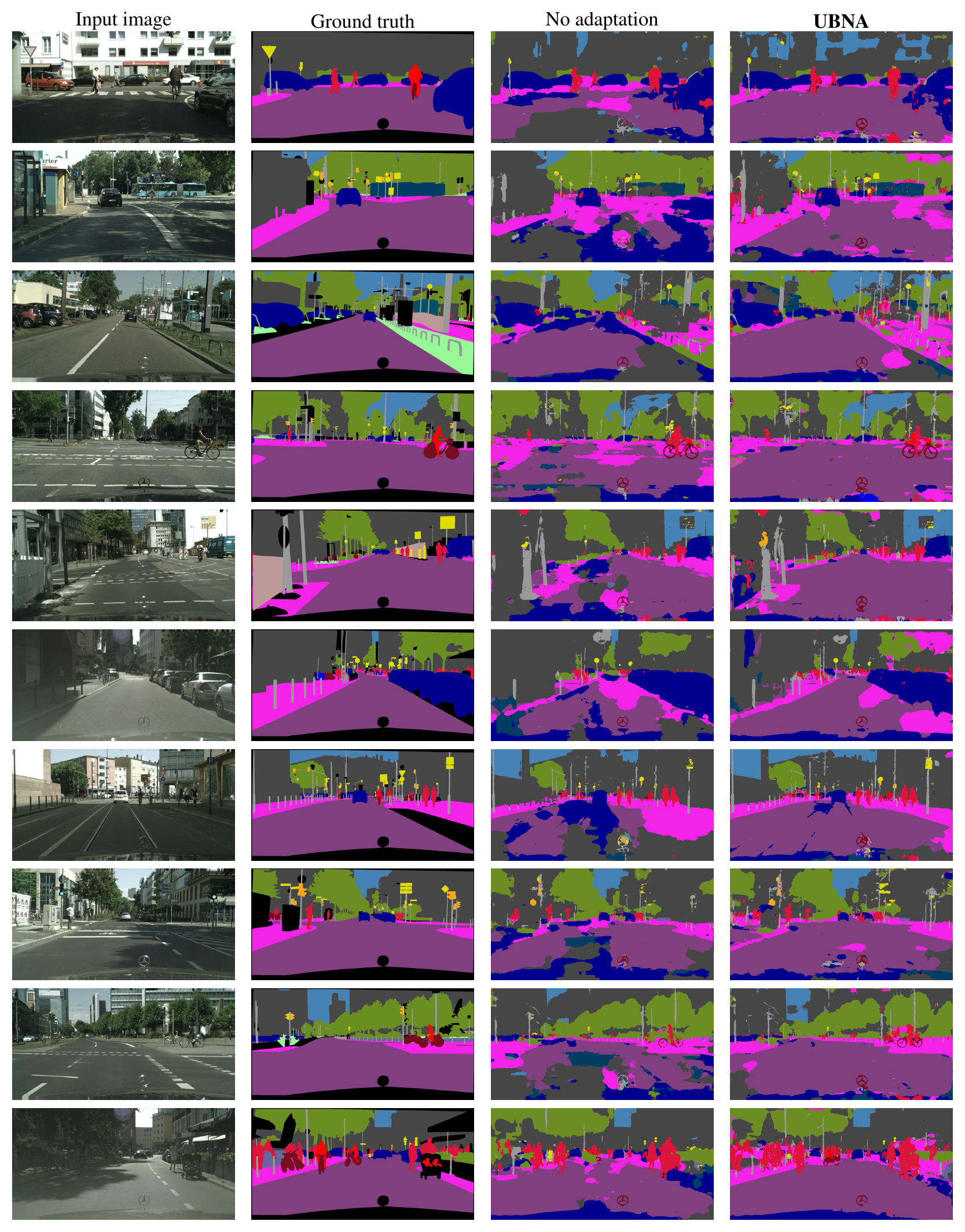}
	\caption{\textbf{Qualitative comparison}: Qualitative results of \textbf{UBNA} in \textbf{comparison to no adaptation and the ground truth} on the \textbf{Cityscapes validation set} for the adaptation from \textbf{SYNTHIA} ($\mathcal{D}^{\mathrm{S}}$) \textbf{to Cityscapes} ($\mathcal{D}^{\mathrm{T}}$). The figure is best viewed on screen and in color.}
	\label{fig:qualitative_synthia_to_cs}
\end{figure*}

\begin{figure*}[t] 
	\centering
	\includegraphics{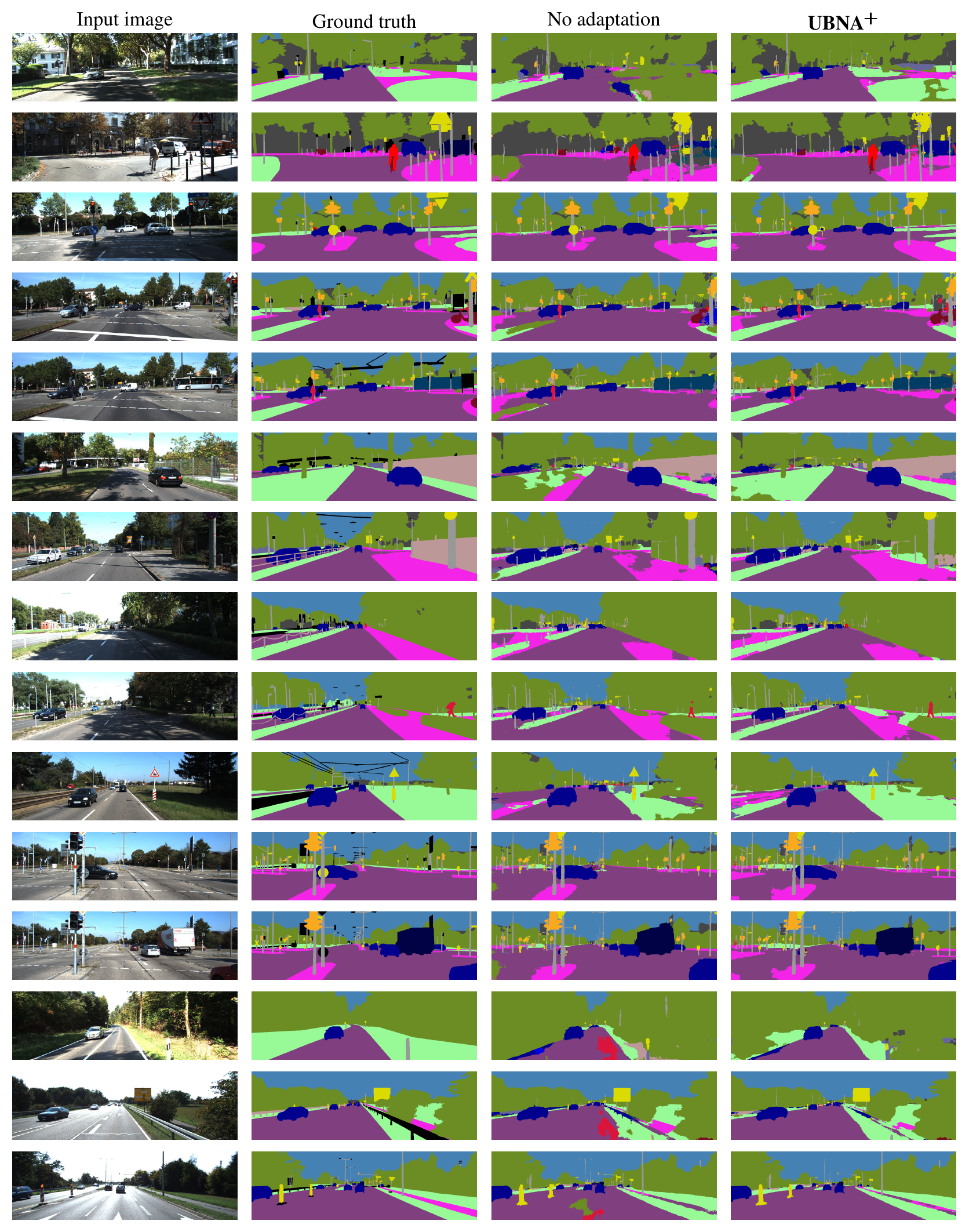}
	\caption{\textbf{Qualitative comparison}: Qualitative results of \textbf{UBNA}$^{\boldsymbol{+}}$ ($\alpha^{\mathrm{LAYER}}=0.3$) in \textbf{comparison to no adaptation and the ground truth} on the \textbf{KITTI validation set} for the adaptation from \textbf{Cityscapes} ($\mathcal{D}^{\mathrm{S}}$) \textbf{to KITTI} ($\mathcal{D}^{\mathrm{T}}$). The figure is best viewed on screen and in color.}
	\label{fig:qualitative_cs_to_kitti}
\end{figure*}

To emphasize also the qualitative effects our method has on the results, we show images of the adaptations from GTA-5 ($\mathcal{D}^{\mathrm{S}}$) to Cityscapes ($\mathcal{D}^{\mathrm{T}}$), SYNTHIA ($\mathcal{D}^{\mathrm{S}}$) to Cityscapes ($\mathcal{D}^{\mathrm{T}}$), and Cityscapes ($\mathcal{D}^{\mathrm{S}}$) to KITTI ($\mathcal{D}^{\mathrm{T}}$) in Figures \ref{fig:qualitative_gta_to_cs}, \ref{fig:qualitative_synthia_to_cs}, and \ref{fig:qualitative_cs_to_kitti}, respectively. In these figures, we show the input images, the ground truth segmentation mask, the predicted segmentation mask when using the model without adaptation, and the results from the adapted model using the UBNA method, from left to right. Interestingly, the qualitative effects are quite similar for all three dataset setups as already observed in the class-wise improvements in Sec.~\ref{sec:benchmark_results_sup}. First of all, the overall artifacts present in different parts of the image are significantly reduced, which is especially well observable on the synthetic-to-real adaptation (\eg, Fig.~\ref{fig:qualitative_gta_to_cs}, row 1, or Fig.~\ref{fig:qualitative_synthia_to_cs}, row 4). However, even in the real-to-real adaptation setting, where the baseline approach already is significantly better, we observe that artifacts are removed by UBNA$^{+}$ from the image, as visible in the last three rows of Fig. \ref{fig:qualitative_cs_to_kitti}, where the human-class artifact in the center of the image is removed. Also, in all three dataset setups, the detection of the road class is significantly improved (\eg Fig.~\ref{fig:qualitative_gta_to_cs}, rows 4 and 5). Also, cars are better distinguished from their surrounding and less wrongfully predicted cars are present in the semantic segmentation prediction (\eg Fig.~\ref{fig:qualitative_synthia_to_cs}, rows 6 and 7). Nevertheless, there are also some classes that are not detected as good as before (cf.~Tables~\ref{tab:gta5_to_cityscapes_val_sup} and \ref{tab:synthia_to_cityscapes_val_sup}) which is also visible qualitatively, \eg, the vegetation class contains slightly more artifacts than before the adaptation (cf.~Fig.~\ref{fig:qualitative_synthia_to_cs}, row 2).

\end{document}